\documentclass[superscriptaddress]{article}

\usepackage{color, colortbl}
\definecolor{LightCyan}{rgb}{0.75,1,1}
\definecolor{LightGray}{rgb}{0.95,0.95,0.95}

\newcolumntype{g}[1]{>{\columncolor{LightGray}}c|}

\newcommand*\rot{\rotatebox{90}}
\newcommand*\rottwo{\rotatebox{45}}

\newcommand*\ZeroshotCLassificationResultsTable{

\begingroup 
\setlength{\tabcolsep}{4.5pt} 
\renewcommand{\arraystretch}{1.5} 
\begin{table*}
\centering
\caption[add short caption]{\textbf{Zeroshot evaluation} top1 accuracy on different datasets. Training CLIP on 63\% of the data gives a higher performance in 17/24 datasets. In the first row, model names are represented by the pruning method (Dedup, Baseline, and Rand for SemDeDup, no pruning, and random pruning respectively), and the fraction of data used for training.}
\label{table:zeroshot_results_table}
\begin{tabular}{l|g|g|g|g|g|g|c|c|c|c|c}
\toprule
        \rottwo{Data / Model} & \rot{Dedup20} & \rot{Dedup40} & \rot{Dedup50} 
        & \rot{Dedup63} & \rot{Dedup72} & \rot{Dedup80} 
        & \rot{Baseline100} & \rot{Rand80} & \rot{Rand60} & \rot{Rand40} & \rot{Rand20} \\ 
        
\midrule

                               Cars &    63.33 &    78.14 &    80.26 &    81.43 &    82.05 &    \textbf{82.61} &        81.42 &   80.96 &   79.26 &   77.74 &   71.57 \\
                         Country211 &    14.16 &    17.74 &    18.26 &    18.44 &    18.70 &    18.20 &   \textbf{19.03} &   18.39 &   16.75 &   15.97 &   12.88 \\
                      Fgvc Aircraft &     4.44 &    11.49 &    12.42 &    \textbf{15.42} &    15.27 &    15.09 &        12.66 &   14.31 &   13.11 &    9.27 &    8.85 \\
                              GTSRB &    38.22 &    37.20 &    36.22 &    43.06 &    41.00 &    35.74 &        42.00 &   \textbf{43.33} &   41.88 &   25.72 &   32.28 \\
                         Imagenet1k &    60.24 &    66.90 &    68.27 &    68.66 &    \textbf{68.93} &    68.80 &        68.74 &   68.29 &   66.12 &   64.82 &   58.86 \\
                              MNIST &    44.29 &    31.87 &    22.93 &    48.55 &    42.75 &    \textbf{48.86} &        33.23 &   43.82 &   35.73 &   36.32 &   19.22 \\
                       Renderedsst2 &    51.46 &    53.65 &    52.72 &    50.80 &    52.99 &    \textbf{57.17} &        51.13 &   52.72 &   51.29 &   52.22 &   45.47 \\
                              STL10 &    96.06 &    96.85 &    97.50 &    \textbf{97.71} &    97.69 &    97.21 &        97.62 &   97.49 &   97.38 &   97.08 &   94.31 \\
                             SUN397 &    64.81 &    67.98 &    68.26 &    68.89 &    69.25 &    \textbf{69.76} &        68.79  &   69.08 &   67.96 &   65.51 &   60.76 \\
                            VOC2007 &    77.94 &    79.51 &    79.74 &    \textbf{80.37} &    79.75 &    78.61 &        80.01 &   77.97 &   79.43 &   77.96 &   74.42 \\
                    Caltech101 &    83.05 &    84.40 &    84.98 &    \textbf{85.06} &    84.35 &    84.75 &        83.42 &   83.69 &   83.93 &   83.38 &   80.62 \\
                      CIFAR100 &    72.09 &    75.71 &    75.17 &    \textbf{77.19} &    77.16 &    77.08 &        74.61 &   76.02 &   74.08 &   72.37 &   67.79 \\
                       CIFAR10 &    92.80 &    93.78 &    94.01 &    94.00 &    \textbf{94.49} &    94.13 &        93.56 &   94.25 &   93.95 &   92.68 &   89.14 \\
               Clevr Dist &    15.75 &    \textbf{23.05} &    15.75 &    19.48 &    21.95 &    21.82 &        23.03 &   18.45 &   15.59 &   18.60 &   16.21 \\
               Clevr Count &    25.37 &    26.36 &    30.85 &    31.87 &    \textbf{34.73} &    20.31 &        24.14 &   15.37 &   26.43 &   14.85 &   21.67 \\
                         DMLAB &    13.16 &    17.62 &    17.99 &    19.20 &    18.52 &    20.23 &        20.46 &   18.50 &   \textbf{21.05} &   17.12 &   19.36 \\
                           DTD &    49.73 &    53.51 &    54.31 &    56.76 &    \textbf{58.94} &    57.66 &        57.34 &   57.02 &   53.35 &   50.96 &   41.76 \\
                       Eurosat &    44.07 &    51.28 &    51.70 &    59.46 &    57.02 &    59.72 &        55.81 &   \textbf{59.81} &   48.63 &   51.26 &   50.00 \\
                       Flowers &    45.21 &    62.21 &    67.67 &    69.78 &    \textbf{70.48} &    66.29 &        67.88 &   68.39 &   65.43 &   62.42 &   58.16 \\
                       Kitti Dist &    20.39 &    13.36 &    14.35 &    14.77 &    19.97 &    \textbf{26.72} &        11.25 &   11.11 &   20.68 &   17.02 &   11.25 \\
                          PCAM &    49.69 &    48.83 &    47.62 &    \textbf{52.66} &    50.09 &    52.14 &        49.09 &   50.11 &   41.28 &   55.02 &   56.59 \\
                          Pets &    77.87 &    87.30 &    89.72 &    90.02 &    90.16 &    90.57  &        \textbf{90.60} &   90.49 &   89.86 &   88.72 &   82.50 \\
                      Resisc45 &    46.76 &    57.56 &    51.69 &    51.49 &    50.14 &    53.57 &        \textbf{57.93} &   54.06 &   51.65 &   49.29 &   46.72 \\
                          SVHN &    34.80 &    33.64 &    26.24 &    \textbf{40.87} &    33.96 &    32.68 &        35.18 &   26.77 &   26.67 &   32.70 &   25.26 \\
                         \midrule
                         Average    &  49.4 &    52.91 &    52.44 &    \textbf{55.66} &    \textbf{55.43} &    \textbf{55.41} &    54.12 &   53.77 &   52.56 &   51.21 &   47.73  \\

\bottomrule
\end{tabular}
\end{table*}
\endgroup

}

\newcommand*\ZeroshotOutOfDistributionResultTable{%

\begingroup 
\setlength{\tabcolsep}{2.5pt} 
\renewcommand{\arraystretch}{1.1} 
\begin{table*}
\centering
\caption[add short caption]{\textbf{Out-of-distribution Robustness} for CLIP models we trained on a different number of examples. The two models trained on 63\% and 72\% of LAION440M with our de-duplication method have higher average accuracy over 6 datasets. In the first column, model names are represented by the pruning method (Dedup, Baseline, and Rand for SemDeDup, no pruning, and random pruning respectively), and the fraction of data used  for training.}
\label{table:out_of_distribution_results_table}

\begin{tabular}{C{1.8cm}|C{1.6cm}C{1.6cm}C{1.6cm}C{1.6cm}C{1.8cm}C{1.2cm}|C{1.2cm}}
\toprule
Model/Dataset     &  ImageNet-A &  ImageNet-O &  ImageNet-R &  ImageNet-Sketch &  ImageNet-V2 &  ObjectNet &  Average \\
\midrule
\rowcolor{LightGray}
Dedup20     &       31.35 &       52.25 &       72.69 &            46.98 &       52.71 &       51.0 &             51.16 \\
\rowcolor{LightGray}
Dedup40     &       38.73 &        49.3 &       77.08 &            51.93 &       59.21 &      54.98 &             55.21 \\
\rowcolor{LightGray}
Dedup50     &       39.68 &       48.55 &       77.74 &            53.54 &       60.37 &      55.36 &             55.87 \\
\rowcolor{LightGray}
Dedup63     &       39.07 &       48.45 &       78.24 &            53.86 &       60.56 &      56.33 &             \textbf{56.08} \\
\rowcolor{LightGray}
Dedup72     &       39.53 &        47.6 &       78.61 &             53.7 &       61.23 &      56.28 &             \textbf{56.16} \\
\rowcolor{LightGray}
Dedup80     &       39.12 &       47.95 &       78.53 &            53.82 &       60.59 &      54.72 &             55.79 \\
Baseline100 &       38.79 &       48.05 &       78.77 &            53.91 &       60.77 &      55.36 &             55.94 \\
Rand80      &       37.87 &        47.7 &       78.04 &            52.81 &       60.02 &       54.3 &             55.12 \\
Rand60      &        34.6 &        47.5 &       75.61 &            51.18 &       57.97 &      53.22 &             53.35 \\
Rand40      &       31.88 &        49.1 &       73.65 &            49.02 &       56.83 &      49.57 &             51.67 \\
Rand20      &       23.43 &        49.4 &       66.74 &            43.76 &       50.67 &      43.57 &             46.26 \\
\bottomrule
\end{tabular}
\end{table*}
\endgroup
}

\usepackage[preprint]{corl_2022}

\usepackage{microtype}
\usepackage{graphicx}
\usepackage{caption}
\usepackage{subcaption}
\usepackage{booktabs} 
\usepackage[export]{adjustbox}
\usepackage[wide]{sidecap}
\usepackage{array}
\usepackage{wrapfig,lipsum,booktabs}
\usepackage{url}

\usepackage{multirow}
\usepackage{algorithm2e}
\SetKwComment{Comment}{/* }{ */}

\newcolumntype{L}[1]{>{\raggedright\let\newline\\\arraybackslash\hspace{0pt}}m{#1}}
\newcolumntype{C}[1]{>{\centering\let\newline\\\arraybackslash\hspace{0pt}}m{#1}}
\newcolumntype{R}[1]{>{\raggedleft\let\newline\\\arraybackslash\hspace{0pt}}m{#1}}

\usepackage{color, colortbl}
\definecolor{myorange}{RGB}{255, 199, 51}
\definecolor{mybrown}{RGB}{239, 116, 25}
\definecolor{myblue}{RGB}{51, 199, 255}
\definecolor{mygray}{gray}{0.6}
\definecolor{LightCyan}{rgb}{0.75,1,1}

\definecolor{codegreen}{rgb}{0,0.6,0}
\definecolor{codegray}{rgb}{0.5,0.5,0.5}
\definecolor{codepurple}{rgb}{0.58,0,0.82}
\definecolor{backcolour}{rgb}{0.95,0.95,0.92}

\usepackage{listings}

\lstdefinestyle{mystyle}{
    backgroundcolor=\color{backcolour},   
    commentstyle=\color{codegreen},
    keywordstyle=\color{magenta},
    numberstyle=\tiny\color{codegray},
    stringstyle=\color{codepurple},
    basicstyle=\ttfamily\footnotesize,
    breakatwhitespace=false,         
    breaklines=true,                 
    captionpos=b,                    
    keepspaces=true,                 
    numbers=left,                    
    numbersep=5pt,                  
    showspaces=false,                
    showstringspaces=false,
    showtabs=false,                  
    tabsize=2
}

\usepackage{hyperref}

\usepackage[T1]{fontenc}
\usepackage[outline]{contour}
\renewcommand{\arraystretch}{1.5}
\contourlength{0.1pt}
\contournumber{10}

\usepackage{amsmath}
\usepackage{amssymb}
\usepackage{mathtools}
\usepackage{amsthm}
\usepackage{wrapfig}
\usepackage{multirow}
\usepackage[compact]{titlesec}      
\usepackage{breakcites}
\usepackage{cite}

\usepackage[capitalize,noabbrev]{cleveref}

\theoremstyle{plain}

\theoremstyle{definition}

\theoremstyle{remark}

\title{SemDeDup: Data-efficient learning at web-scale through semantic deduplication}

\author{
  Amro Abbas$^{1}$
  \hspace{0.5cm}
  Kushal Tirumala$^{1}$\thanks{Equal Contribution. Correspondence to: Amro Abbas amroabbas@meta.com, Ari Morcos arimorcos@meta.com}
  \hspace{0.5cm}
  D\'{a}niel Simig$^{1*}$
  \hspace{0.5cm}
  Surya Ganguli$^{2}$
  \hspace{0.5cm}
  Ari S. Morcos$^{1*}$\\
  $^1$Meta AI (FAIR)\hspace{0.5cm}$^2$Department of Applied Physics, Stanford University\\
}

\begin{document} 

\maketitle

\begin{abstract}
Progress in machine learning has been driven in large part by massive increases in data. However, large web-scale datasets such as LAION are largely uncurated beyond searches for exact duplicates, potentially leaving much redundancy. Here, we introduce SemDeDup, a method which leverages embeddings from pre-trained models to identify and remove ``semantic duplicates'': data pairs which are semantically similar, but not exactly identical. Removing semantic duplicates preserves performance and speeds up learning. Analyzing a subset of LAION, we show that SemDeDup can remove 50\% of the data with minimal performance loss, effectively halving training time. Moreover, performance increases out of distribution. Also, analyzing language models trained on C4, a partially curated dataset, we show that SemDeDup improves over prior approaches while providing efficiency gains. SemDeDup provides an example of how simple ways of leveraging quality embeddings can be used to make models learn faster with less data.   
\end{abstract}

\section{Introduction}
\label{sec:introduction}


\begin{figure}[ht!]
\centering
\begin{subfigure}{.4\columnwidth}
    \centering
    \includegraphics[width=\columnwidth]{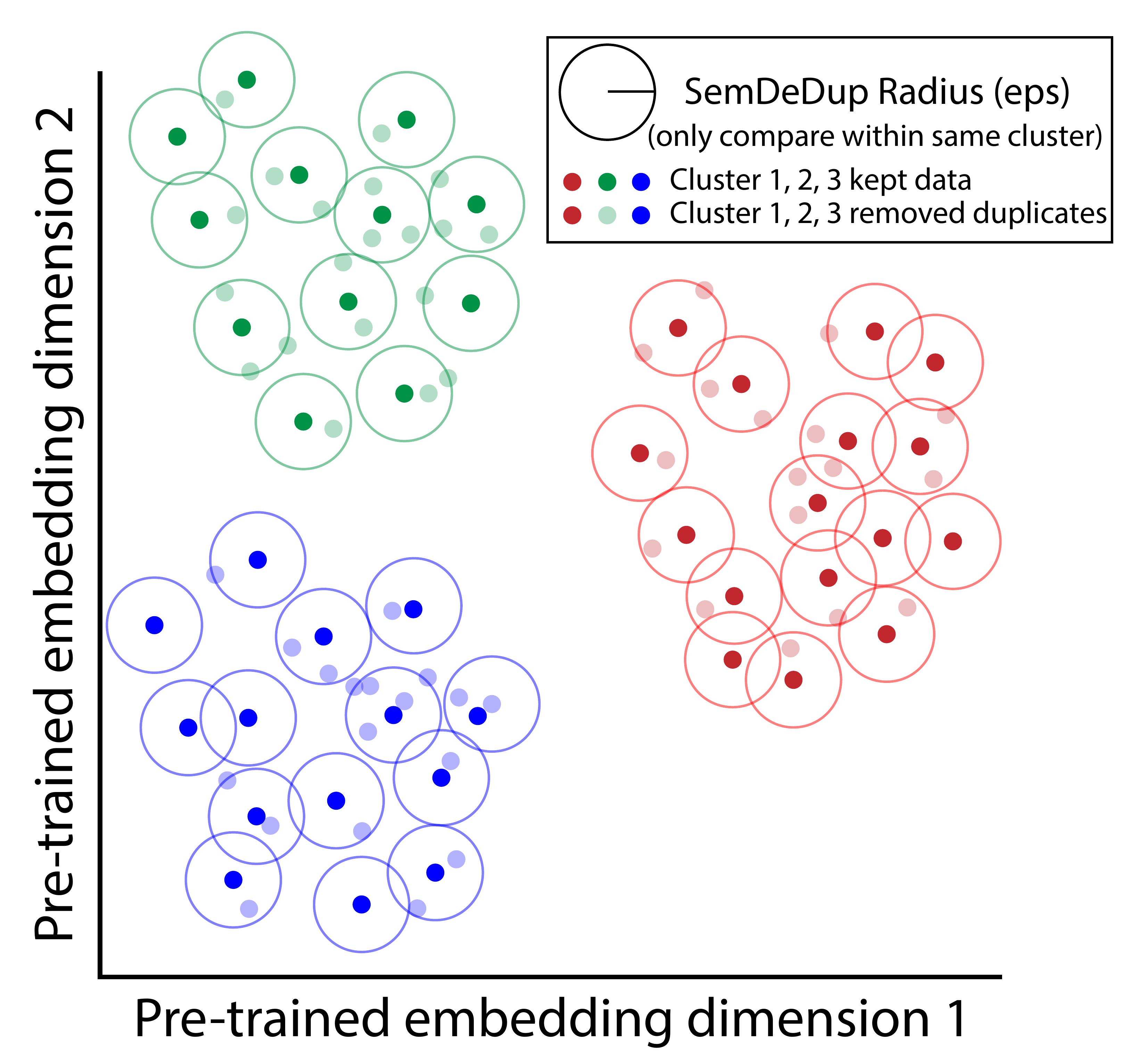}
    \caption{}

\end{subfigure}
\hfill
\begin{subfigure}{.5\columnwidth}
    \centering
    \includegraphics[width=\columnwidth]{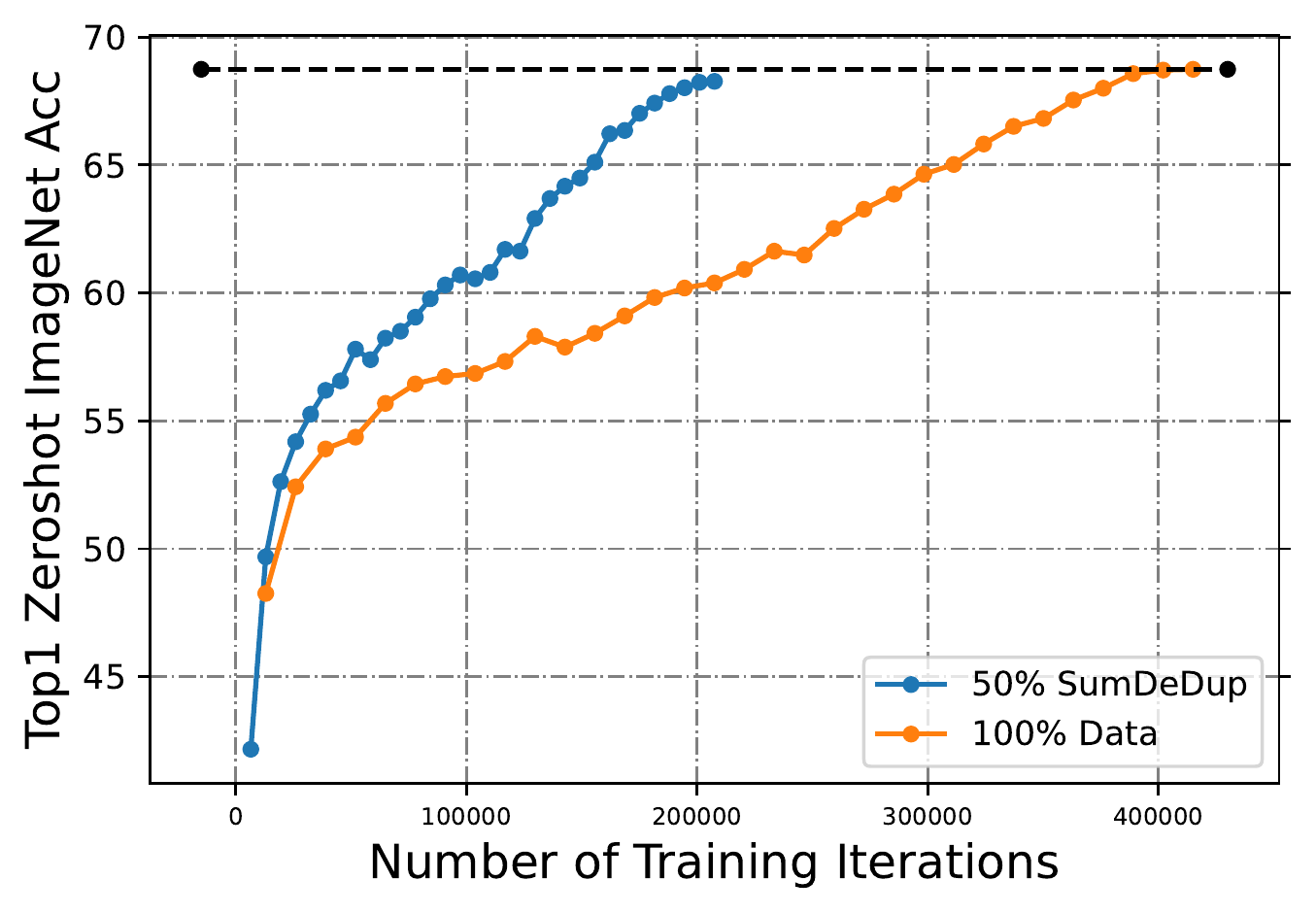}
    \caption{}

\end{subfigure}
 
\caption{\textbf{Data efficiency from semantic deduplication (SemDeDup)} (a): A schematic of the SemDeDup algorithm which efficiently removes semantic duplicates from web-scale data. (b): When SemDeDup removes 50\% of the LAION-440M dataset, training on this semantically {\it nonredundant} subset achieves almost the {\it same} performance as training on the {\it entire} 440M dataset. Also, training speed is {\it twice} as fast and completes in {\it half} the time.}
\label{fig:introfig}

\end{figure}

\begin{figure*}[ht!]
\centering
\begin{center}
\includegraphics[width=\textwidth]{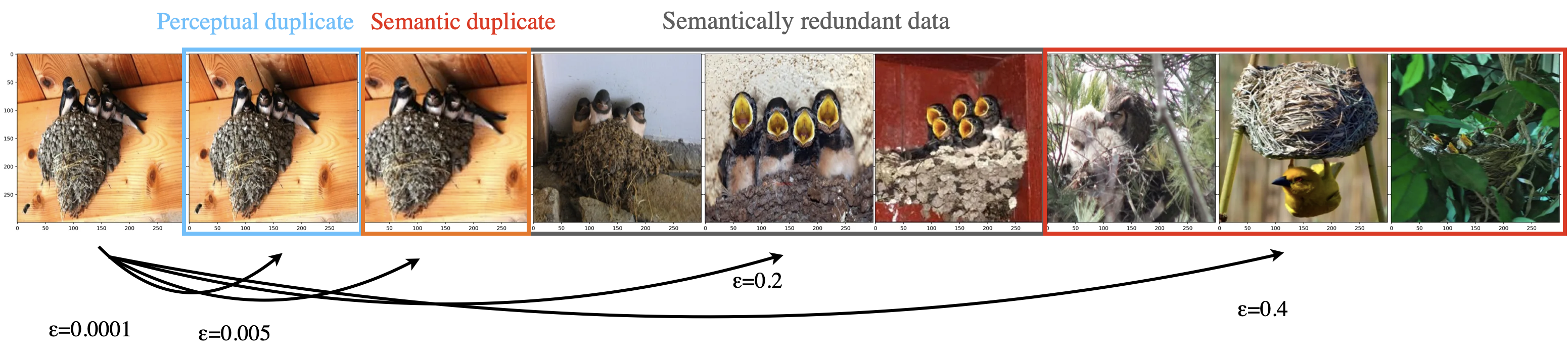}
\caption{\textbf{Mapping cosine similarity to perceptual and semantic similarity.} We visualize pairs of images with cosine similarity $1-\epsilon$ in the CLIP image encoder embedding space. The left most image is a random seed image from LAION, while the remaining images are sorted by their dissimilarity $\epsilon$ to the seed image. Roughly, as $\epsilon$ increases from left to right, we move from perceptual to semantic duplicates, while at large values of $\epsilon$ we see semantically redundant pairs. Note the red labelled ``semantic duplicate" is a view of the original left-most seed image from a slightly different perspective. We visualize more examples in Figure \ref{fig:semantically_similar_images_appendix}.}
\label{fig:visduplicates}
\end{center}
\end{figure*}

A primary driver of recent success in machine learning has been the rise of self-supervised learning (SSL) scaled to ever larger models and unlabelled datasets \citep{Hestness2017-yq,Kaplan2020-ti,Henighan2020-jf,rosenfeld2020a,Gordon2021-az,Hernandez2021-ix,Zhai2021-dl,HoffmannChinchilla}. In particular, modern large datasets are often derived at global web-scale and are generally unfiltered, with the exception of NSFW filters. One such public dataset is LAION \citep{schuhmann2022laion}, a multi-modal dataset of 5 billion image/text pairs. Multi-modal models such as CLIP \citep{radford2021learning} are trained for many epochs on these large datasets achieving impressive performance but at the cost of extremely long training durations. 

The critical role of large datasets has led to increasing interest in scaling laws which enable us to predict how a model's performance will change given more data and/or parameters, leading to the observation that test error generally scales as a power law with respect to data quantity \citep{Kaplan2020-ti}. Power law scaling, however, is unsustainable as diminishing marginal returns are quickly hit such that ever increasing amounts of data are required to achieve ever diminishing improvements in performance. Notably, many of these models appear never to converge, as test performance continues to increase even after 10s of passes through these massive datasets \citep{openclip, multi-modal-scaling}. This result suggests that our best models are underfitting, likely as a result of spending an increasing fraction of learning time focusing on redundant data. 

Improving data efficiency would therefore be quite impactful, either by enabling models to achieve the same performance much faster, or by enabling models to achieve better performance given the same computational budget. These observations have inspired recent work which suggests that by pruning training data according to an intelligent criterion, power law scaling with respect to data can be beaten and, given an optimal data ranking metric, exponential scaling might in principle be achieved \citep{Sorscher2022-wo}. Recent explorations of this direction have shown promising results, with some works able to reduce data size by almost 5-fold with minimal performance loss \citep{radenovic2023filtering}.

However, optimal approaches to select data remain poorly understood. Such approaches might focus on one of several different classes of examples to be removed, roughly ordered by the complexity of their discovery:

\begin{enumerate}
    \item \textbf{Perceptual duplicates}: We loosely define such data pairs to be perceptually identical to a typical human observer. The most straightforward version would be exact duplicates at the pixel or token level that could easily be found via exact duplicate detection in input space. However, such approaches might miss pairs of images with human imperceptible pixel level distortions. Most widely-used datasets have some exact duplicate filter already applied, though perceptual duplicates with slight pixel-level differences may pass through such filters. 
    \item \textbf{Semantic duplicates}: these are examples which contain largely identical information content, but remain perceptually distinct. For example, a pair of image views which are derived from the same image, but feature different margins, aspect ratios, color distributions, etc. could be considered semantic duplicates. A pair of sentences with the same structure but some words exchanged for synonyms would also be considered a semantic duplicate. Such pairs would rarely, if ever, be detected by exact duplicate filters as they would be far apart in pixel/token space. 
    \item \textbf{Semantically redundant data}: in contrast to semantic duplicates, semantically redundant data are not derived from the same underlying objects and would be clearly distinguishable to a human. However, the information contained in such examples may still contain substantial overlap. For example, consider the case of two different images of two different golden retrievers in two different parks. These images are neither perceptually nor semantically identical as the content of the images differs. However, the information contained in them is quite similar, leading us to think of such pairs as semantically redundant. Each additional semantically redundant data point will provide less and less new information, eventually converging to near-zero information gained from additional such data. Methods such as SSL Prototypes \citep{Sorscher2022-wo} and memorization \citep{Feldman2020-yv} search for semantically {\it non-redundant} data subsets to train on.  
    \item \textbf{Misleading data}: these are data which rather than providing zero information (as in the previous categories) provide negative or \textit{harmful} signal, in the sense that removing these data actually improves performance, rather than having a neutral effect. While such data are easy to conceive of in supervised learning (i.e. mislabeled examples), it is much less clear what such examples may be in the context of self-supervised learning.
\end{enumerate}

In this work, we focus on the category of semantic duplicates: data which are semantically highly similar but which would be difficult to discover using simple deduplication approaches. These data points are challenging to identify because distance measures in input space are unlikely to uncover semantic duplicates. To overcome this limitation, we leverage pre-trained foundation models to compare data similarity in the learned embedding space rather than in input space. Comparing every data point to every other data point, however, is intractable, especially for web-scale datasets containing billions of examples. To make this computation possible, we use the clustering approach described in \citep{Sorscher2022-wo} to segment the embedding space, allowing us to only search for duplicate pairs within a cluster. Using this approach, we make the following contributions: 

\begin{itemize}
    \item We propose SemDeDup (Fig. \ref{fig:introfig}, a), a simple, yet effective and computationally tractable way to identify semantic duplicates. Using this approach, we show that large web-scale datasets such as LAION contain large numbers of semantic duplicates, with 50\% of examples containing at least one semantic duplicate.  
    \item Large fractions of semantic duplicates can be removed with little-to-no performance impact, greatly increasing training efficiency. We reduced the size of our LAION training set by 50\% with minimal performance loss, and improved learning speed, achieving nearly the same performance 2x faster (Fig. \ref{fig:introfig}, b), and moreover \textit{improved} performance out-of-distribution.
    \item We apply SemDeDup to C4, a large text corpus, beating prior SoTA deduplication while providing efficiency gains of 15\%, sometimes even improving performance. 
\end{itemize}

Overall, our results demonstrate a simple yet surprisingly effective approach to reduce the cost of training through the removal of semantic duplicates which is likely applicable to all web-derived datasets and may help to democratize the training of large-scale foundation models by improving data and compute efficiency.

\section{Related Work} \label{sec:related_work}
Much of the work in language and vision on deduplication has focused on the removal of exact duplicates. For example, \citep{Liao2022-ft} removed duplicates between the YFCC15M dataset \citep{Thomee2016-zd} and the ImageNet validation set to prevent train-test leakage.  The C4 text corpus - used for training T5 \citep{ColinT5} - has been deduplicated by discarding repeated occurrences of any three-sentence spans.  \citep{LeeDedup} showed that it’s possible to further deduplicate this dataset without loss of performance by computing approximate n-gram overlap between documents using the MinHash technique \citep{BroderMinHash}. \citep{RaeGopher} also applied MinHash based deduplication to curate training data for the Gopher model and demonstrated that training on the deduplicated dataset can result in lower perplexity across various validation sets. ~\citep{KandpalPrivacy}  found that deduplication prevents memorization in LLMs and thus mitigates privacy concerns. More recent works use forms of model-based feature extraction to improve the robustness of the similarity metric used for deduplication. ~\citep{Silcock2022-tp} created a supervised dataset for detecting duplicate news articles and trained models to predict those labels.  In the domain of computer vision, ~\citep{Choi2022-pg} improves on SSL techniques by removing near-duplicates in some high dimensional feature space they learn.


Beyond deduplication, a host of classical machine learning approaches seek to achieve data efficiency by finding {\it coresets}, defined as small subsets of the training data that can be used to train a machine learning algorithm to the same test accuracy achievable when training on the entire training data (see e.g. \citep{Guo2022-qy,Phillips2016-ob} for reviews). However, many coreset algorithms are computationally prohibitive and therefore are difficult to scale to web-scale data.  In contrast to many traditional coreset algorithms, we develop an exceedingly simple and tractable algorithm that achieves both computational and data efficiency at scale.  

Recent approaches to achieve data efficiency in deep learning have operated in a supervised setting by defining and finding ``hard'' examples not easily learned by partially or fully trained (ensembles of) models \citep{Toneva2019-hj,Paul2021-ci, Chitta2021-se,Feldman2020-yv,meding2022trivial}.  Perhaps the closest to our work is a recent effort to break beyond neural power law scaling by pruning unlabelled data, using the embedding space of a pre-trained foundation model \citep{Sorscher2022-wo}.  However, the largest dataset for which these works examined data pruning was ImageNet. In contrast, we move from relatively small, highly curated ImageNet scale to highly uncurated, web-scale data. Our analysis, at this new large and uncurated scale, reveals a possibly fundamental role for semantic deduplication as an important initial step in data-pruning for self-supervised learning that was not considered in prior data-pruning works.

\section{SemDeDup} \label{sec:methods}

\paragraph{Defining and identifying semantic duplicates} \label{sec:methods_id_dup}

While identifying perceptual duplicates can be easily done in input space, identifying semantic duplicates is more difficult as they may be distant in either pixel or token space. To identify these pairs, we leverage the embedding space of a large pre-trained foundation model to provide a more semantically meaningful distance metric. To detect and remove semantically similar images, we use the following semantic de-duplication (SemDeDup) algorithm (Fig. \ref{fig:introfig}, a). First, we embed each data point using a foundation model (CLIP \citep{openclip, clip} for images and OPT \citep{ZhangOPT} for language). We then cluster the embeddings into $k$ clusters via k-means. Below, we choose $k=50,000$ clusters in CLIP image encoder embeddings and $k=11,000$ clusters in OPT-language model embeddings. 
Within each cluster, we compute all pairwise cosine similarities and set a threshold cosine similarity above which data pairs are considered semantic duplicates.
Finally, from each group of semantic duplicates within a cluster, we keep the image with the lowest cosine similarity to the cluster centroid and remove the rest. We note that to determine duplicates, this method considers only the images and ignores the captions. A simplified pseudo code for SemDeDup is shown in Algorithm \ref{alg:semdedup} in the appendix. We provide more details about the method in addition to experiments on choosing the value of $k$ in section \ref{sec:more_analysis}.

\paragraph{Utilizing pre-trained foundation Models} Our method makes use of pre-trained foundation models to embed data examples. Considering that there are many of these ready-to-use pre-trained models available to the public, we can use embeddings from these models to guide curation of other datasets. Pre-trained models like Vision Transformers \citep{vit} for vision tasks, OPT \citep{ZhangOPT} for natural language and CLIP \citep{clip} for vision-language data have been used widely. In this work, we utilize pre-trained CLIP and OPT models for deduplication. In addition, in Section \ref{sec:more_analysis}, we show that one can effectively use an on-the-shelf model pre-trained on one dataset to prune another dataset resulting in a considerable training cost saving.

\paragraph{Clustering to reduce computation} \label{sec:methods_clustering}

The time complexity of naive de-duplication is $\mathcal{O}(n^2)$ where $n$ is the number of data points, making this approach impractical for large web-scale data. For example, the LAION-440M dataset would require $\approx 1.9 \mathrm{x} 10^{17}$ similarity computations. The k-means clustering step in SemDeDup reduces this complexity substantially from $\mathcal{O}(n^2)$ to $\mathcal{O}(n^2/k)$ assuming approximately uniform cluster size\footnote{Note that our choice of $k$ depends on $n$, it is not a constant in the context of this complexity analysis.}. This means we only require $\approx 4.6 \mathrm{x} 10^{12}$ intra-cluster comparisons instead of $\approx 1.9 \mathrm{x} 10^{17}$ across all pairs, a 5-order of magnitude improvement. 


\begin{figure}[ht]
\vskip 0.2in
\begin{center}
\begin{subfigure}{.32\columnwidth}
\includegraphics[width=\columnwidth]
{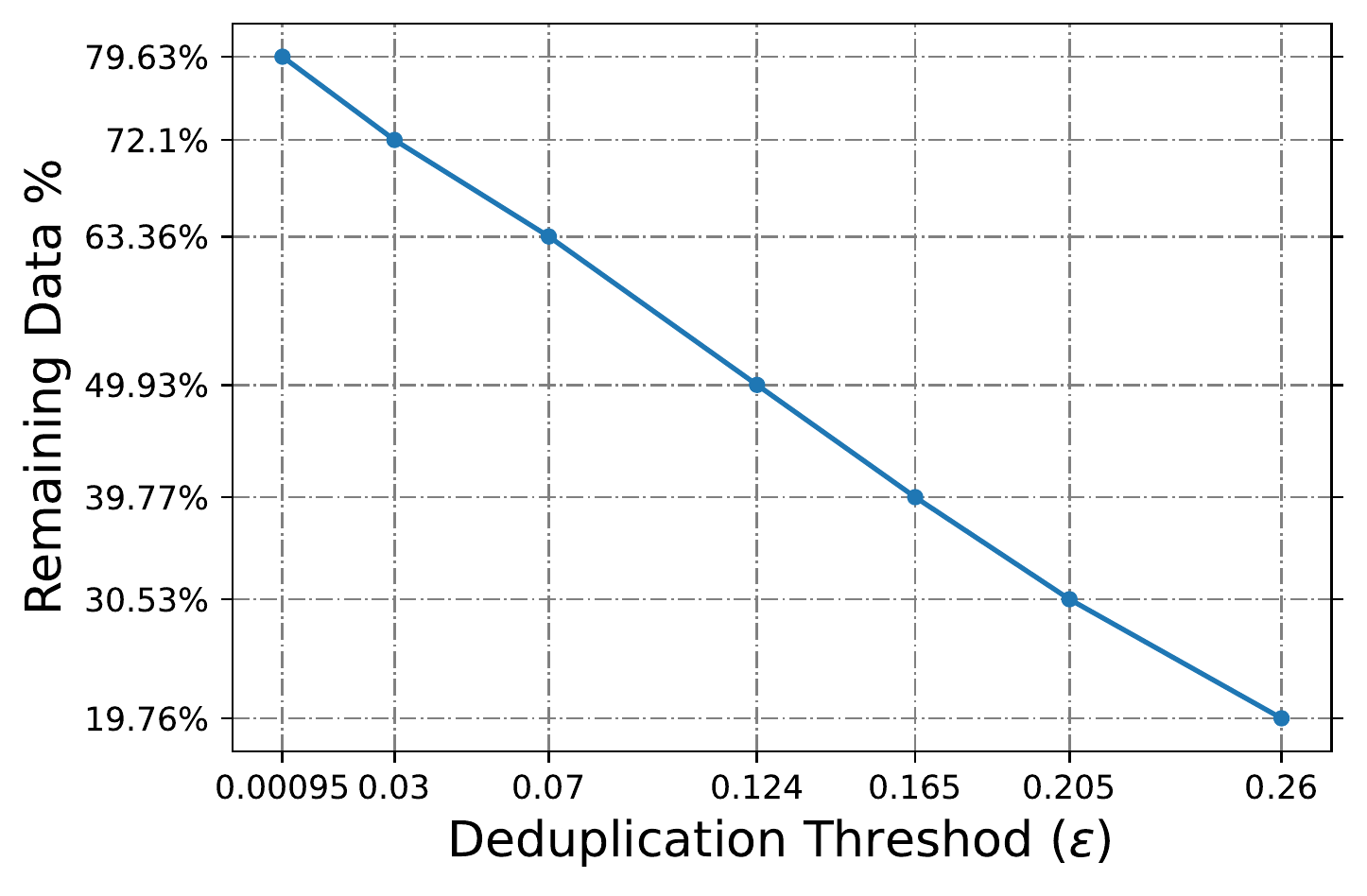}
\caption{}
\end{subfigure}
\begin{subfigure}{.32\columnwidth}
\includegraphics[width=\columnwidth, height=0.658\columnwidth]{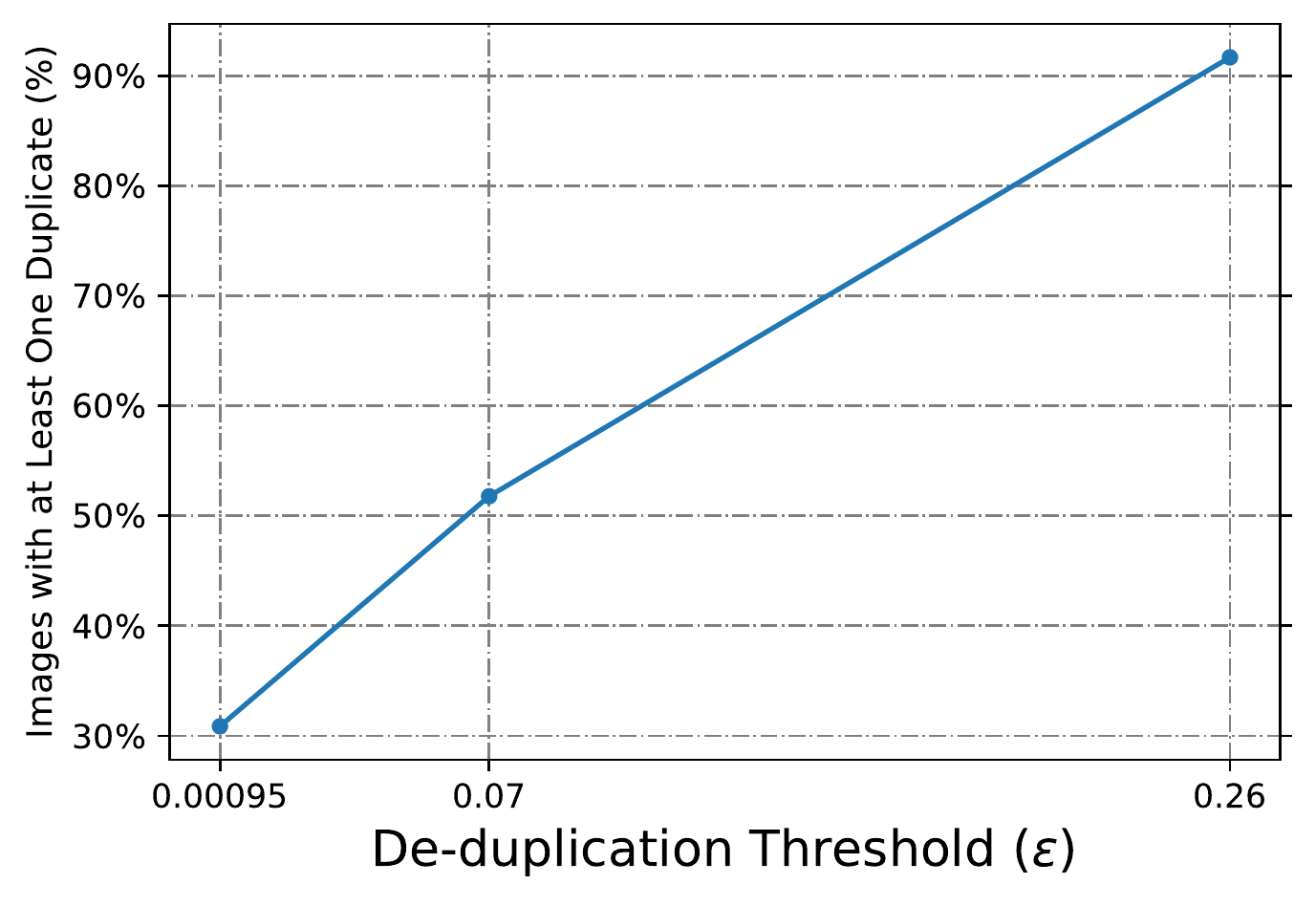}
\caption{}
\end{subfigure}
\begin{subfigure}{.32\columnwidth}
\includegraphics[width=\columnwidth]{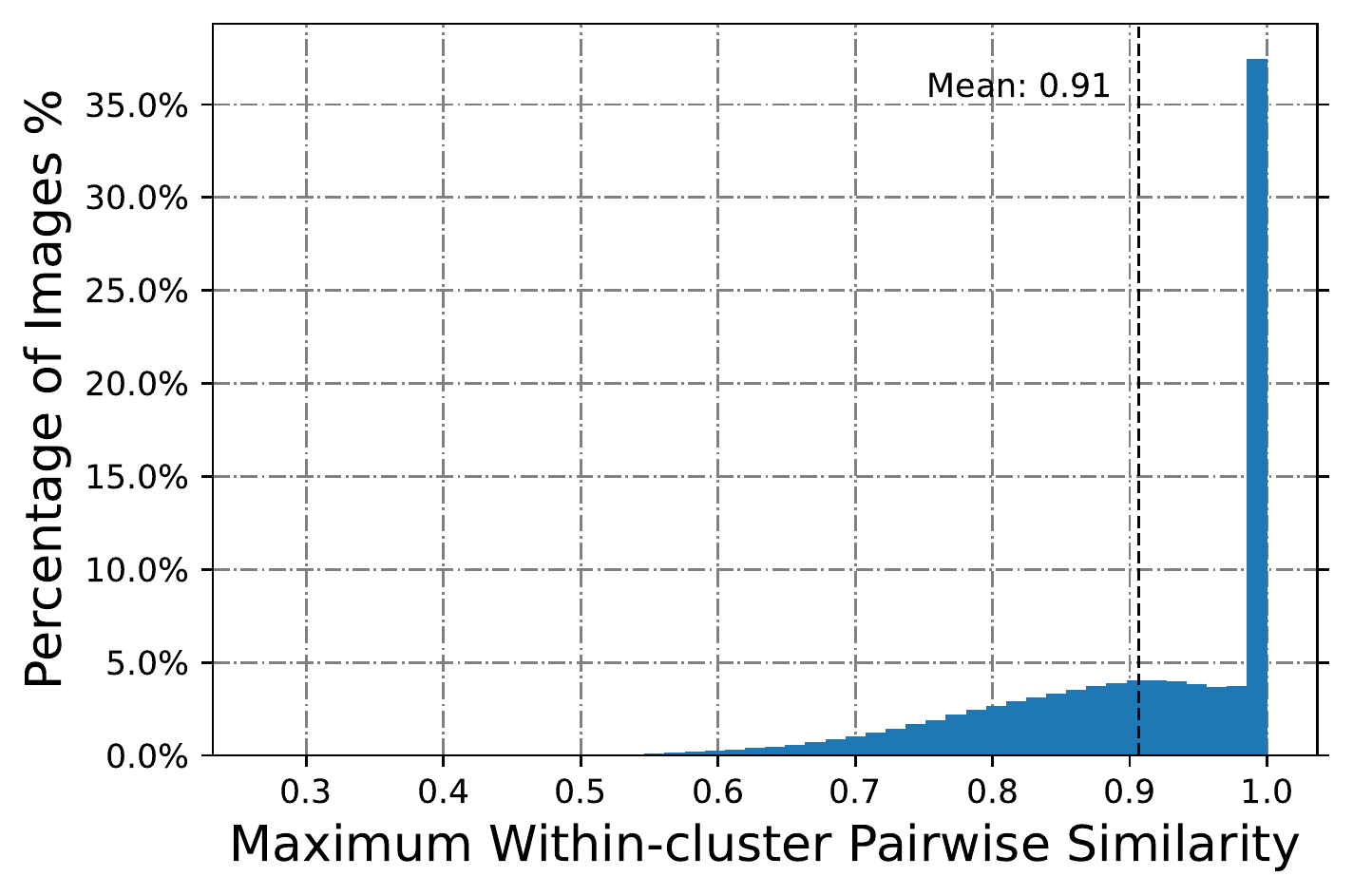}
\caption{}

\end{subfigure}

\caption{\textbf{Extreme semantic redundancy in LAION-440M.} (a) Fraction of data remaining as a function of deduplication threshold $\epsilon$ for LAION-440M. (b) Percentage of images in LAION-440M with at least one semantic duplicate as a function of $\epsilon$. (c) Histogram of the number of within-cluster image pairs in LAION-440M at a given cosine similarity.}
\label{fig:eps_vs_dataset_size}
\end{center}
\vskip -0.2in
\end{figure}

\section{SemDeDup on LAION} \label{sec:results}

If we consider pairs of data points to be semantic duplicates when their cosine similarity is at least $1-\epsilon$, then $\epsilon$ can be thought of as a deduplication dissimilarity threshold, with increasing $\epsilon$ reflecting an increasingly coarser notion of semantic equality. We expect that low thresholds of $\epsilon$ will find semantic duplicates, while higher thresholds will allow semantically redundant data pairs as well. 

To evaluate SemDeDup's ability to discover semantic redundancy in multi-modal data, we train CLIP models on the LAION dataset (Section \ref{sec:methods}). We first show that LAION contains extreme amounts of semantic redundancy (Section \ref{sec:res-quantification}) and provide examples of the semantic duplicates discovered by SemDeDup (Section \ref{sec:res-examples}). Most critically, we demonstrate that removing the semantic duplicates discovered by SemDeDup has minimal to no impact on converged performance and increases learning speed (Section \ref{sec:res-train-dedup}).

\subsection{Datasets and Training} \label{sec:methods_data}

\paragraph{The LAION dataset.} To train large-scale multi-modal models, we used the LAION dataset \citep{laion5b}, an open multi-modal dataset containing up to 5 billion image-text pairs scraped from the web. LAION data were filtered using a pre-trained CLIP model to only retain image-text pairs with an embedding similarity greater than 0.28. Image-text pairs containing very short captions or small images were also removed. A simple de-duplication method based on the image url was also performed. 

The majority of our experiments were performed on the LAION-440M filtered subset of LAION-2B introduced by \citep{radenovic2023filtering}. This dataset was filtered using a Complexity, Action, and Text (CAT) filtering according to three criteria: (1) high enough caption \textbf{complexity}; (2) the caption must contain an \textbf{action}; (3) any \textbf{text} present in the image cannot substantially overlap with the caption.

To ensure this CAT filtered LAION-440M subset did not impact our results, we also performed experiments on unfiltered data derived from LAION. Much of the original LAION-400M subset \citep{laion400m} is no longer available due to broken urls, so we used a reduced version of the LAION-400M subset containing the 233 million data points we were able to collect, which we call LAION-233M. 


\paragraph{CLIP training.} For CLIP training on LAION, we use the OpenCLIP implementation \citep{openclip}. We use CLIP-ViT-Base/16 in all our experiments. The model has Vision Transformer Base (ViT-B-16) \citep{vit} as an image encoder and Text Transformer \citep{vaswani2017attention} as a text encoder.
We train all models with a global batch size of 33k image-caption pairs and fix the number of training epochs to 32 regardless of the dataset size. This results in training for a fewer number of iterations when training on deduplicated data, thereby achieving efficiency gains. We train with AdamW \citep{adamw} and cosine learning rate schedule with warmup. The same peak learning rate of $ 5 \mathrm{x} 10^{-4}$ is used for all models. Table \ref{table:clip_training_parameters} shows training parameters we use for CLIP.\\


\paragraph{CLIP Evaluation} For CLIP evaluation we use zero-shot evaluation on 30 different datasets. Tables \ref{table:zeroshot_results_table} and \ref{table:out_of_distribution_results_table} in the Appendix list all the datasets we use for evaluation. 


\subsection{Extreme semantic redundancy at web-scale} \label{sec:res-quantification}

How many semantically redundant pairs are there in LAION? Remarkably, we find that even tiny thresholds $\epsilon$ lead SemDeDup to remove large fractions of data in LAION440M (Fig. \ref{fig:eps_vs_dataset_size}a), showing that LAION-440M contains large quantities of semantic duplicates. Surprisingly, $30\%$ of images in LAION-440M have a semantic duplicate at the highly stringent distance threshold of $\epsilon=0.00095$, while $50\%$ have a duplicate at the tight threshold of $\epsilon=0.03$ (Fig. \ref{fig:eps_vs_dataset_size}c). Moreover, a histogram of pairwise cosine similarity in LAION-440M (Fig. \ref{fig:eps_vs_dataset_size}d) reveals a high density of pairs at high cosine similarity, including a large contribution at $1$, reflecting highly similar semantic duplicates. These results demonstrate that LAION-440M contains large amounts of semantic redundancy.

\subsection{What do semantic duplicates look like?} \label{sec:res-examples}

What leads to semantic duplicates? In Fig. \ref{fig:visduplicates}, we show examples of semantic duplicates found at different thresholds $\epsilon$. At extremely low values of $\epsilon$ we find perceptual duplicates, and at slightly higher values of $\epsilon$, we find semantic duplicates, which are the same image but with distortions which evade exact de-duplication approaches such as different margins, crops, aspect ratios, and color filters, or slightly different peripheral details.  Fig. \ref{fig:before_and_after_1}, and \ref{fig:before_and_after_2} show examples of clusters that are semantically deduplicated at increasing levels of $\epsilon$, clearly indicating more semantic diversity in deduplicated clusters as $\epsilon$ increases. 

Many semantic duplicates are of products which may have been displayed on multiple e-commerce websites, each with a slightly different style. As a result, semantic duplicates often contain different, but highly similar captions.  While most clusters contained 20-40\% duplicates, there are several remarkable outliers in redundancy in LAION-440M (Fig. \ref{fig:cluster_size}), including one cluster containing $\approx 307,000$ copies of the European Union flag and another with $\approx 318,000$ copies of an icon of ``Image not found."

At higher levels of $\epsilon$ in Fig. \ref{fig:visduplicates}, and \ref{fig:semantically_similar_images_appendix},  we find fewer semantic duplicates, which are generally derived from the same source image, and more pairs which exhibit semantic redundancy instead, in which the same concept is present, but not derived from the same image source. For example, semantically redundant pairs may contain different images of similar objects or scenes. 

\subsection{Training on semantically deduplicated data improves efficiency} \label{sec:res-train-dedup}

If SemDeDup is effective at finding semantic duplicates, we should be able to remove these duplicates with a minimal performance impact. To test this, we train CLIP models on subsets of LAION-440M deduplicated at different thresholds $\epsilon$, corresponding to smaller fractions of data as $\epsilon$ rises. 

\begin{figure}[t]
\begin{center}

\begin{subfigure}{.48\columnwidth}
    \centering
\includegraphics[width =\columnwidth]{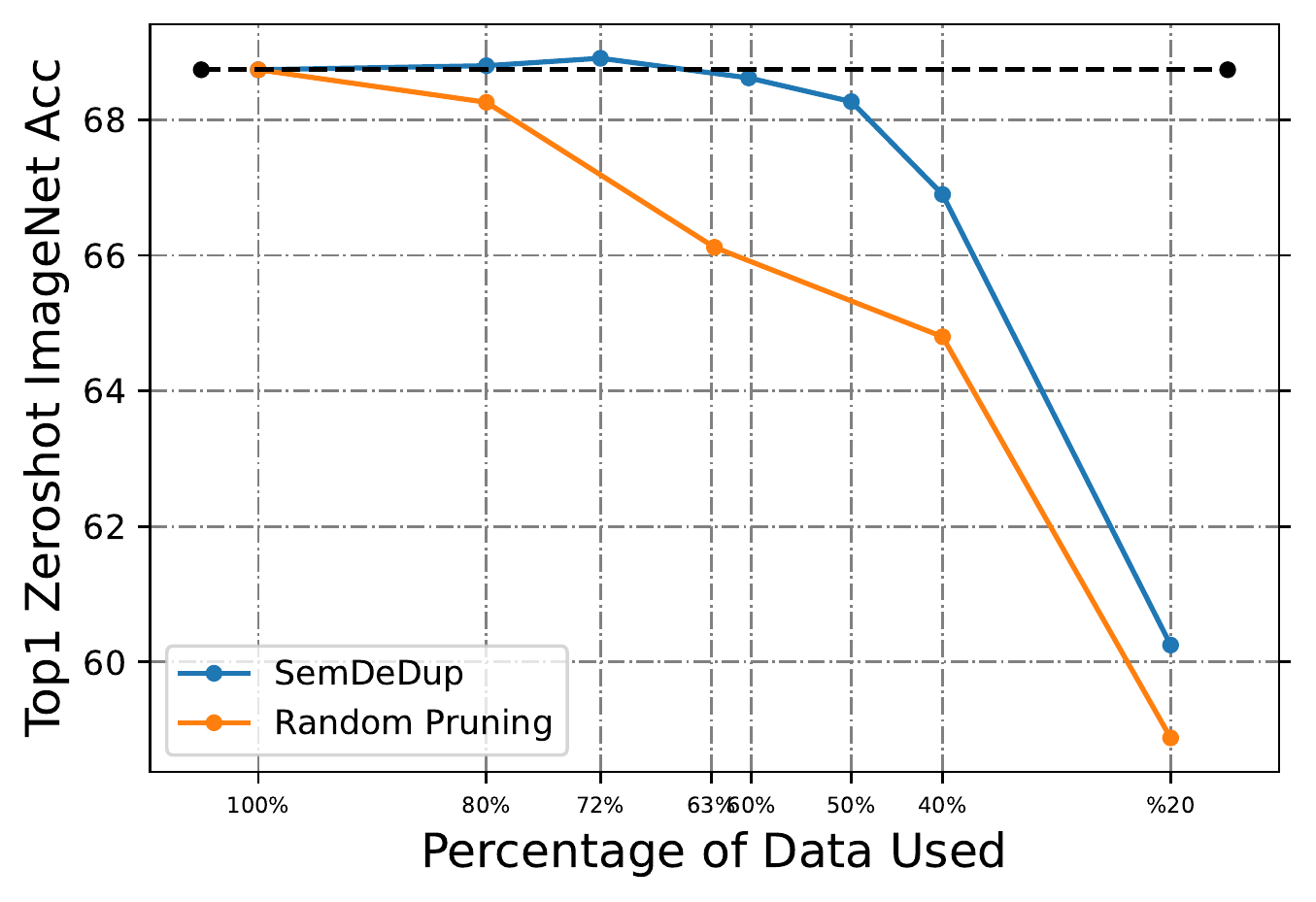}
    \caption{}
\end{subfigure}
\hfill
\begin{subfigure}{.48\columnwidth}
    \centering
\includegraphics[width = \columnwidth]{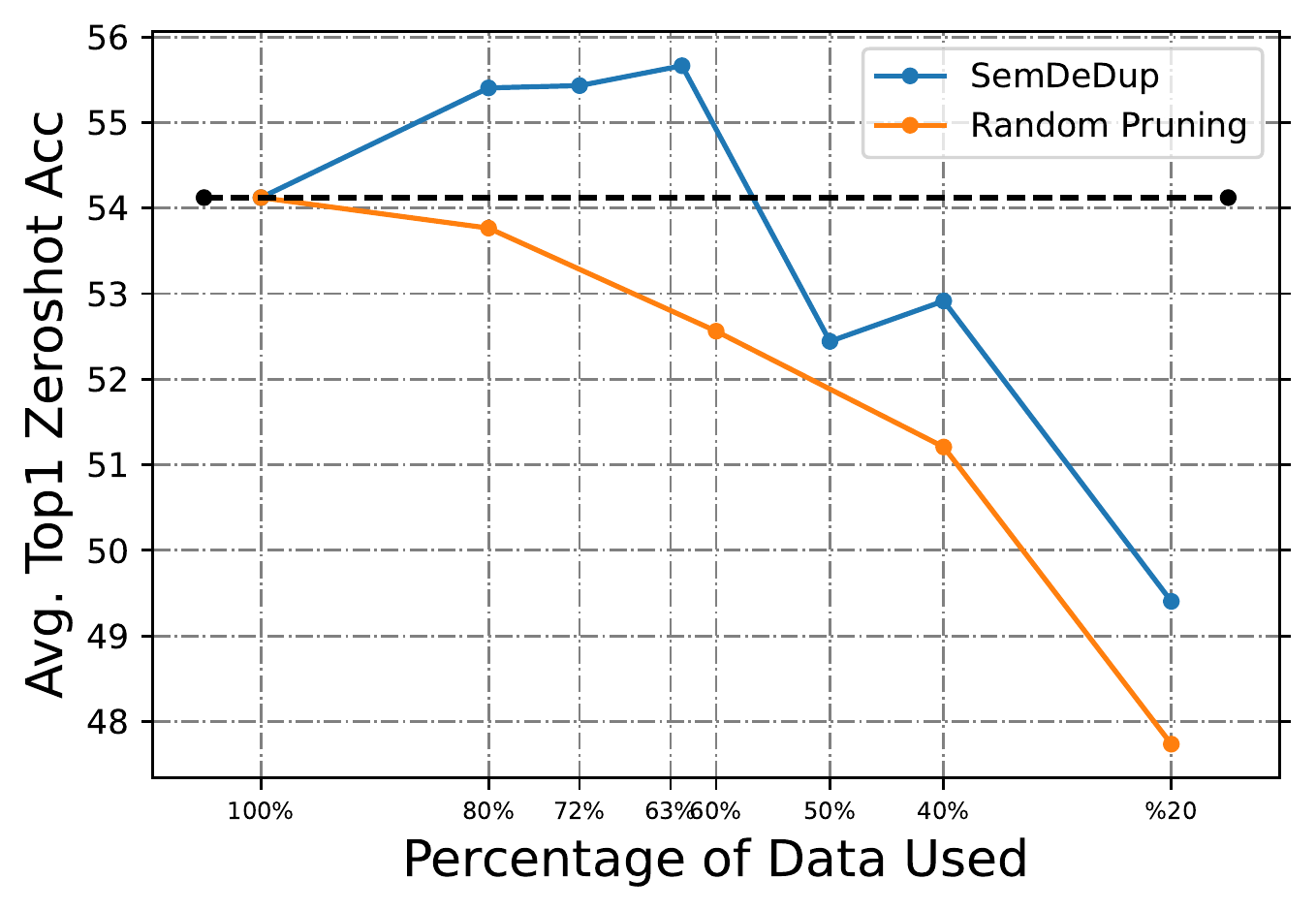}
    \caption{}
   \end{subfigure}
   
\caption{\textbf{SemDeDup allows better average zero-shot accuracy across 24 tasks with less data and faster pre-training.} (a): Performance of SemDeDup (blue) and random pruning (orange) for different amounts of retained data. Down to using only 50\% of LAION-440M for pre-training CLIP, we are able to match the zero-shot ImageNet accuracy of the baseline model trained on 100\% of the data (black dashed line) with a small drop of 0.47\% only, while we outperform the baseline model with only 63\% of data. (b): Average zero-shot performance for CLIP measured on 24 datasets. Average performance {\it improves} across $24$ tasks down to $63\%$ of the pre-training data, yielding {\it better} performance with almost $1.6\times$ {\it faster} pre-training.}
\label{fig:LAION-440_training}
\end{center}
\end{figure}

In Fig. \ref{fig:LAION-440_training} (a), we plot the top-1 zero-shot accuracy of our CLIP models on ImageNet-1k. Encouragingly, we found that SemDeDup can remove up to 37\% of LAION440M with no performance drop, and 50\% with minimal performance drop ($<0.5\%$). In contrast, randomly removing data results in much larger drops. In Fig. \ref{fig:LAION-440_training} (b), we show the average zero-shot performance across $24$ tasks, finding that on average, performance increased on de-duplicated data. See Table {\ref{table:zeroshot_results_table}} for detailed performance on all $24$ tasks at $6$ deduplication thresholds as well as $1$ baseline and $4$ random controls. See also Fig. \ref{fig:all_zershot_line_plots} for performance on $24$ individual tasks. 

\begin{figure}[ht]
\begin{center}

\begin{subfigure}{.48\columnwidth}
    \centering
\includegraphics[width=\columnwidth]{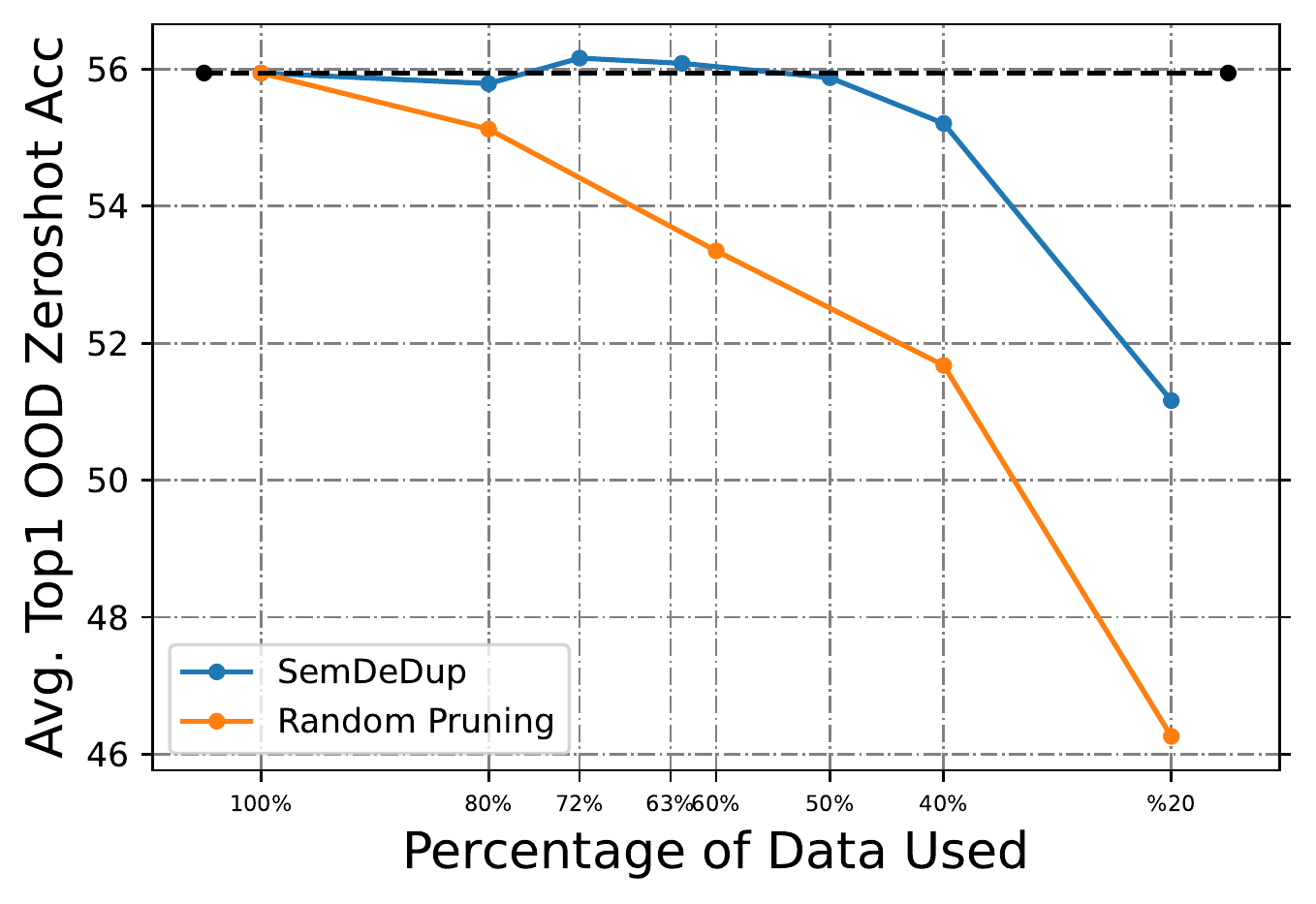}
    \caption{}
\end{subfigure}
\hfill
\begin{subfigure}{.48\columnwidth}
    \centering
\includegraphics[width=\columnwidth]{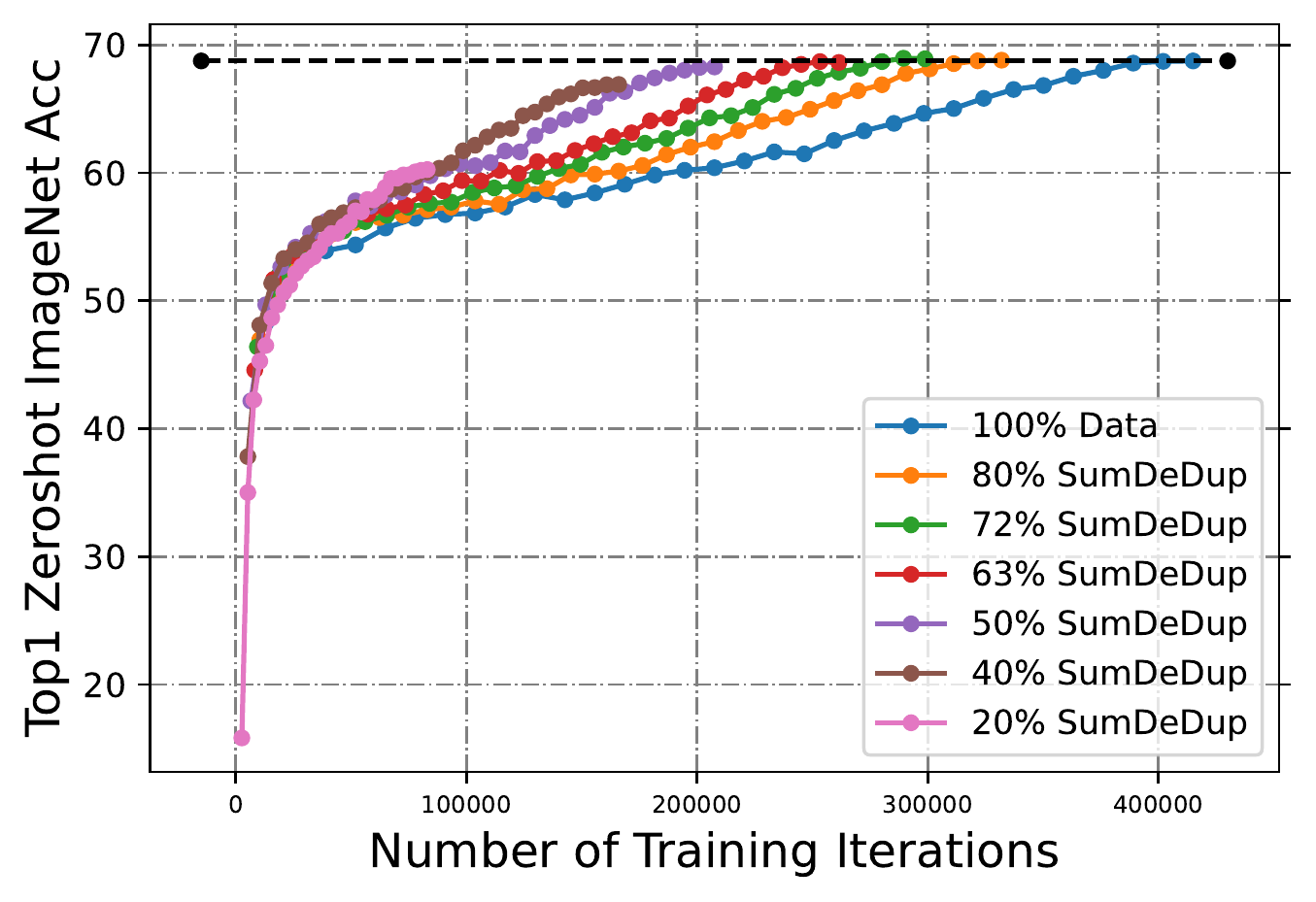}
    \caption{}
   \end{subfigure}
 
\caption{\textbf{SemDeDup allows better average performance across 6 ImageNet OOD tasks with less data and faster pre-training.} \textbf{(a)} zeroshot validation accuracy averaged over 6 ImageNet-1k OOD tasks for CLIP models pre-trained on deduplicated LAION data with different thresholds $\epsilon$. We outperform the baseline model with only 63\% of pre-training data from LAION-440M.
\textbf{(b)} We track zeroshot ImageNet-1K performance as a function of LAION-440M pre-training iterations at different deduplication thresholds. The models trained on smaller deduplicated datasets actually learn faster, thereby allowing them to converge to almost baseline performance (black dashed line) in far fewer iterations.}
\label{fig:laion440_acc_vs_iterations}
\end{center}
\end{figure}

We also evaluated out-of-distribution robustness on $6$ datasets commonly used for this task: ImageNet-A, ImageNet-O \citep{hendrycks2021nae}, Imagenet-R \citep{hendrycks2021many}, Imagenet-sketch \citep{imagenet-sketch}, ImageNetV2 \citep{imagenet-v2}, and ObjectNet \citep{objectnet}. We again found that SemDeDup {\it increased} average performance over baseline when {\it removing} 37\% of the data, and matched performance when 50\% was removed as shown in Fig. \ref{fig:laion440_acc_vs_iterations} (a). See Table \ref{table:out_of_distribution_results_table}  for detailed performance on $6$ OOD tasks at $6$ deduplication thresholds as well as $1$ baseline and $4$ random controls.  We also note that SemDeDup outperforms random pruning on all individual out-of-distribution robustness  datasets for all fractions of dataset kept. See Fig. \ref{fig:fig:all_OOD_zershot_line_plots} for performance on the $6$ individual tasks. 

\begin{figure}[ht]
\begin{center}
\includegraphics[width = 0.4\columnwidth]{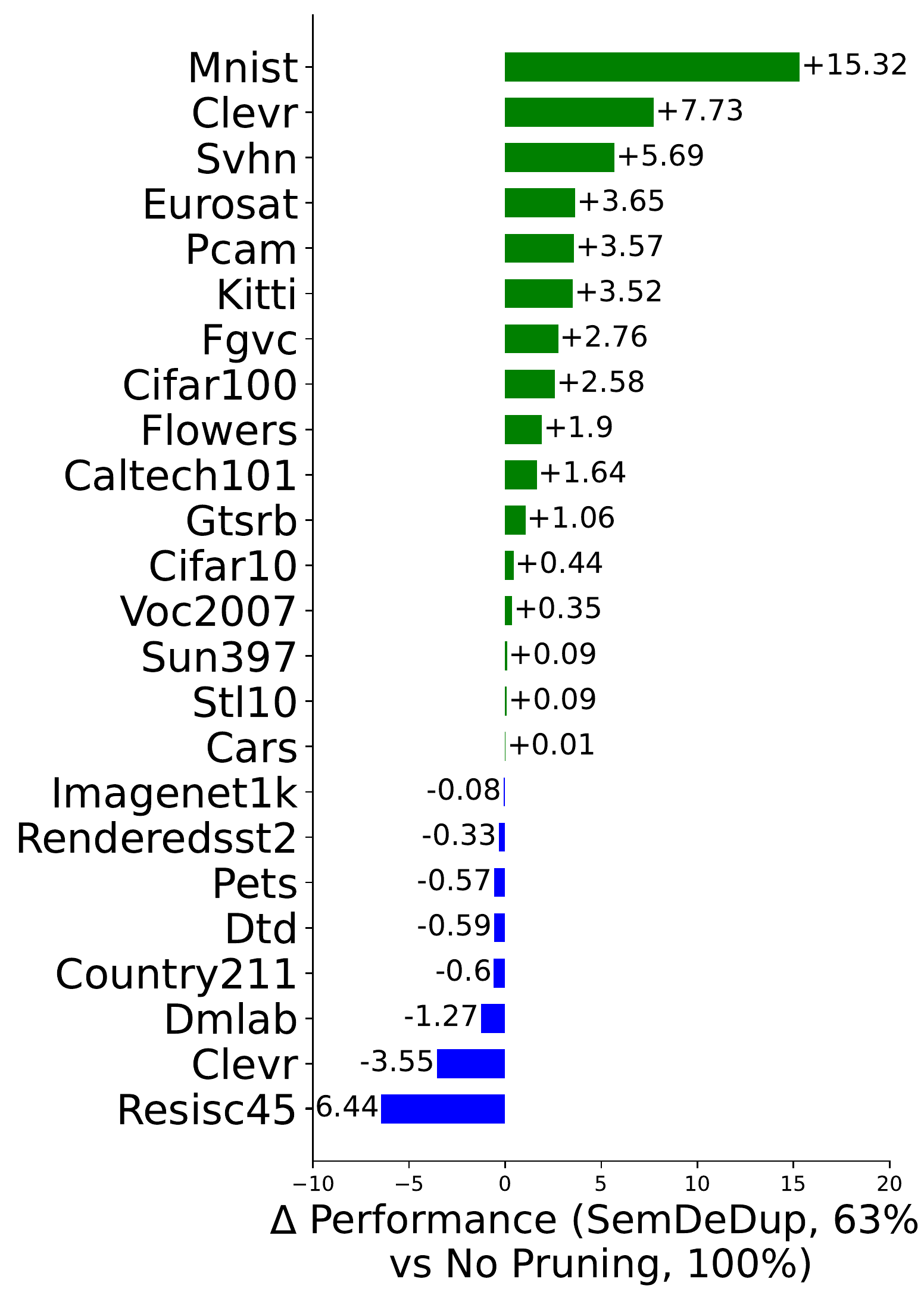}
\includegraphics[width = 0.4\columnwidth]{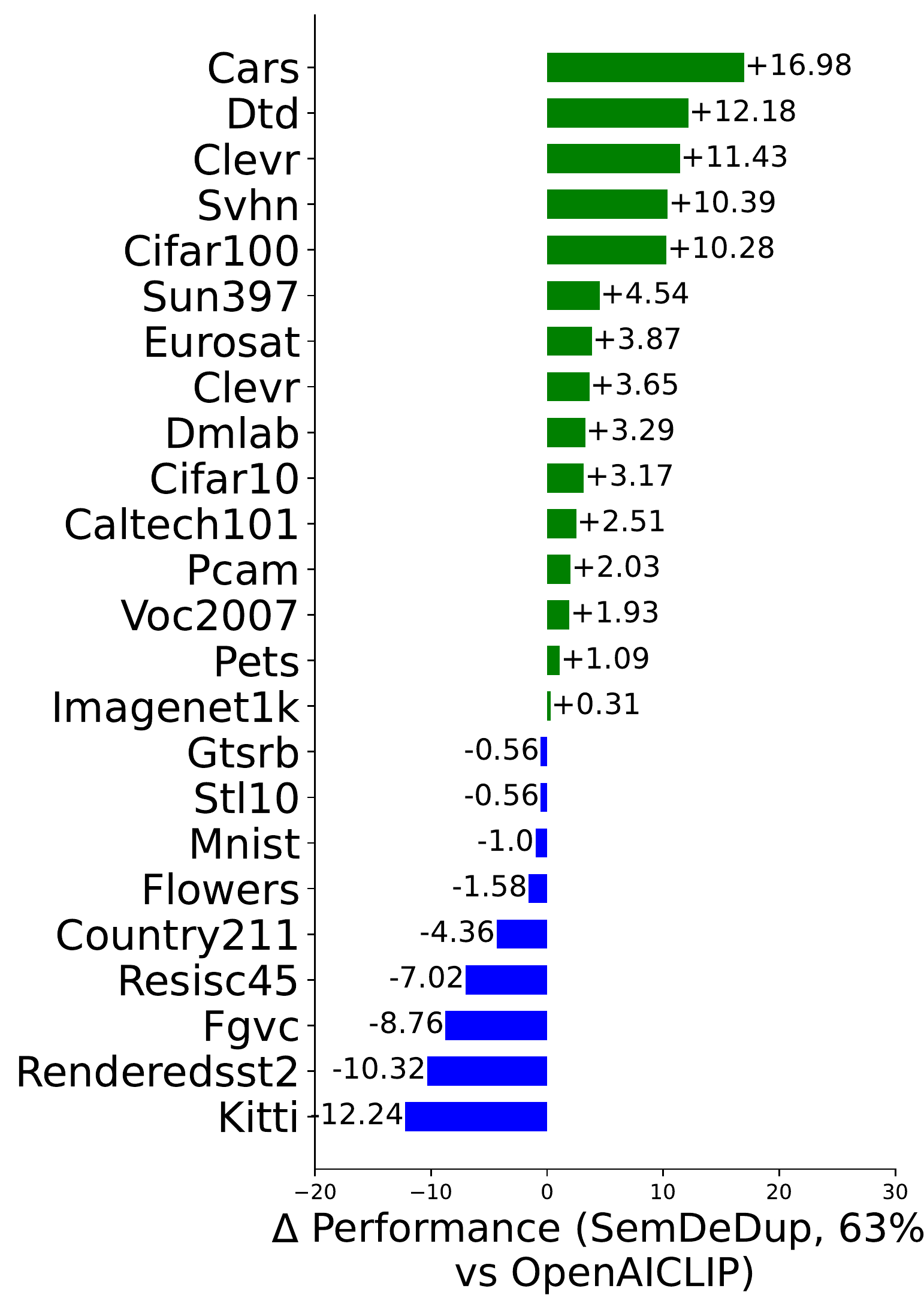}

\caption{\textbf{SemDeDup improves zeroshot and OOD performance in many tasks with less pre-training.} A comparison of zeroshot evaluation performance between our CLIP model trained on 63\% of LAION-440M after de-duplication to a baseline CLIP model trained on 100\% of the data (\textbf{left}), and OpenAI CLIP \citep{clip}  (\textbf{right}) on $30$ tasks. The green bars show when SemDeDup outperforms the baseline model. }
\label{fig:zeroshot_bar_plot}
\end{center}
\end{figure}

Fig \ref{fig:zeroshot_bar_plot} shows SemDeDup performance across $30$ combined zero-shot and OOD tasks when removing $37\%$ of the data, relative to a CLIP baseline trained on all the data. Remarkably, on about $20$ out of $30$ tasks, performance actually {\it improves} after {\it removing} pre-training data, whereas on all but about $3$ of the remaining tasks performance is not substantially reduced.  Our observation that SemDeDup can improve performance in many cases is consistent with prior work which has found that removing duplicates may improve performance by discouraging memorization \citep{lee2021deduplicating}.

We emphasize that SemDeDup achieves these results on LAION-440M, an already highly curated dataset derived from LAION-2B which was found to have similar performance despite the almost five-fold reduction in data \citep{radenovic2023filtering}. However, to ensure that this curated subset did not bias our results, we also evaluated on LAION-233M, an uncurated subset of LAION-2B, finding qualitatively similar results (Fig. \ref{fig:laion233m_result}).

Because SemDeDup reduces the number of training points, it enables substantially faster training. In Fig. \ref{fig:laion440_acc_vs_iterations} (b), we plot the top-1 zero-shot accuracy on ImageNet-1k as a function of the number of iterations for different deduplication thresholds $\epsilon$. Notably, models trained on deduplicated data reach convergence in substantially fewer iterations. 

Why do models trained on uncurated data exhibit slower learning? We posit that successive learning iterations involving semantic duplicates yield redundant information, thereby wasting valuable computation on data points that are highly similar to those the model has already seen. By removing these semantic duplicates, we increase the fraction of data points which provide a marginal information gain to the model, thereby increasing learning speed \citep{Sorscher2022-wo}. 


\section{SemDeDup on Natural Language} \label{sec:nlp}


\subsection{Methods}
\label{sec:langmethods}



 We train language models on deduplicated versions of the C4 dataset \citep{ColinT5}. Since pre-training large language models on the entire C4 corpus is beyond our compute budget, we train on subsets of this data whose sizes are compute optimal given model size as per \citep{HoffmannChinchilla}. We use the OPT model and training configurations \citep{ZhangOPT} to train 125M and 1.3B parameter models (see Table 1 in \citep{ZhangOPT} for full specifications). We use the original number of warmup updates but adjust the learning rate schedule such that all training runs anneal learning rate to 0 by the end of the training — this allows for fair comparisons of model performances across different dataset sizes. 
 For 1.3B model size experiments, we increase the number of warmup updates to 5550 and reduce the peak learning rate to $6 \mathrm{x} 10^{-5}$ to stabilize training. 

We evaluate our trained language models on two independent validation sets: the validation text corpora used by OPT \citep{ZhangOPT}  (referred to as "opt\_valid") and a random sample of the instruction finetuning corpus used to train the OPT-IML family of models \citep{IyerOPTIML}, composed of verbalized prompts corresponding to a wide range of NLP tasks and their solutions (referred to as "prompts\_with\_answers").

To perform SemDeDup, we pass documents through the open-sourced pre-trained 125M OPT model \citep{ZhangOPT} and save the last layer embedding for the last token in the document. We then apply the same method described in Section~\ref{sec:methods_clustering} with $\mathcal{K} = 11000$ to cluster these embeddings. We compare to random pruning and the NearDup method described in~\citep{lee2021deduplicating}. Note that the deduplication threshold values associated with different fractions of data remaining change compared to LAION-440M, as seen in Fig.~\ref{fig:nlp_appendix_125M_eps_vs_frac_data}.


\subsection{Results on Language Modeling} \label{sec:results_c4}

\begin{figure}[ht]
\begin{center}
\includegraphics[width = 0.4\textwidth]{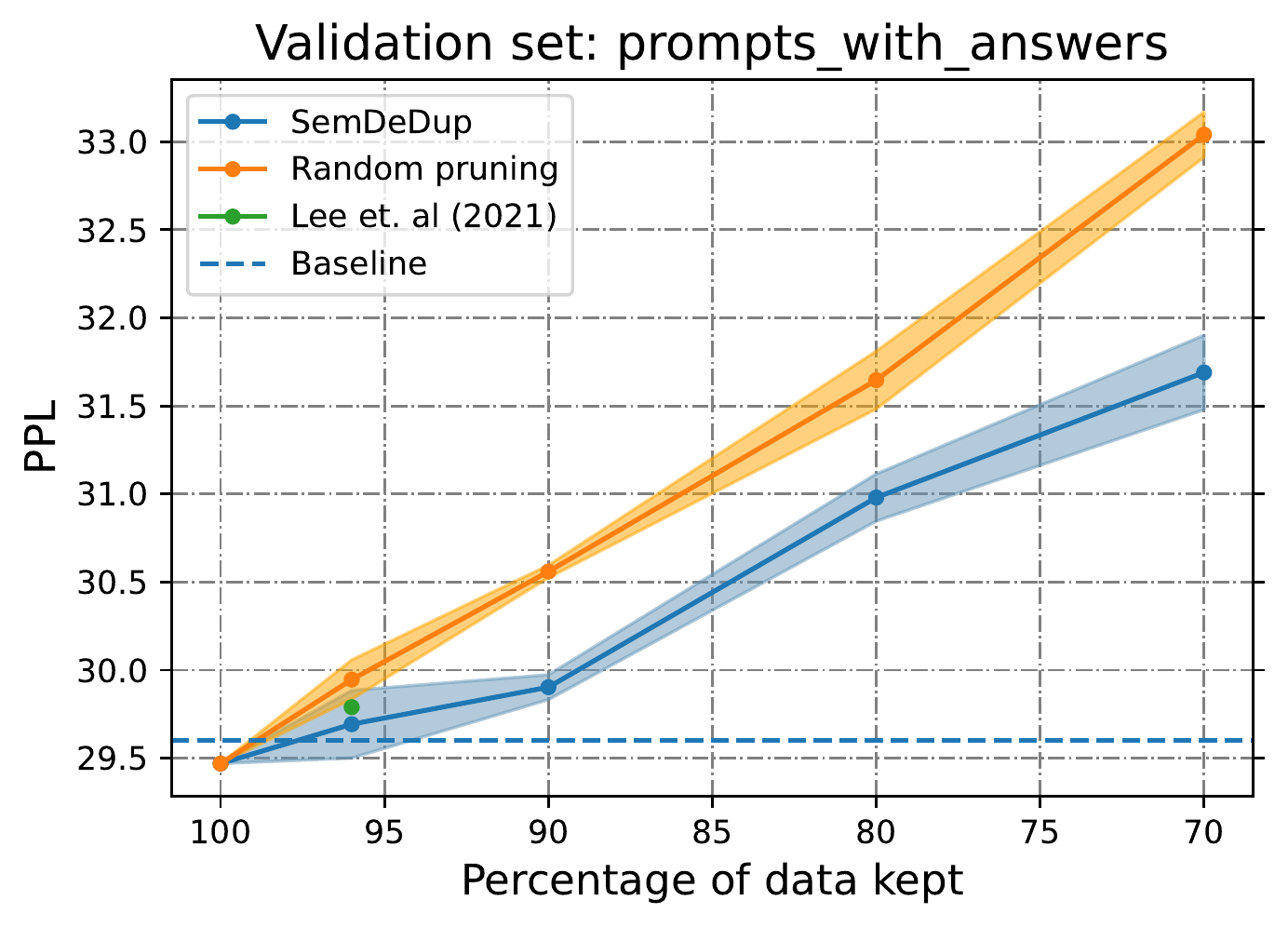}
\includegraphics[width = 0.4\textwidth]{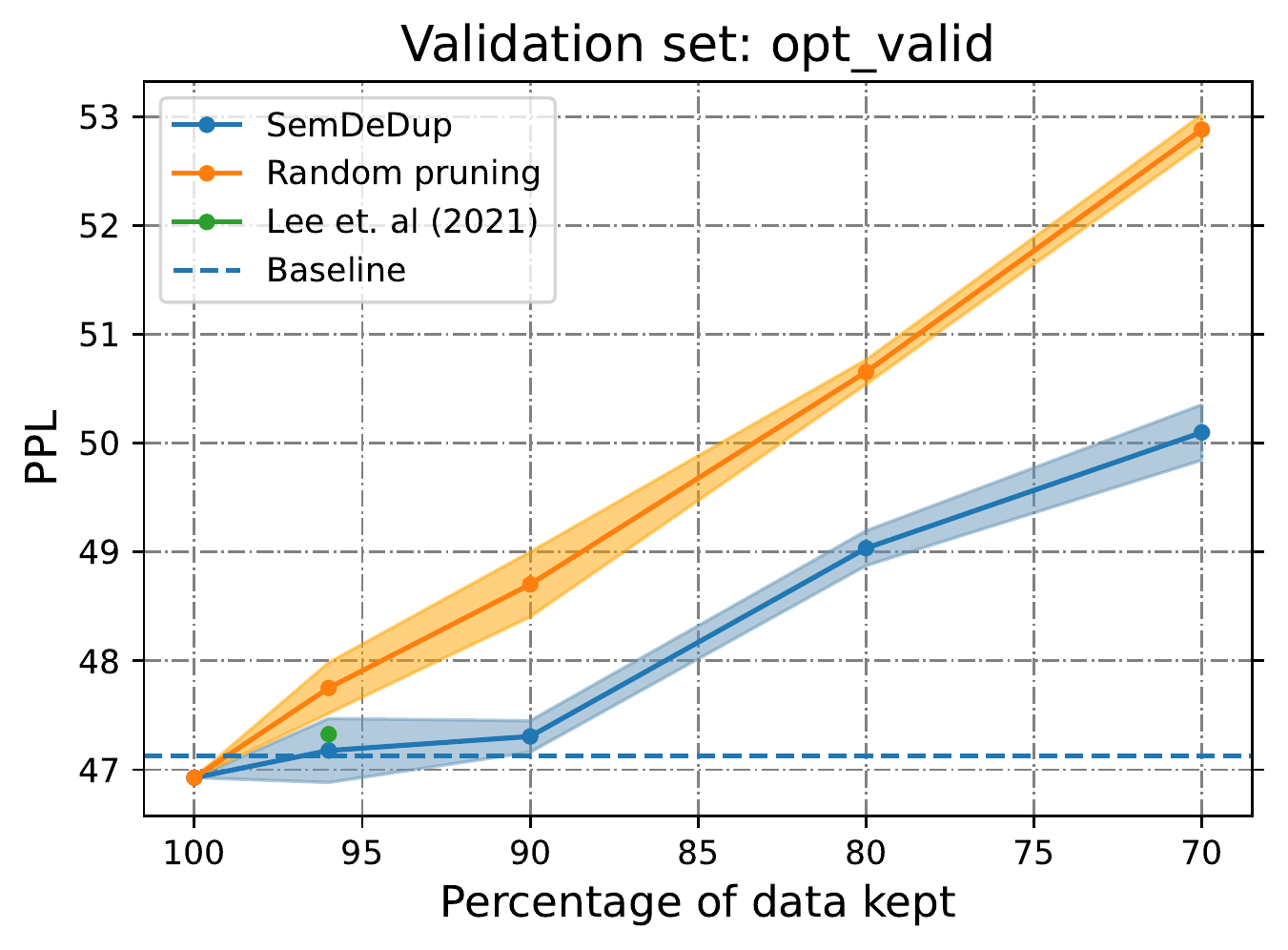}

\caption{\textbf{SemDeDup applied to C4}.  The x-axis corresponds to different percents of data kept, and the y-axis represents the perplexity on validation sets described in Section~\ref{sec:langmethods} (lower is better). Each point is a separate 125M model trained on one-pass of its respective pruned dataset (mean and standard deviation across 3 random training seeds). The green point represents a 125M model trained on a version of C4 deduplicated via the NearDup method \citep{LeeDedup}. Note that NearDup (the single green point) keeps 96.1\% of the data. SemDeDup can match this baseline performance while keeping only 80\% of the data (see Table~\ref{table:nlp_appendix_125m_c4_matched_epoch_80} for numerical comparison).}
\label{fig:nlp_125M_matched_epoch_c4}
\end{center}
\end{figure}

In Fig.~\ref{fig:nlp_125M_matched_epoch_c4}, we show the performance of SemDeDup versus random pruning. We observe that SemDeDup significantly outperforms random pruning as measured by perplexity on prompts\_with\_answers and average opt\_valid performance. For a breakdown of performance on individual validation sets in opt\_valid, see Fig.~\ref{fig:nlp_appendix_big_ood_plot_matched_epochs} where we observe that SemDeDup beats random pruning on every single validation set in opt\_valid.

\begin{figure}[ht]
\begin{center}
\includegraphics[width = 0.4\textwidth]{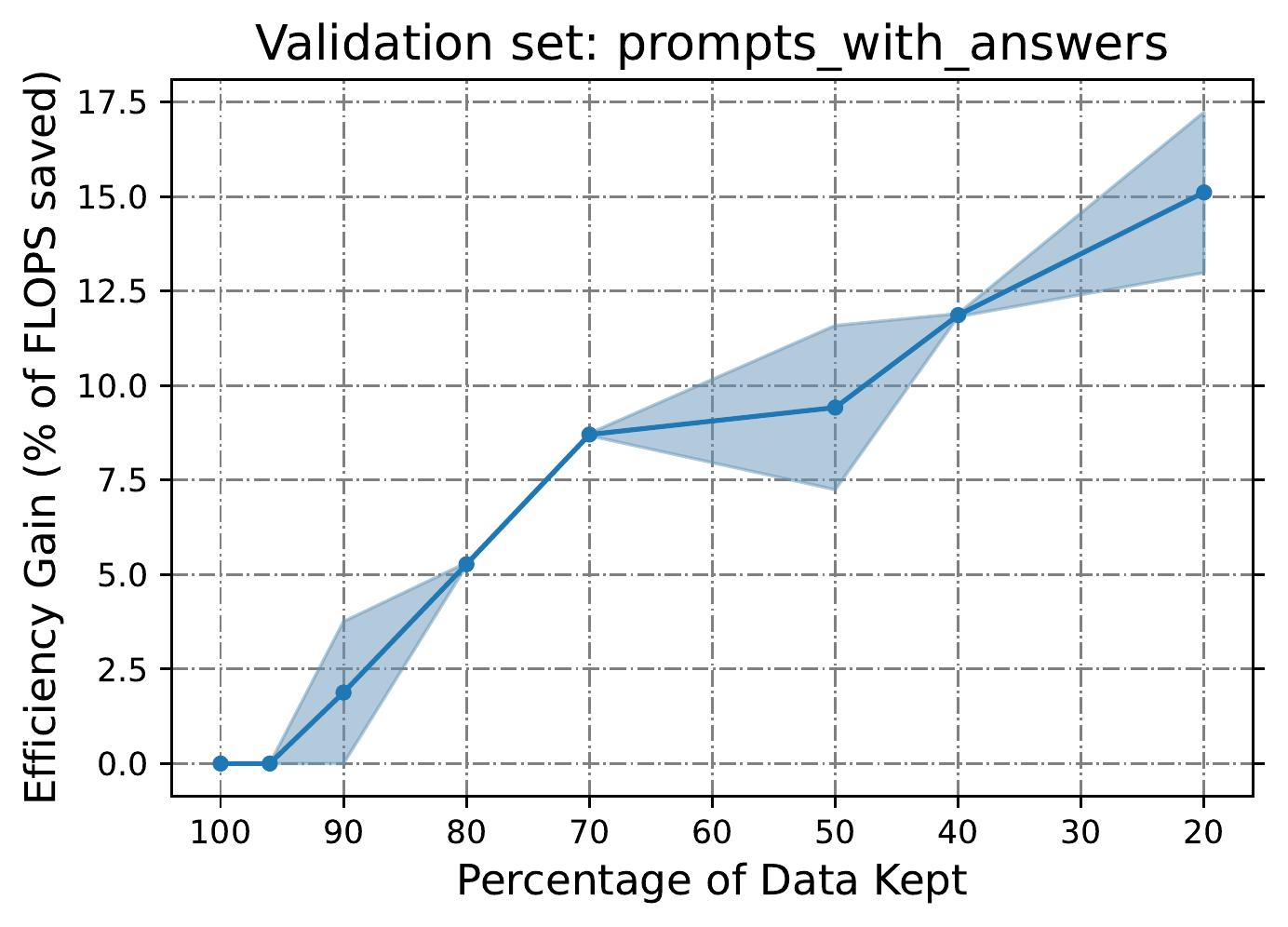}
\includegraphics[width = 0.4\textwidth]{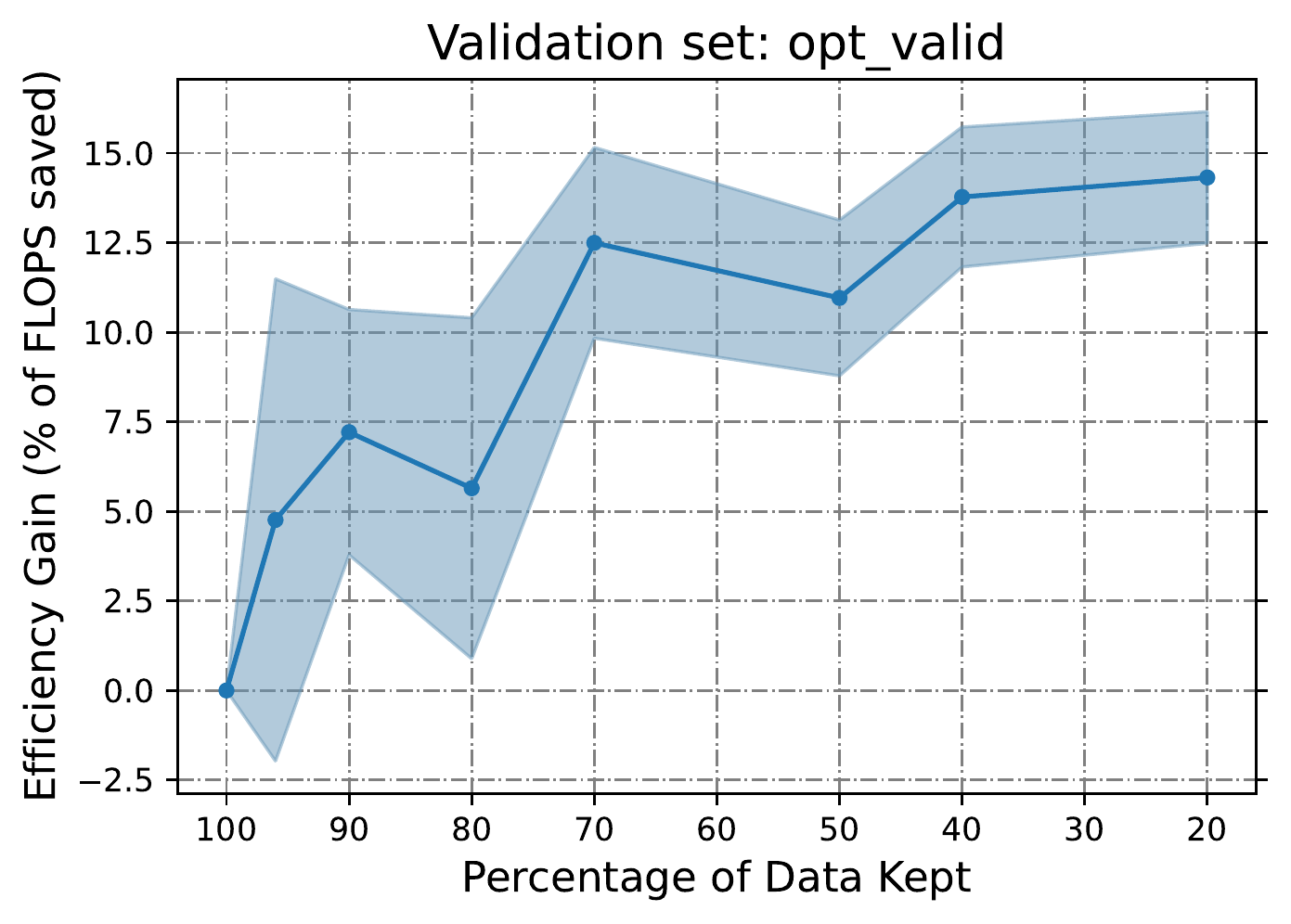}

\caption{\textbf{SemDeDup allows compute efficiency gains by training on much smaller datasets for slightly longer.} We prune datasets via SemDeDup and continue training past one epoch until we reach baseline model perplexity. The x-axis is the percentage of data kept, and the y-axis is the percentage of FLOPs saved. For example, training on the 80\% pruned dataset reaches baseline model perplexity on prompts\_with\_answer in ~95.0\% of the baseline training, saving ~5.0\% compute. Mean and standard deviation provided across 3 random training seeds.}
\label{fig:nlp_125M_efficiency_graphs}
\end{center}
\end{figure}

Training on less data for one epoch naturally causes performance to decrease. Thus, we also explore whether continuing to train on the same smaller pruned datasets for more epochs will match the performance of a baseline model trained on a larger dataset. In Fig.~\ref{fig:nlp_125M_efficiency_graphs}, we train on datasets pruned with SemDeDup, but perform the same number of total training steps as the baseline model on the larger dataset (which was trained for $1$ epoch).  This causes the model to do multiple epochs over the pruned dataset. We observe that by training for multiple epochs over significantly pruned datasets we can reach the performance of a single-epoch run on the full dataset using 10-15\% less compute.  This is similar to the finding in Section~\ref{sec:res-train-dedup}. Notably, this efficiency gain is larger at higher pruning percentages, indicating that more aggressive pruning can yield more efficiency gains. This trend generally holds across the individual validation sets in opt\_valid (see Fig.~\ref{fig:nlp_appendix_big_ood_plot_efficiency}).

On the C4 validation set, we observe that SemDeDup still outperforms random pruning in Fig.~\ref{fig:nlp_appendix_125m_full_plots_matched_epoch}. In Table~\ref{table:nlp_appendix_125m_c4_matched_epoch} we compare SemDeDup to the NearDup baseline from \citep{LeeDedup}. We observe that NearDup and SemDeDup have comparable performance as is expected, because with 4\% pruning there is very little change to the underlying dataset.

\subsection{What is being pruned in language data?}

In Fig.~\ref{table:nlp_appendix_examples_one} and Fig.~\ref{table:nlp_appendix_examples_two} we choose specific clusters and show a random sample of documents retained in the cluster after performing SemDeDup for different values of $\epsilon$. In Fig.~\ref{table:nlp_appendix_examples_one}, we observe that at low values of $\epsilon$, we find semantic duplicates in the form of templated text, where typically few words (e.g. a geographic location or a name) is changed. This successfully evades exact-string deduplication methods but contains highly redundant information as seen in Fig.~\ref{table:nlp_appendix_examples_one}. In Fig.~\ref{table:nlp_appendix_examples_two}, we show an example of a cluster with semantically redundant duplicates — most examples in this cluster are advertisements about Nike shoes. These examples are not necessarily templated text or have exact string matches, but are highly redundant nonetheless. We see in Fig.~\ref{table:nlp_appendix_examples_two} that at more aggressive pruning (i.e. higher $\epsilon$) these semantically redundant duplicates get pruned. We note that exact string duplicates (i.e.``perceptual duplicates for text") are rare since duplicate occurrences of any three-sentence spans were removed in C4 already.


\section{Analysis of hyperparameter choices} \label{sec:more_analysis}
\subsection{Number of k-means clusters for SemDeDup}
Here  we study the impact of changing the number of clusters $k$ in the k-means clustering step in SemDeDup described in section \ref{sec:methods_id_dup}. In all our experiments in the main paper, we set $k$ = 50,000 for the LAION dataset and $k$ = 11,000 for the C4 dataset. To study the impact of the $k$ on the performance, we deduplicate LAION440M using different values for $k$ and train different CLIP models on the deduplicated data. We compare three values for $k$ (70,000, 50,000, and 10,000) when deduplicating LAION440M to 40\% of its size. As we see in Table \ref{table:effect_of_number_of_clusters} the exact choice of $k$ has a very small impact on performance as measured by the zeroshot accuracy on ImageNet with a small improvement in the top1 accuracy as $k$ increases. 

The key intuition is that the choice of $k$ implements a tradeoff in the probability of recovering all semantic duplicates of any data point, and the computational complexity of doing so. For example, assuming k-means finds equal cluster sizes, each data point will lie in a cluster of size $N/k$, and we are only searching for $\epsilon$-nearest neighbors (with cosine similarity > $1-\epsilon$) within each cluster.  As $k$ decreases, cluster size $N/k$ increases, and the error probability of substantially many $\epsilon$ nearest neighbors of a data point lying outside it's own cluster decreases, while the computational complexity of searching for all nearest neighbors within the cluster increases. As long as $k$ is small enough relative to the total dataset size $N$, so that $N/k$ is large enough to contain most nearest neighbors of each data point, the performance of SemDeDup should be robust to the choice of $k$.

\begin{table}
\centering
\caption[add short caption]{Performance of CLIP when keeping 40\% of LAION440M as a function of the number of k-means clusters $k$ used for SemDeDup. SemDeDup is robust to the choice of $k$ and the impact on the zeroshot accuracy on ImageNet is small with slight performance improvement as we increase $k$.}
\label{table:effect_of_number_of_clusters}
\begin{tabular}{C{4.cm}|C{2.cm}|C{2.cm}|C{2.cm}}
\toprule
{Metric / Num. of Clusters} &  70K Clusters &  50K Clusters &  10K Clusters \\ 
\midrule
Top1 zeroshot IN Acc.     &        67.11  &            66.90   &         66.56    \\  
Top5 zeroshot IN Acc.    &        90.96  &            90.74   &         91.04    \\  

\bottomrule
\end{tabular}
\end{table}

\subsection{Pre-trained models for extracting embeddings}
As we describe in section \ref{sec:methods_id_dup}, SemDeDup clusters the example embeddings extracted from a pre-trained foundation model and uses them for deduplication. To study the effect of the pre-training dataset of the foundation model on SemDeDup we deduplicate LAION440M using an OpenAI CLIP model \citep{clip} pre-trained on a different dataset than LAION. We use the Open AI CLIP ViT-Base model pre-trained on a private dataset of 400 million image-caption pairs. We use the embeddings from this model to deduplicate LAION440M dataset to 40\% of its size. As we see in Table \ref{table:using_different_foundation_model}, using Open AI CLIP model for extracting embeddings has a negligible impact on the performance.
\begingroup
\begin{table}
\centering
\caption[add short caption]{The impact of the foundation model used for extracting embeddings. Using a foundation model pre-trained on a different (and private) dataset has no impact on the performance. The table shows the performance when training OpenCLIP on 40\% of LAION440M dataset. In each column in the table, the dataset is deduplicated by SemDeDup using embeddings from a different model.}
\label{table:using_different_foundation_model}
\begin{tabular}{C{7.5cm}|C{2cm}|C{3cm}}
\toprule
{Metric / Model  Used for Extracting Embeddings} &  CLIP Pre-trained on LAION440M &  OpenAI CLIP \citep{clip} Pre-trained on Private 400M dataset \\ 
\midrule
Top1 zeroshot IN Acc. After Training on DeDup Data     &        66.90  &            66.96     \\  
Top5 zeroshot IN Acc. After Training on DeDup Data    &        90.74  &            90.80   \\  

\bottomrule
\end{tabular}
\end{table}
\endgroup

\subsection{Different strategies for choosing which semantic duplicates to keep}
In section \ref{sec:methods_id_dup} and Algorithm \ref{alg:semdedup}, we describe the steps for deduplication with SemDeDup. From each group of duplicates (the circles in Figure \ref{fig:introfig}), we keep the example with the lowest cosine similarity to the cluster centroid in the embedding space. This is the default setting for all experiments we run unless otherwise mentioned. In Table \ref{table:which_example_to_keep} we study the strategy we follow to choose the example to keep from each group of duplicates. We train three CLIP models on 40\% of LAION440M deduplicated by SemDeDup for 32 epochs. We try three options for choosing the examples we keep 1) keeping examples with low similarity to centroids, 2) keeping random examples, and 3) keeping examples with high similarity to cluster centroids. We obverse that the difference between the three methods in zero-shot accuracy on ImageNet is negligible.
\begingroup
\begin{table}
\centering
\caption[add short caption]{ Different strategies to choose the example to keep from each group of duplicates.}
\label{table:which_example_to_keep}
\begin{tabular}{C{3.5cm}|C{3.3cm}|C{2.cm}|C{3.3cm}}
\toprule
{Metric / Examples to Keep} &  Examples with low similarity to centroids & Random examples & Examples with high similarity to centroids \\ 
\midrule
Top1 zeroshot IN Acc.     &        66.90  &            66.90   &         66.73    \\  
Top5 zeroshot IN Acc.    &        90.74  &            90.95   &         90.82    \\  

\bottomrule
\end{tabular}
\end{table}
\endgroup

\subsection{Training on deduplicated data for more iterations improves performance}
Training on deduplicated data comes with the advantage that we train for fewer iterations under the match-epochs setting. For example, training on 50\% of LAION440M for the same number of epochs as the baseline model (100\% of the data) means that we train for only 50\% of the number of training iterations. We find that we can achieve a good trade-off between performance and training speed when training on deduplicated data.  We show that training on deduplicated LAION440M for more iterations improves the accuracy while still being below the number of iterations we train the baseline model for. In Table \ref{table:training_for_more_iterations}, we show results for different CLIP models, trained on 50\% of LAION440M, for a different number of training iterations. We see that by continuing training the model until we reach 75\% of the iterations relative to the baseline model, we outperform the baseline model on not only ImageNet, but also on average accuracy over 24 datasets, and on the $6$ out-of-distribution datasets. 

\begingroup 
\setlength{\tabcolsep}{2.5pt} 
\renewcommand{\arraystretch}{1.5} 
\begin{table}
\centering
\caption[add short caption]{By training on only 50\% of LAION440M, deduplicated using SemDeDup, we perform better than training on whole LAION440M (baseline100) with 62.5\% or 75\% of the number of training iterations used for training the baseline model. The table shows zeroshot Top1 accuracy.}
\label{table:training_for_more_iterations}
\begin{tabular}{p{3.5cm}|p{1.cm}|p{1.8cm}|p{1.6cm}}
\toprule
{Model} &   IN Acc &  Avg. Acc (24 datasets) &  Avg. OOD (6 datasets)\\ 
\midrule
100\% data, 100\% iters (Baseline100) &  68.74  &      54.12   &         55.94    \\  
50\% data, 50\% iters     &        68.27  &            \textbf{54.59}   &         55.87\\ 
50\% data, 62.5\% iters   &         68.33  &    \textbf{55.07}  & \textbf{56.38}  \\
\rowcolor{LightCyan}
50\% data, 75\% iters   &  \textbf{69.21}  &  \textbf{55.07}  & \textbf{56.36}   \\
\bottomrule
\end{tabular}
\end{table}
\endgroup

\subsection{Choosing the deduplication threshold $\epsilon$}
 We tune the deduplication threshold $\epsilon$ for each dataset manually to get the desired deduplicated dataset size. To do that, we first run the clustering step of SemDeDup. Then we sample 10\% of the clusters and tune $\epsilon$ on them. We found that using only 10\% of clusters gives a good approximation of the final dataset size. 
 We  notice that the  relationship between $\epsilon$ and the deduplicated dataset size is semi-linear for both LAION and C4 datasets  (see Fig. \ref{fig:eps_vs_dataset_size}, \ref{fig:eps_for_different_num_clusters}, and \ref{fig:nlp_appendix_125M_eps_vs_frac_data}). When tuning  $\epsilon$ we start with two values and run SemDeDup on 10\% of the clusters (the time needed for this step is a few minutes. See the DeDup. Time column in Table \ref{table:semdedup_time} ).  Then we  linearly interpolate the two values of $\epsilon$  knowing their correspondence dataset size and the target dataset size to get a better value for $\epsilon$. In Fig.  \ref{fig:eps_for_different_num_clusters} we plot the duplicated dataset size as a function of  $\epsilon$ for  different values of  the number of clusters $k$  used. We show that $k$ has a small impact on the value $\epsilon$ only when the duplicated dataset size is less than 50\%.

\section{Compute cost of running SemDeDup}
We report in Table \ref{table:gpu_hours} the cost of running SemDeup on LAION440M in GPU hours. We see in the table that the overhead of deduplicating LAION440M doesn't exceed 1\% of the training cost in GPU hours. This results in substantial savings in the overall cost after deduplication. For example, training on 50\% of the data saves 50\% of the training cost while requiring only 1\% of the training cost for deduplication. 
We also show in Table \ref{table:semdedup_time} the time needed for deduplicating LAION440M dataset using SemDeDup using 8 GPUs for clustering and 64 GPUs for CLIP training. Our implementation for SemDeDup parallelizes the operations across devices to speed up the deduplication.  The table also shows how the time changes as we change the number of clusters. 

However, we should note that the computational cost of SemDeDup can be amortized across the efficiency gains it can generate in training many downstream models by many other groups.  For example, its typical use case would be to take a large web-scaled dataset, and semantically deduplicate it {\it once}, resulting in a much smaller {\it foundation dataset} \citep{Sorscher2022-wo} that can be widely disseminated to the community.  Then many different groups can train many different foundation models on this deduplicated foundation dataset, and all these groups will reap the training efficiency gains conferred by a less redundant smaller dataset. Thus the computational cost of finding the dataset can be amortized across the efficiency gains achieved on many downstream training runs, in direct analogy to how the computational cost of training a foundation model can be amortized across the computational efficiency gains with which it achieves high zero-shot or fine-tuning performance on many  downstream applications.   

 \begingroup
\setlength{\tabcolsep}{3.5pt} 
\renewcommand{\arraystretch}{1.4} 
\begin{table}
\centering
\caption[add short caption]{SemDeDup requires much fewer GPU hours than training a CLIP ViT-B-16 model on LAION440M for one epoch. When using 50K clusters, it requires only 0.29 of the GPU hours needed for one epoch of training on 100\% of the data. This is equivalent to 0.0091 of the complete training cost in GPU hours.}
\label{table:gpu_hours}
\begin{tabular}{C{2.5cm}|C{2.cm}|C{2.cm}|C{1.7cm}|C{2.cm}|C{2.cm}}
\toprule
{Num. SemDeDup Clusters / Cost } & GPU Hours For Training CLIP on 100\% of LAION440M for 32 Epochs & GPU Hours For Training on 50\% of DeDup. Data & SemDeDup Overhead in GPU Hours & SemDeDup Overhead / 1 Epoch Training GPU Hours & SemDeDup Overhead / 32 Epochs Training GPU Hours \\ 

\midrule
10K Clusters & 11541  &  5770.5  & 163.5 & 0.43 & 0.0132\\
\hline                                         
25K Clusters &  11541 &   5770.5 & 101.2  & 0.26 & 0.0082\\
\hline
\rowcolor{LightCyan}
50K Clusters &  11541 &   5770.5  &  110.3  & 0.29 & 0.0091\\
\hline
70K Clusters &  11541 &   5770.5 & 103.6 & 0.27  & 0.0084\\
\bottomrule
\end{tabular}
\end{table}
\endgroup

\section{Discussion} \label{sec:discussion}

We introduced SemDeDup, a simple yet tractable and effective method which leverages pre-trained embeddings to remove semantic duplicates which are highly semantically similar but not identical. Removing semantic duplicates improves learning speed and out-of-distribution performance while providing efficiency gains of up to 50\% on the largely uncurated LAION and 15\% on the partially curated C4. SemDeDup demonstrates the importance of data quality and the potential of data curation to dramatically improve training efficiency. 

\paragraph{\textbf{Limitations.}} While SemDeDup does an effective job of removing semantic duplicates and some semantically redundant data points, it is only one way to remove uninformative data points. In particular, this work does not capture many aspects of semantic redundancy, nor does it address removal of bad or misleading data, all of which can likely be exploited to make substantial further reductions to dataset size without sacrificing performance.  

SemDeDup also requires access to a pre-trained embedding model relevant to the domain of interest, which may pose a problem for entirely novel domains unrelated to the wide array of publicly available pre-trained models. However, for most domains, pre-trained models are readily available, and many such models have been shown to generalize to related domains. We, therefore, expect that this limitation will only apply to a small fraction of the practical use cases for SemDeDup.

In LAION, we identified semantic duplicates based only on image data, but we ignored the caption information. Leveraging this information may lead to the identification of further semantic duplicates. 

Our results on C4 showcase the potential of SemDeDup for NLP, but the gains were more modest due to the partially curated nature of C4 which has fewer duplicates than LAION. We also trained small models relative to the best models. It is possible that results may change with scale, though following \citep{Sorscher2022-wo}, it is likely increasing scale would further improve the benefits of data curation. 

Overall, the optimal data pruning policy for finding the smallest possible data subset under computational tractability and performance constraints remains, as ever, an extremely difficult open question. However, the remarkable  efficacy of SemDeDup, especially given its underlying simplicity and scalability, suggests that the removal of semantic duplicates may well be an important prerequisite for any more sophisticated data pruning algorithm, especially when working with modern, large, highly uncurated, web-scale datasets.  

\section*{Acknowledgements}
We thank Mido Assran and Mansheej Paul for discussions. We also thank Mitchell Wortsman for support with OpenCLIP. We also thank Armen Aghajanyan for suggestions for handling training instability in our language model experiments.

\bibliography{references}

\newpage
\appendix
\renewcommand{\thefigure}{A\arabic{figure}}
\setcounter{figure}{0}
\renewcommand\thetable{A\arabic{table}}
\setcounter{table}{0}
\onecolumn

\section{Additional Analysis} \label{sec:analysis_appendix}
\subsection{Number of k-means Clusters for SemDeDup}
\begin{figure}[ht]
\vskip 0.2in
\begin{center}
\centerline{
 \includegraphics[width = 0.5\textwidth]{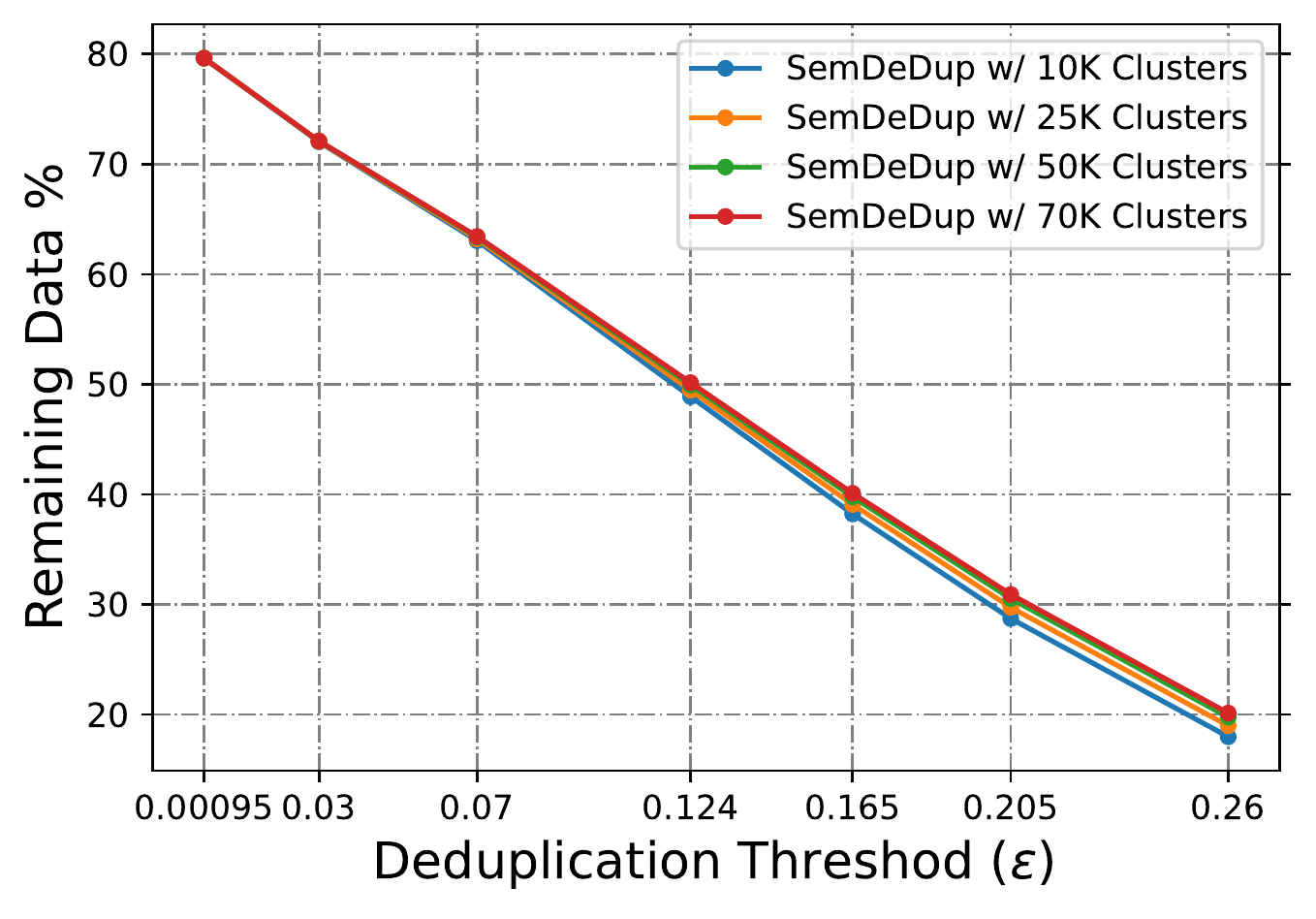}
}
\caption{Deduplicated dataset size as a function of the deduplication threshold for different values of k-means clusters $k$. Note that the range in the dataset size is 0.003\% when $\epsilon$ is 0.00095 and 2\% when $\epsilon$ is 0.26.}
\label{fig:eps_for_different_num_clusters}
\end{center}
\vskip -0.2in
\end{figure}

\begin{figure}[ht]
\vskip 0.2in
\begin{center}
\centerline{
  \includegraphics[width = 0.30\textwidth]{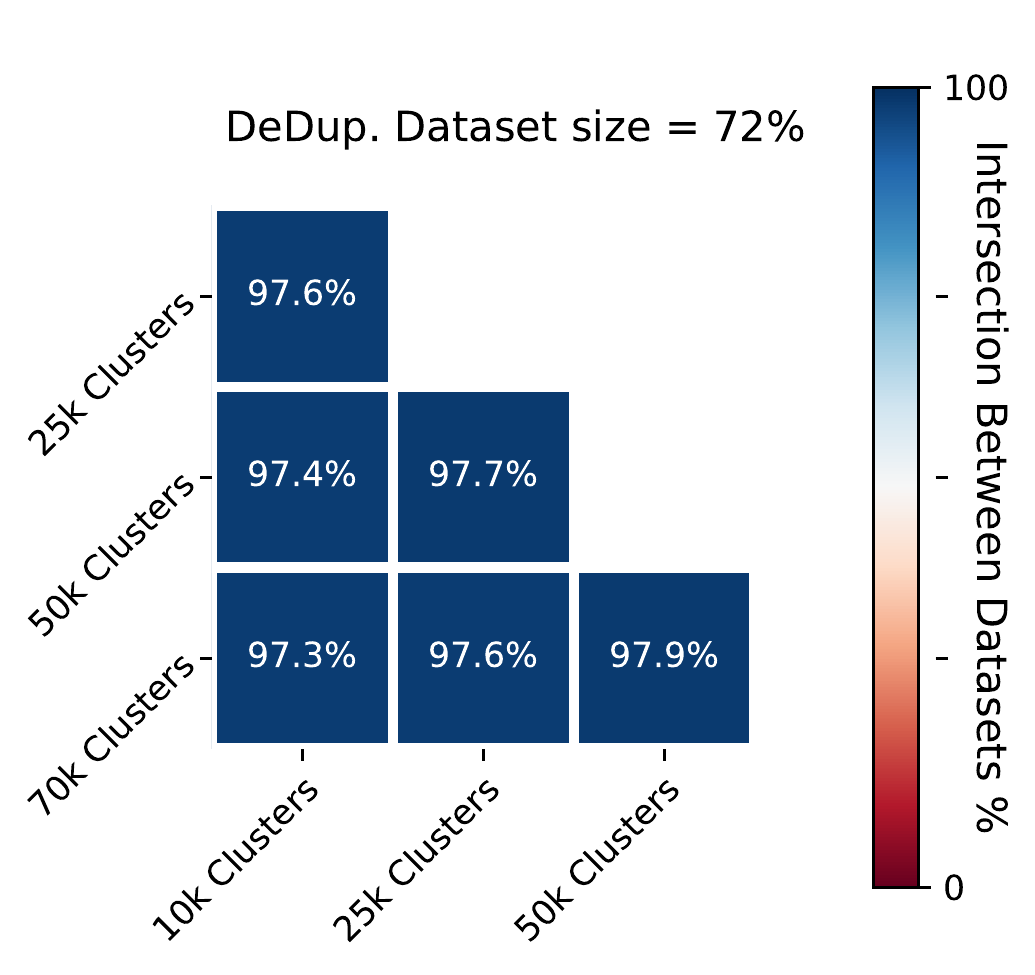}
  \includegraphics[width = 0.30\textwidth]{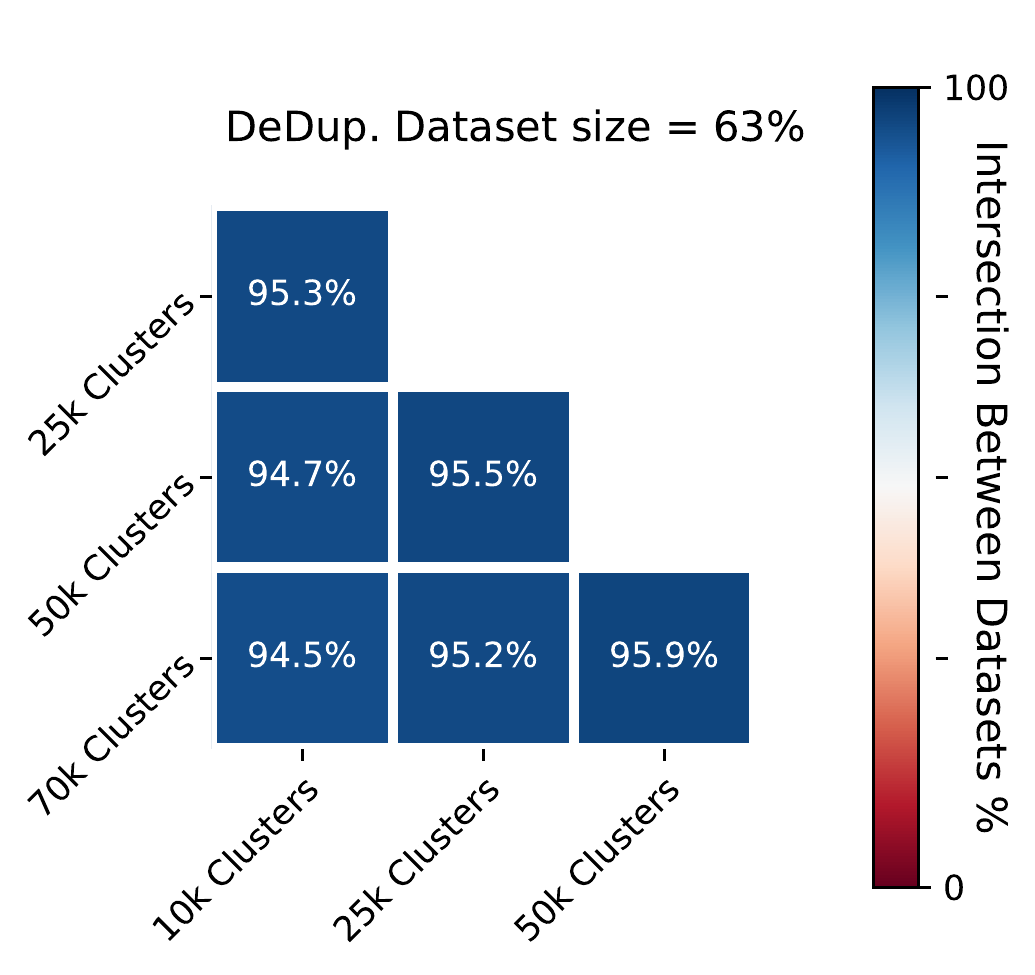}
 \includegraphics[width = 0.30\textwidth]{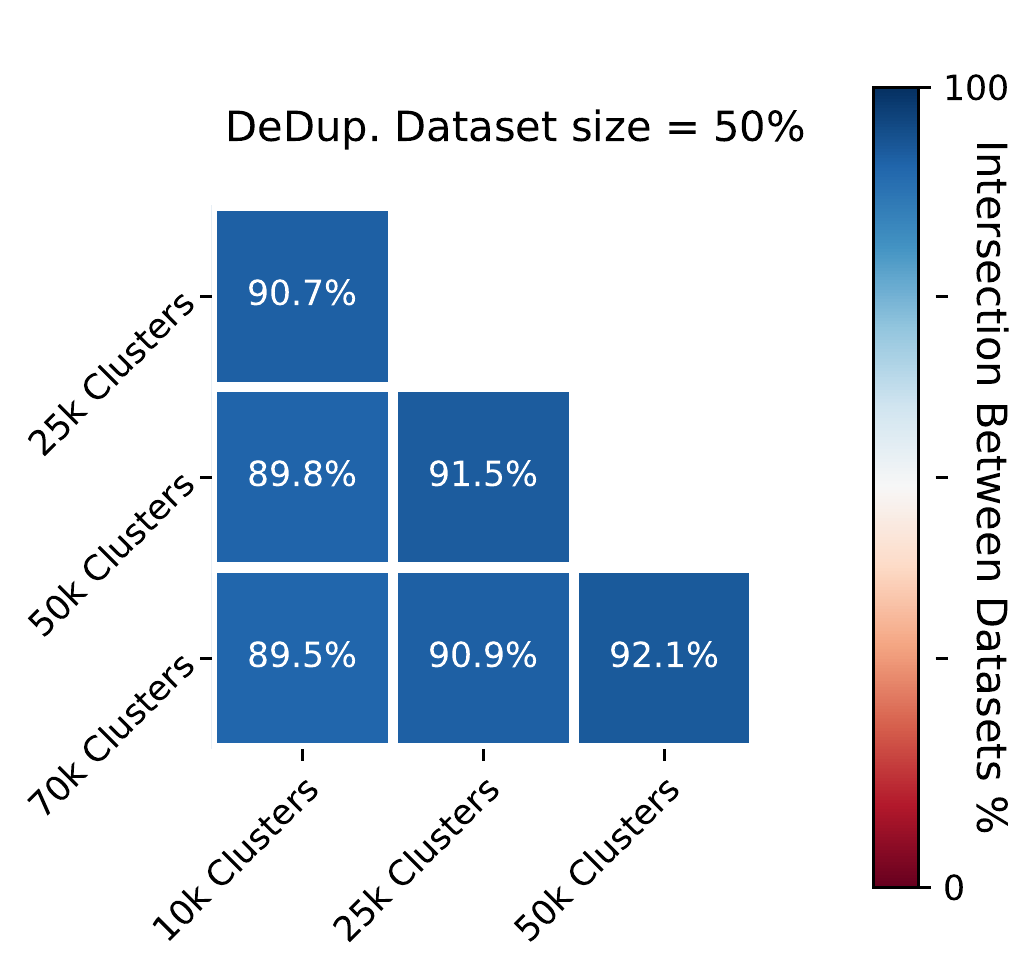}

}

\caption{Intersection between different deduplicated LAION datasets using different values for the number of k-means clusters $k$. Each cell corresponds to the percentage of intersection between two datasets deduplicated using different $k$ values. At the 72\% dataset size, more than 97\% of data examples are shared between all the datasets regardless of the value of $k$. This shows the robustness of SemDeDup to the number of clusters parameter $k$.}
\label{fig:dataset_intersection}
\end{center}
\vskip -0.2in
\end{figure}

To further assess the impact of changing  the value of $k$ we measure  the intersection between datasets deduplicated by SemDeDup using different  values  for $k$. Let $D_{A}=\{a_{1}, a_{2}, ..., a_{N}\}$ and $D_{B}=\{b_{1}, b_{2}, ..., b_{N}\}$ be two datasets of the same size $N$. We define the percentage of intersection $I$ between $D_{A}$ and $D_{B}$ in equation \ref{eq:intersection} as the percentage of data points that appear in both datasets relative to the dataset size $N$. Note that $I(D_{A}, D_{A})=100\%$.\\
We find that deduplicating LAION440M dataset to 72\% of its size using any value of $k$ values (10000, 25000, 50000, 70000) results in almost the same dataset with only 3\% of the examples replaced when changing $k$. This is induced by the 97\%  percentage of intersection $I$ value between any pair of datasets deduplicated using two different values for $k$. We show in Fig. \ref{fig:dataset_intersection} the percentage of intersection ratio between different datasets when changing the number of clusters $k$ at different deduplication thresholds $\epsilon$.\\
We also show in figure \ref{fig:eps_for_different_num_clusters} that by using the same deduplication threshold  value $\epsilon$ we get almost the same deduplicated dataset size for different values for $k$.\\



\begin{equation} \label{eq:intersection}
I(D_{A}, D_{B}) = 100 * \frac{\lvert D_{A} \cap D_{B} \rvert}{N}
\end{equation} 

\subsection{Estimating The Fraction of Duplicates Detected By SemDeDup}
SemDeDup searches for duplicates within clusters. This results in reducing the floating point operations (FLOPs) required for deduplication by 5 order of magnitude for LAION440M dataset as described in section \ref{sec:methods_clustering}. Indeed, by searching for duplicates within clusters, we ignore duplicates across different clusters if they exist. Here we try to estimate the efficiency of SemDeDup in detecting all the duplicates in the dataset.\\ 
Let $D_{\epsilon}$ represent the total number of duplicates in the dataset at a specific value of deduplication threshold $\epsilon$, and $D^s_{\epsilon}$ represent the total number of duplicates detected by SemDeDup. We define the deduplication efficiency $\eta_{\epsilon}$ (eq. \ref{eq:deduplication_efficiency}) as the fraction of duplicates detected by SemDeDup from the total number of duplicates in the datasets at a specific value of $\epsilon$. For example, a deduplication efficiency of 100\% corresponds to detecting all the duplicates in a dataset. As computing the exact value of $D_{\epsilon}$ is computationally expensive, we approximate its value by the number of duplicates between the cluster items and its 20 nearest neighbor clusters and donate this approximated value by $D^{'}_{\epsilon}$. We sampled part (2000 clusters) of the LAION440M dataset randomly and compute the value of the deduplication efficiency $\eta$ in eq. \ref{eq:deduplication_efficiency} for different values of $\epsilon$ and k-means clusters $k$. As we see in Table \ref{table:deduplication_efficiency}, for $k$=50,000, SemDeDup can effectively detect more than 94\% of the duplicates when keeping 63\% of LAION440M dataset and 89\% of the duplicates when keeping 40\%.

\begin{equation} \label{eq:deduplication_efficiency}
\eta = 100 * \frac{D^{s}_{\epsilon}}{D^{'}_{\epsilon}}
\end{equation} 

\begin{table}
\centering
\caption[add short caption]{Percentage of duplicates detected ($\eta$) by SemDeDup at different deduplication thresholds ($\epsilon$). We notice that $\eta$ increases as we reduce the number of clusters $k$ in the clustering step of SemDeDup.}
\label{table:deduplication_efficiency}
\begin{tabular}{C{2.3cm}|C{0.7cm}|C{0.7cm}|C{0.7cm}|C{0.7cm}|C{0.7cm}|C{0.7cm}|C{0.7cm}|C{0.7cm}|C{0.7cm}}
\hline
\multicolumn{1}{c|}{Percentage of Data Kept} & \multicolumn{3}{c|}{ 63\%}  &  \multicolumn{3}{c|}{ 50\%} & \multicolumn{3}{c}{40\%} \\
\hline
\toprule
{Num. of Clusters} & 70K  & 50K  & 10K  & 70K  & 50K  & 10K & 70K  & 50K  & 10K \\ 
\midrule
$\eta$    &     94.4 &  94.6   &    95.3  & 90.1 & 90.6 & 91.3 &  88.3  &      89.0   &     90.8  \\  
\bottomrule
\end{tabular}
\end{table}
\vskip 1.0pt

 \begingroup
\setlength{\tabcolsep}{7.5pt} 
\renewcommand{\arraystretch}{1.4} 
\begin{table}
\centering
\caption[add short caption]{Time for running SemDeDup on LAION440M. Note that we report the total time for deduplication to different dataset size ratios.}
\label{table:semdedup_time}
\begin{tabular}{C{4.cm}|C{2.5cm}|C{2.5cm}|C{2.5cm}}
\toprule
{Operation / Time } &  Clustering Time & DeDup. Time & Total Time \\ 
\midrule

SemDeDup w/10K Clusters & 2h:36 @8 GPUs &  2h:20 @64 GPUs  & 4h:56  \\
\hline                                         
SemDeDup w/25K Clusters & 3h:52 @8 GPUs &  1h:19 @64 GPUs & 5h:11  \\
\hline
\rowcolor{LightCyan}
SemDeDup w/50K Clusters &  5h:59 @8 GPUs &  1h:22 @64 GPUs & 7h:21  \\
\hline
SemDeDup w/70K Clusters &  9h:02 @8 GPUs &  1h:10 @64 GPUs & 10h:12  \\
\hline
Training CLIP on 100\% of LAION440M for 32 Epochs & --- & --- & 69h:52 @176 GPUs  \\

\bottomrule
\end{tabular}
\end{table}
\endgroup

\section{CLIP Zeroshot Evaluation}
\begin{figure}[ht]
\vskip 0.2in
\begin{center}
\centerline{
 \includegraphics[width = 0.9\textwidth]{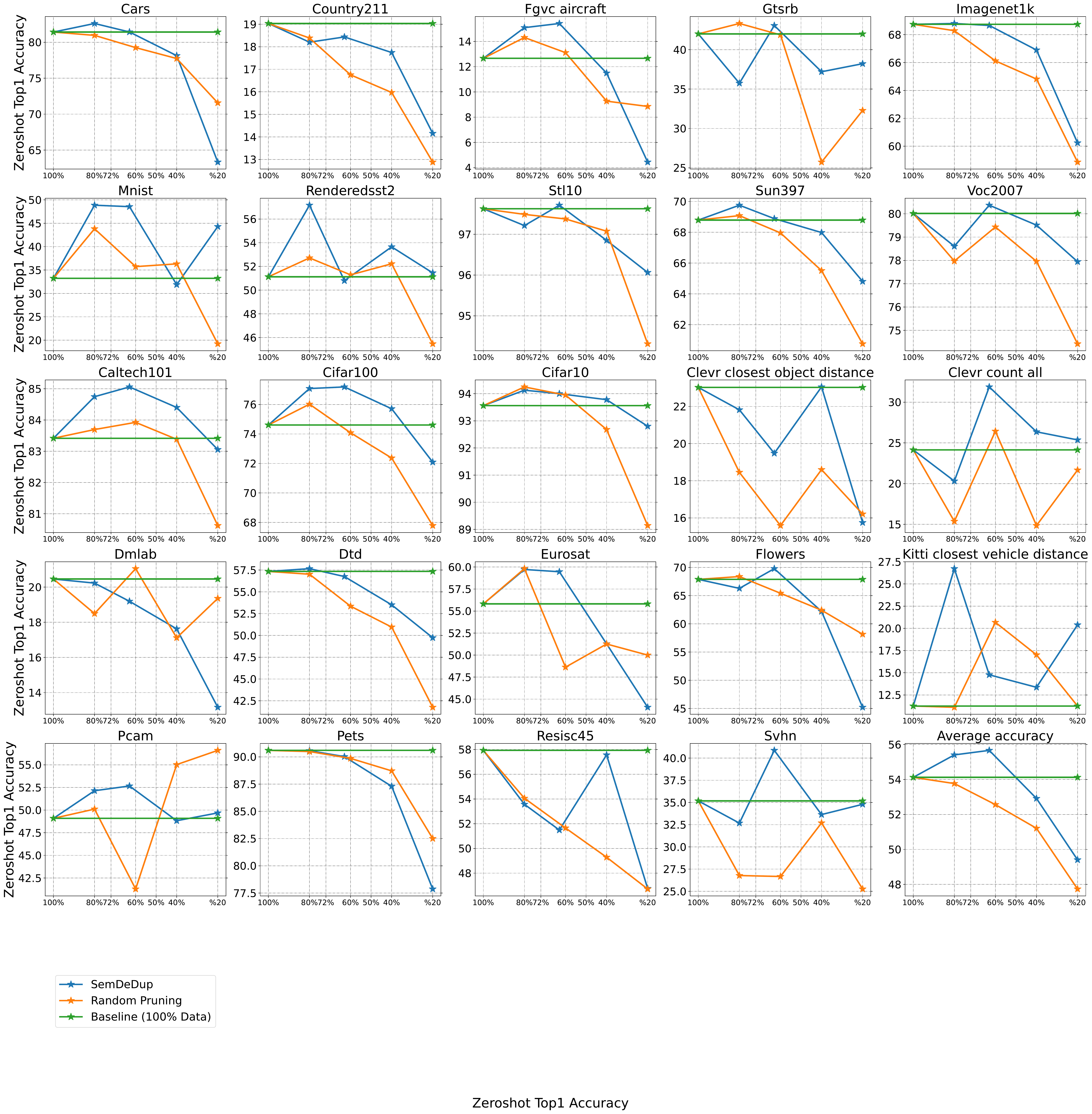}
}
\caption{Zeroshot performance of CLIP on 24 datasets. The last plot shows the average performance over all datasets.}
\label{fig:all_zershot_line_plots}
\end{center}
\vskip -0.2in
\end{figure}

\begin{figure}[ht]
\vskip 0.2in
\begin{center}
\centerline{
 \includegraphics[width = 0.9\textwidth]{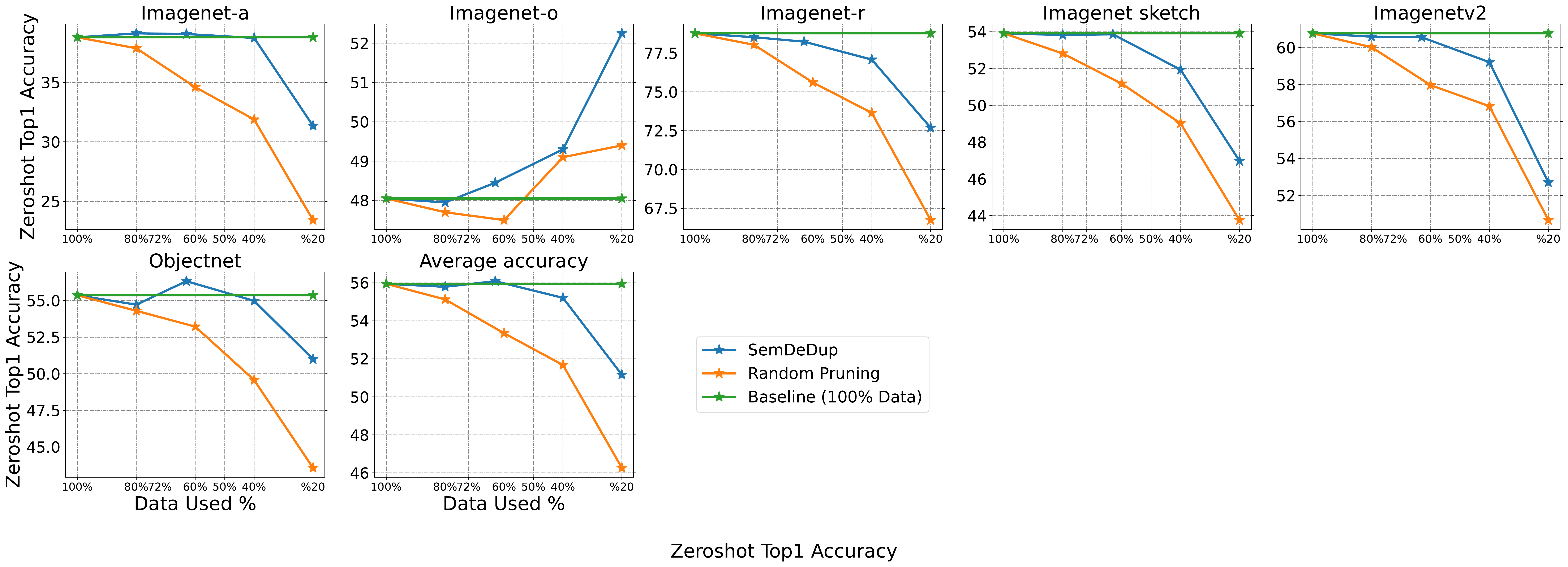}
}
\caption{Out-of-distribution zeroshot performance on 6 datasets. SemDeDup outperforms random pruning on all datasets for all fractions of dataset kept. The last plot shows the average performance over all datasets.}
\label{fig:fig:all_OOD_zershot_line_plots}
\end{center}
\vskip -0.2in
\end{figure}

In this section, we show the result of zeroshot evaluation for CLIP. We note that the models trained on  dataset deduplicated using SemDeDup outperform the baseline model in many tasks. In Table {\ref{table:zeroshot_results_table}} we list the top1 zeroshot accuracy on 24 tasks and in Table \ref{table:out_of_distribution_results_table} we show the top1 zeroshot accuracy on 6 datasets for out-of-distribution robustness evaluation. Our complete evaluation set has $30$ different datasets in total. When using only 63\% of LAION-440M, SemDeDup outperforms the baseline model in 19 out of the 30 tasks. Fig. (\ref{fig:all_zershot_line_plots}) and Fig. (\ref{fig:fig:all_OOD_zershot_line_plots}) show the performance of different models as a function of training dataset size.

\begin{figure}[ht]
\begin{center}
\centering
\begin{subfigure}{.44\columnwidth}
    \centering
\includegraphics[width =\columnwidth]{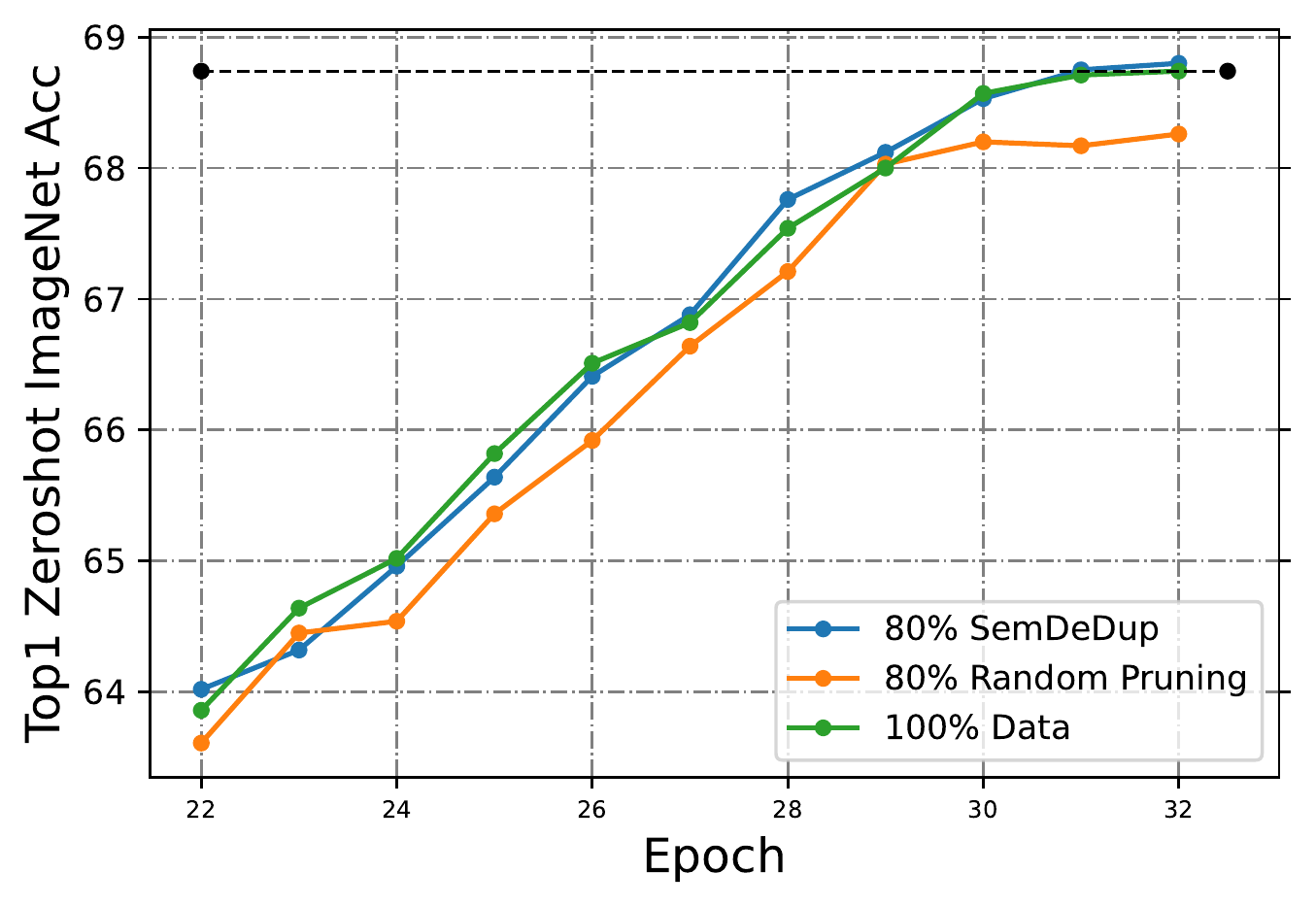}
    \caption{}
\end{subfigure}
\begin{subfigure}{.44\columnwidth}
    \centering
\includegraphics[width =\columnwidth]{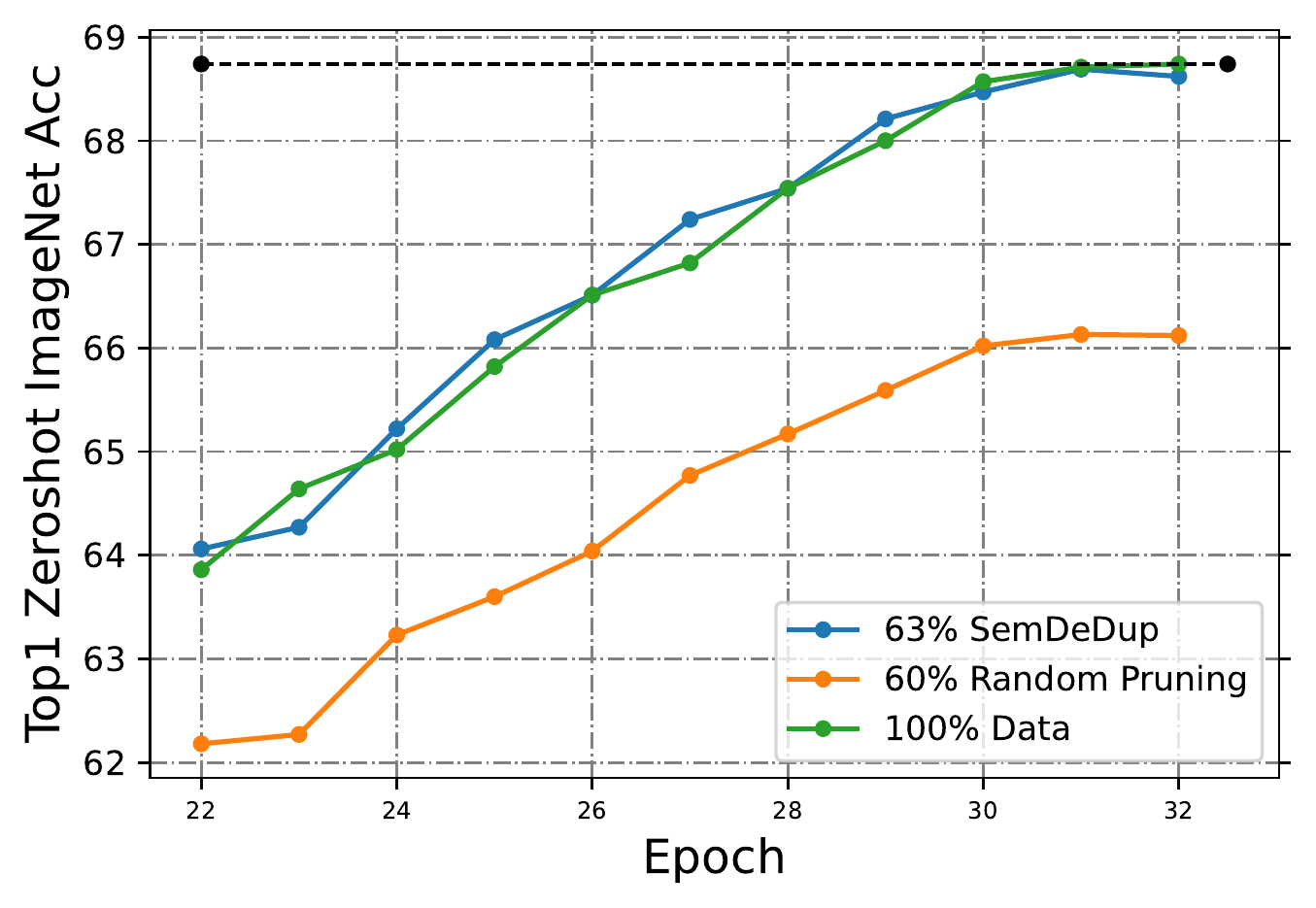}
    \caption{}
\end{subfigure}
\begin{subfigure}{.44\columnwidth}
    \centering
\includegraphics[width =\columnwidth]{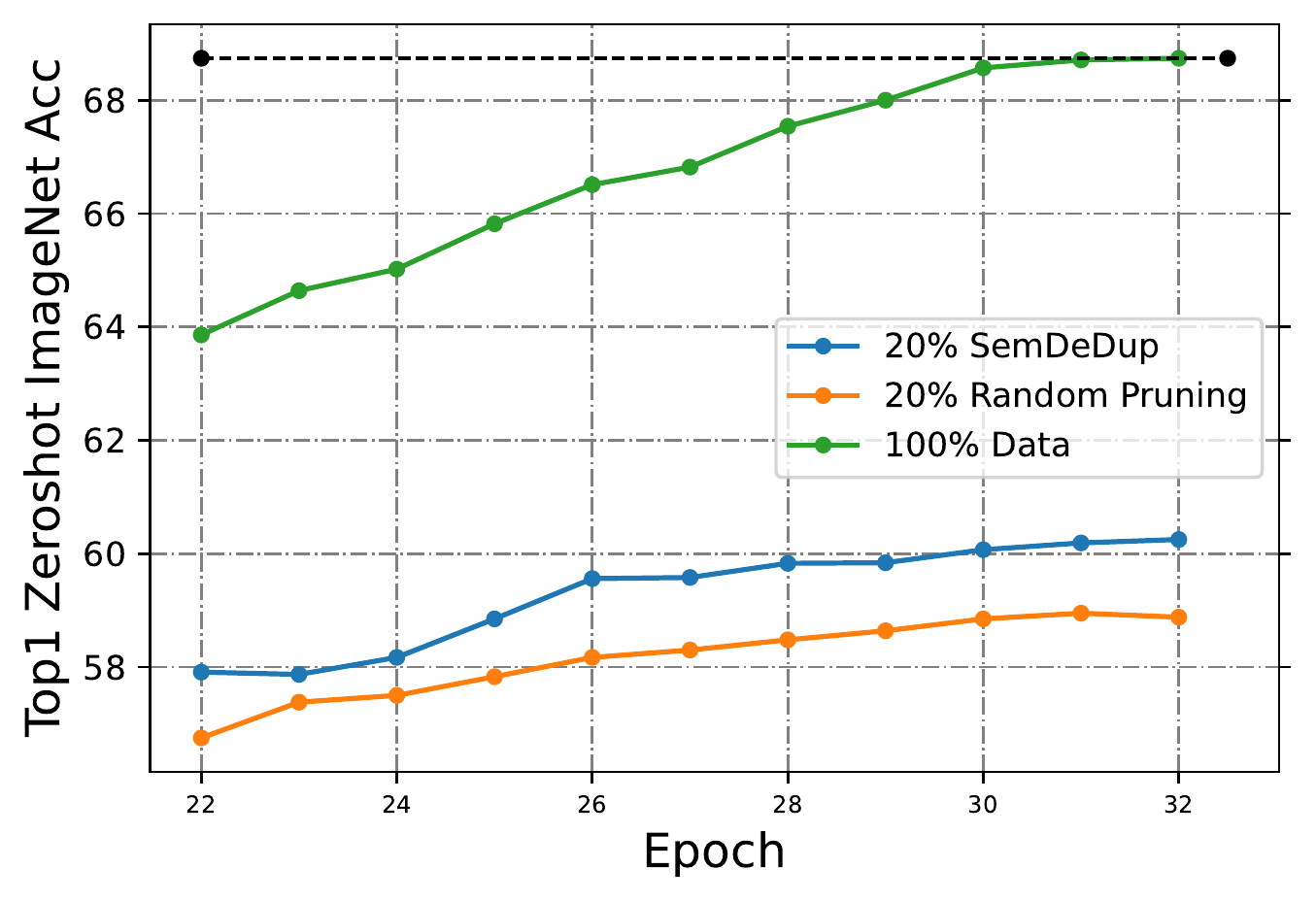}
    \caption{}
\end{subfigure}
\begin{subfigure}{.44\columnwidth}
    \centering
\includegraphics[width =\columnwidth]{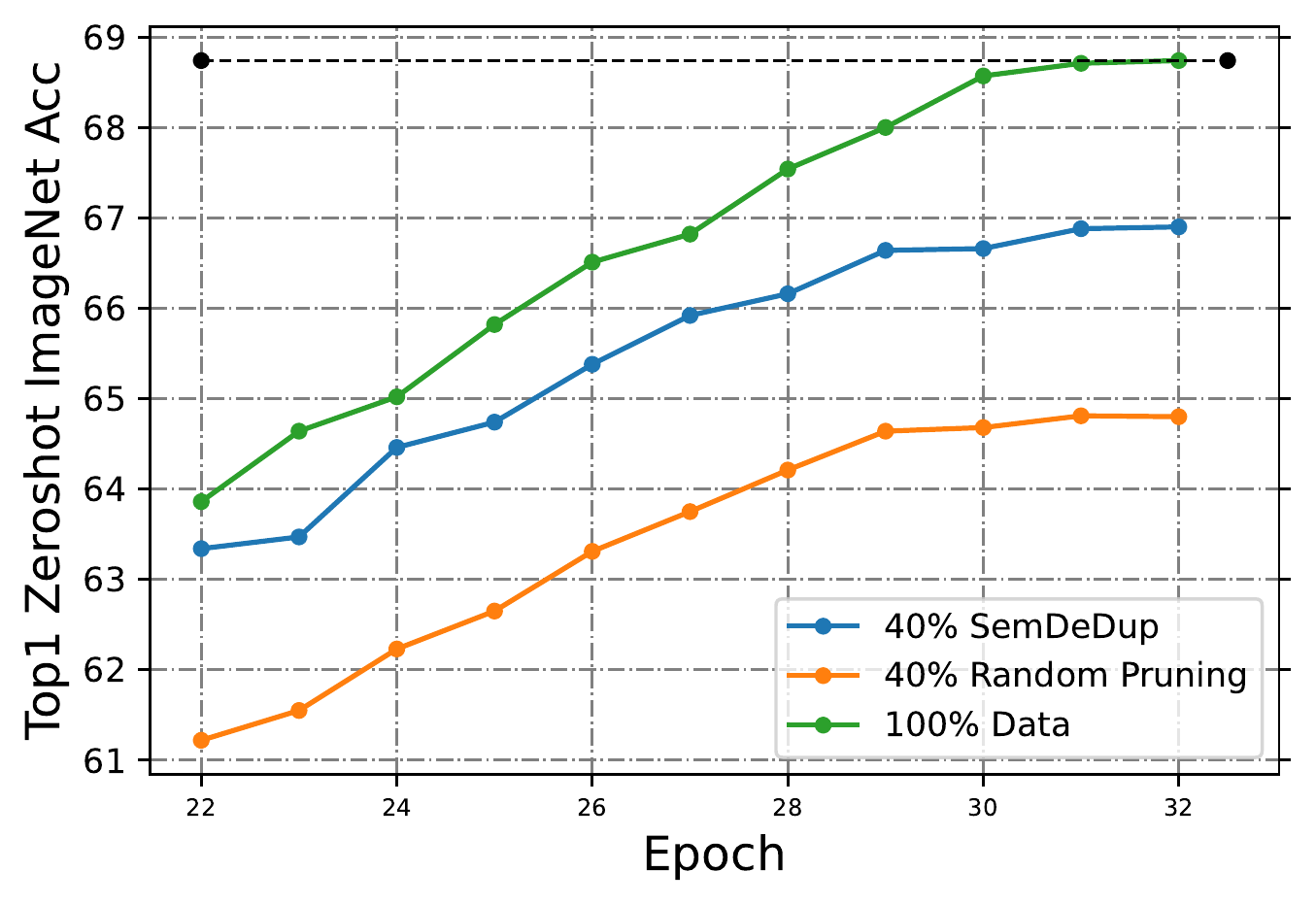}
    \caption{}
\end{subfigure}
   
\caption{\text{SemDeDup} is always better than training on random subset from LAION-440M. The plots show zeroshot top1 accuracy on ImageNet for CLIP models trained on different fractions of data.}
\label{fig:dedup_vs_rand}
\end{center}
\vskip -0.2in
\end{figure}

\begingroup 
\setlength{\tabcolsep}{1.5pt} 
\renewcommand{\arraystretch}{1.5} 
\begin{table}
\centering
\caption[add short caption]{Training parameters for CLIP}
\label{table:clip_training_parameters}
\begin{tabular}{c|c}
\toprule
Parameter & Value \\
\midrule
Model &   CLIP ViT-B-16 \\
Warmup & 2000\\
Epochs & 32\\
Batch size & 33,792\\
Learning rate   &  5.0e-4, cosine scheduler \\
Optimizer & AdamW, wd=0.2, betas=(0.9, 0.98), eps=1.0e-6\\
\bottomrule
\end{tabular}
\end{table}
\endgroup

\ZeroshotCLassificationResultsTable

\ZeroshotOutOfDistributionResultTable

\section{LAION-233M De-duplication} \label{sec:app-laion233}
\begin{figure}[ht]
\begin{center}
\includegraphics[width = 0.5\columnwidth]{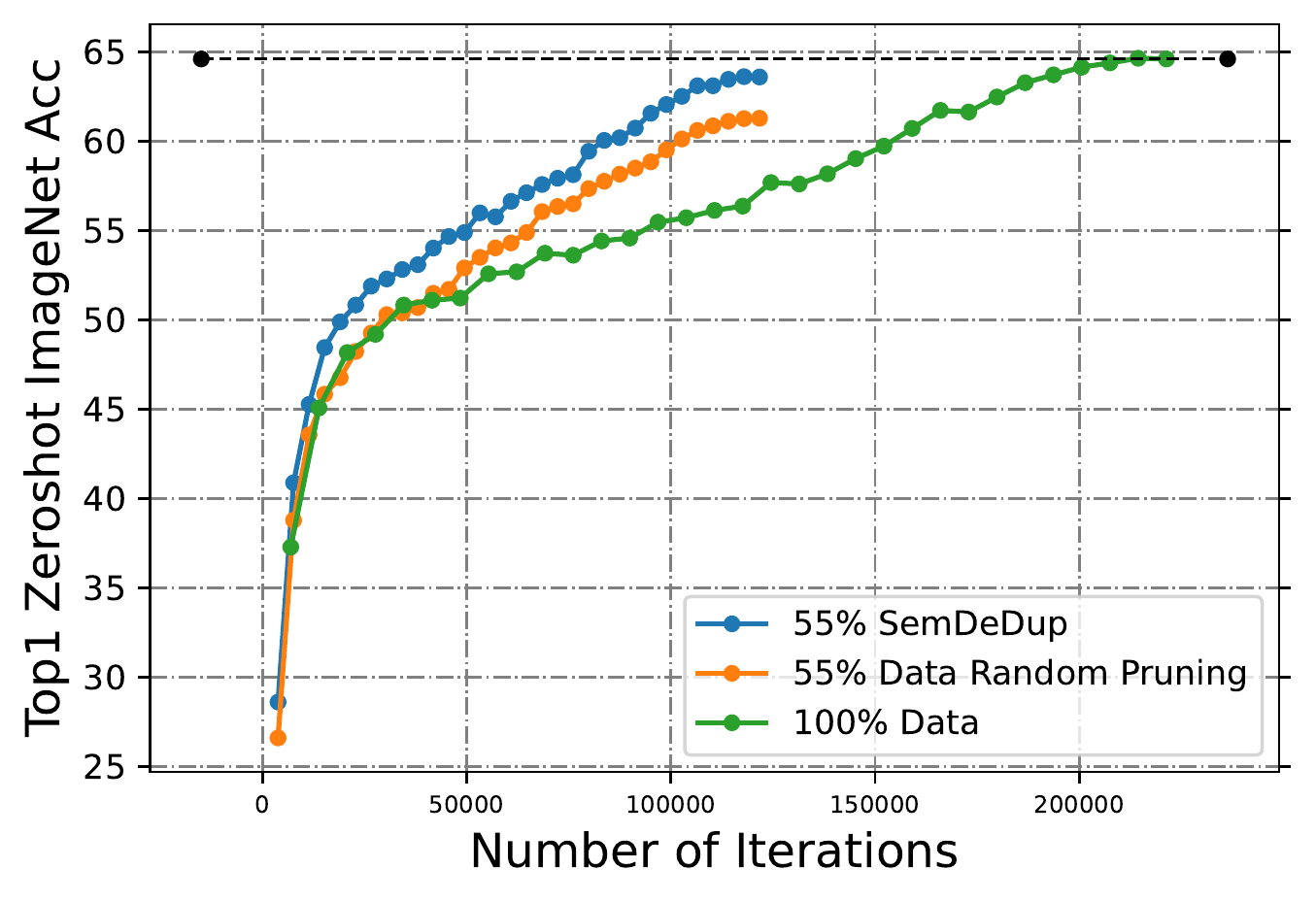}

\caption{Performance when deduplicating 233 million images from LAION-2B. We deduplicte LAION233M to 55\% of its size and train CLIP model on it. SemDeDup performs better than random pruning (63.61\%  vs 61.3\%). Training on  the whole 233 million examples gives 64.62\%. Note that the deduplicated dataset size here is 128 million only.}
\label{fig:laion233m_result}
\end{center}
\end{figure}

To support our results on LAION-440M, we also de-duplicate a much smaller dataset of 233 million images. We call this dataset LAION-233M. Usually, CLIP needs to be trained on more than 400 million images as introduced in \citep{clip}, so de-duplicating LAION-233M is more challenging in this respect.
We train a baseline model on the 233 million images and two models on 55\% of the data, one on a random subset and the other on deduplicated subset using SemDeDup. We trained all the models using the same hyperparameters we used for training on LAION-440M. We show ImageNet top1 zeroshot accuracy for these models in Fig. \ref{fig:laion233m_result}. The baseline model achieved 64.62\% accuracy, while the SemDeDup model achieved 63.61\% outperforming the model trained on the random subset (61.3\% accuracy).

\begin{table}
\caption{Performance after training on deduplicated data for the same number of iterations as training on 100\% of the data.}
\label{table:match_iterations_training}
\begin{tabular}{c|c|c|c|c|c}
\toprule
{Metric / Model} &   dedup40 &  dedup50 &  dedup60 & dedup70 & Baseline (100\%) \\ 
\midrule
Top1 IN Zeroshot Acc. & 68.35 & \textbf{68.92} & \textbf{69.04} & \textbf{69.14} & 68.74 \\  
Top5 IN Zeroshot Acc. & \textbf{91.64} & \textbf{91.82} & \textbf{91.86} & \textbf{91.73} & 91.42 \\
\bottomrule
\end{tabular}
\end{table}

\begin{figure}[ht]
\vskip 0.2in
\begin{center}
\centerline{
 \includegraphics[width = 0.6\textwidth]{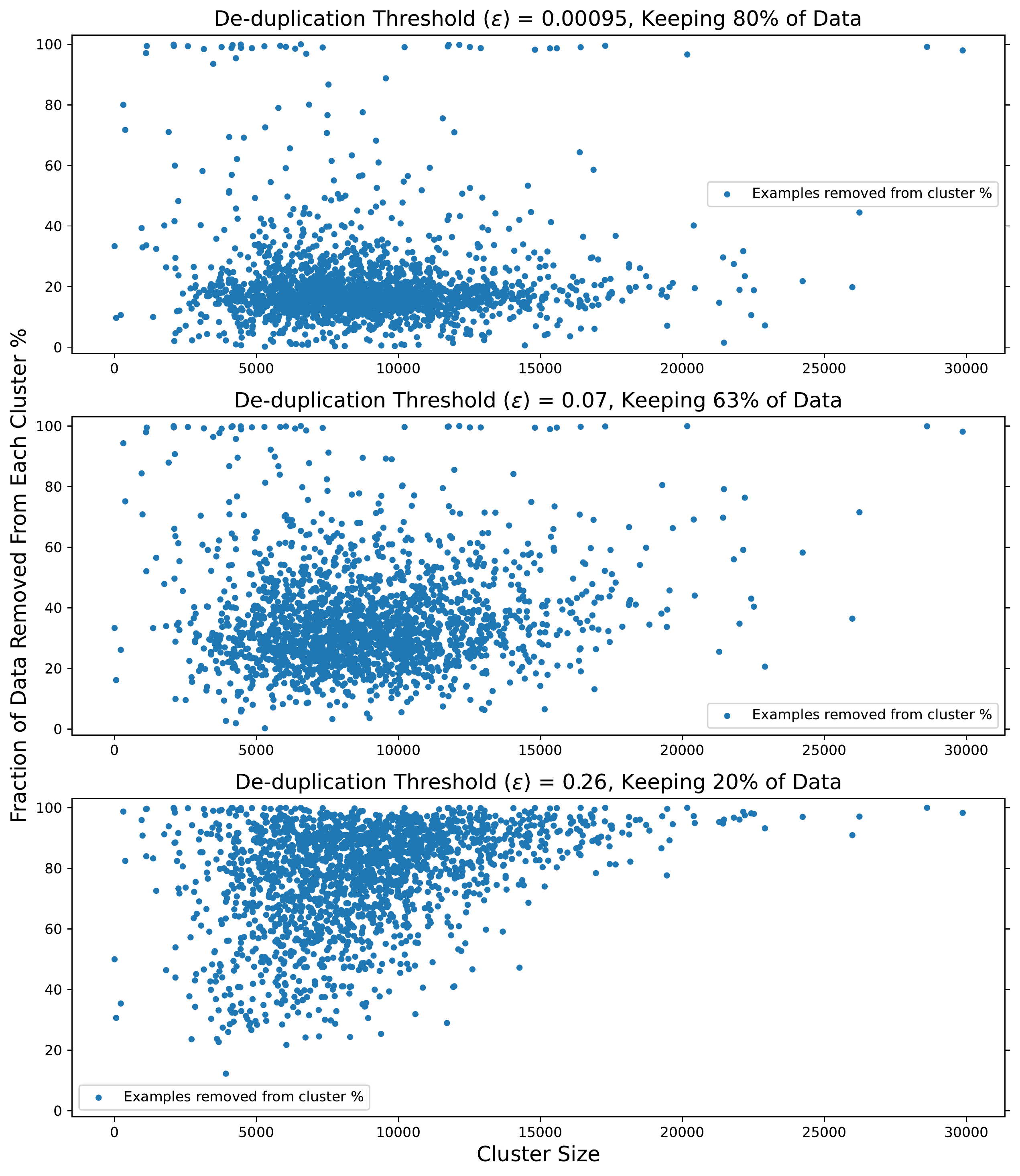}
}
\caption{How many images can we remove from each cluster? Moving from top to down we increase $\epsilon$ value. The x-axis corresponds to the cluster size. The y-axis corresponds to the fraction of data removed from each cluster by SemDeDup. As we increase $\epsilon$, more examples are removed from each cluster. We notice that most of the examples from the large clusters (the points to the right) are removed when $\epsilon$ becomes large. The points in this figure are for 2000 clusters sampled randomly from a total of 50,000 clusters.}
\label{fig:Fraction_of_Data_Removed_From_Cluster}
\end{center}
\vskip -0.2in
\end{figure}

\begin{figure}[ht]
\vskip 0.2in
\begin{center}
\centerline{
 \includegraphics[width = 0.5\textwidth]{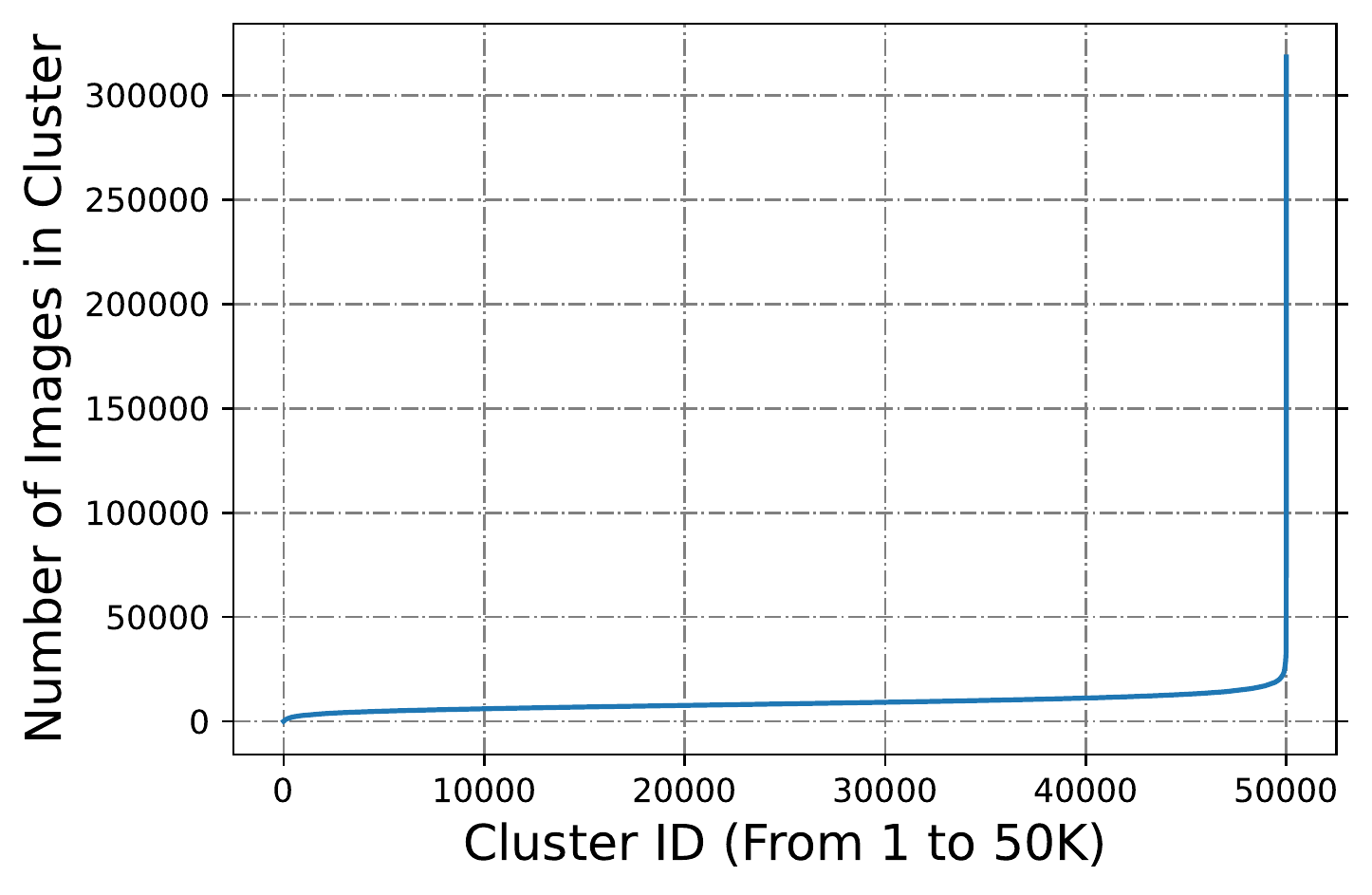}
}

\caption{The number of images in each cluster for 50,000 clusters of LAION-440M images after running k-means clustering in the embedding space. The average cluster size is 8748, but we also see a few clusters with more than 300,000 examples. }
\label{fig:cluster_size}
\end{center}
\vskip -0.2in
\end{figure}



\begin{table}
\centering
\caption[add short caption]{}
\label{alg:semdedup}
\begin{tabular}{C{14.5cm}}
\par\rule{\textwidth}{0.5pt} 
PyTorch-style Pseudo Code For SemDeDup
\par\rule{\textwidth}{0.5pt} 
\begin{lstlisting}[language=Python]
#Input: cluster_embeddings, num_clusters, epsilon

for i in range(num_clusters):
    # Load cluster embeddings.
    cluster_i_embeddings = cluster_embeddings[i]
    
    # Sort the cluster embeddings by the distance to the cluster centroid.
    cluster_i_embeddings = sort_by_distance_to_cluster_centroid(cluster_i_embeddings, descending = True) 
    
    # We use descending=True/False for keeping examples with low/high similarity to cluster centroids. We  ignore this step for keeping random examples from each group of similar examples. See Appendix D for more details about this step.

    # Compute the pairwise cosine similarity between embeddings
    pairwise_sim_matrix = cluster_i_embeddings @ cluster_i_embeddings.T 
    
    triu_sim_matrix = torch.triu(pairwise_sim_matrix, diagonal = 1) 
    
    M = torch.max(triu_sim_matrix, dim=0)[0]

    # Check if the maximum similarity <= the threshold.
    points_to_keep_from_cluster_i = M <= 1-epsilon  
\end{lstlisting}  
\par\rule{\textwidth}{0.5pt} 
\end{tabular}
\end{table}

\section{Visualizing Examples Before and After De-duplication}
\begin{figure}[ht]
\vskip 0.2in
\begin{center}
\centerline{\includegraphics[width = 0.75\textwidth]{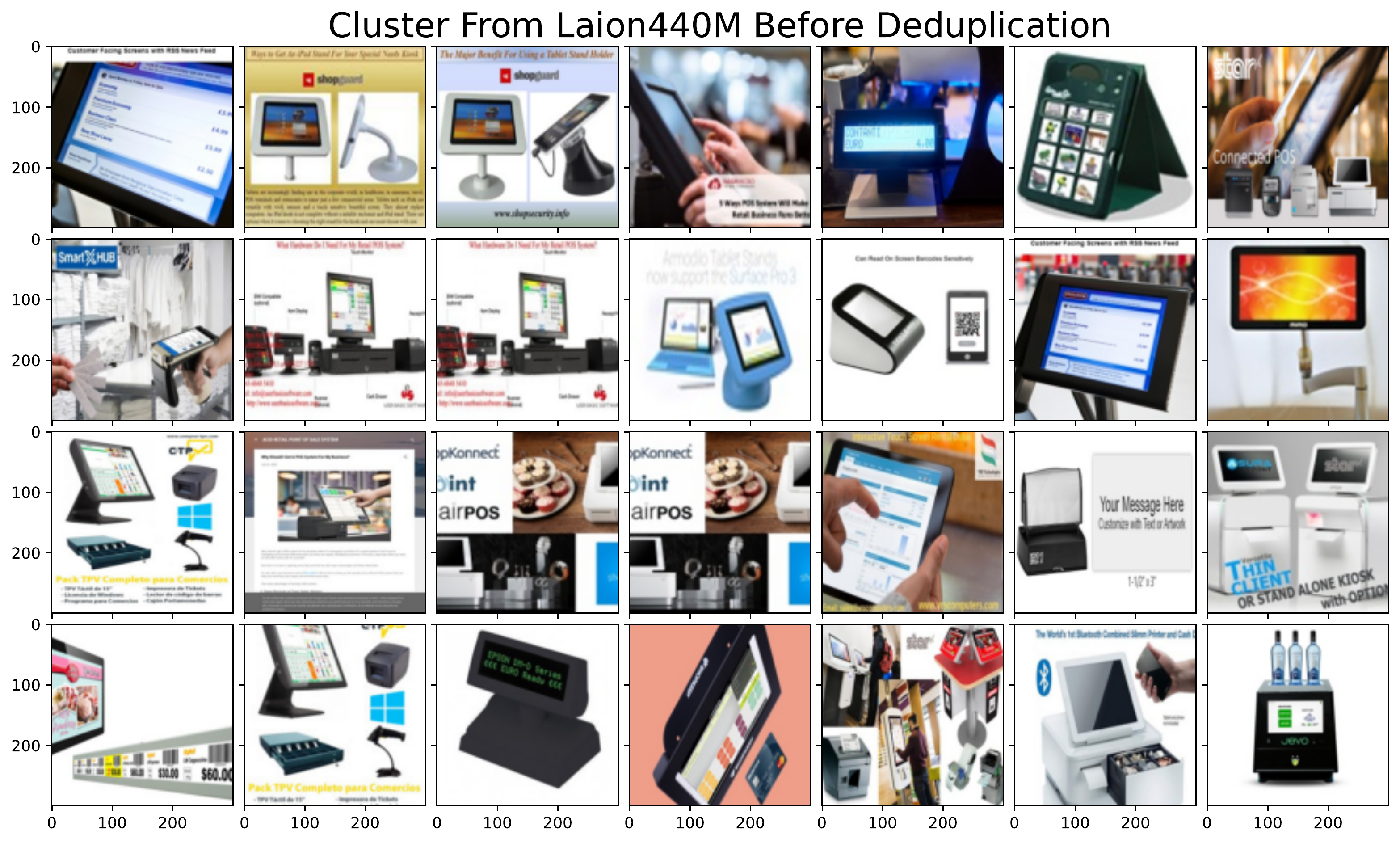}}
\centerline{\includegraphics[width = 0.75\textwidth]{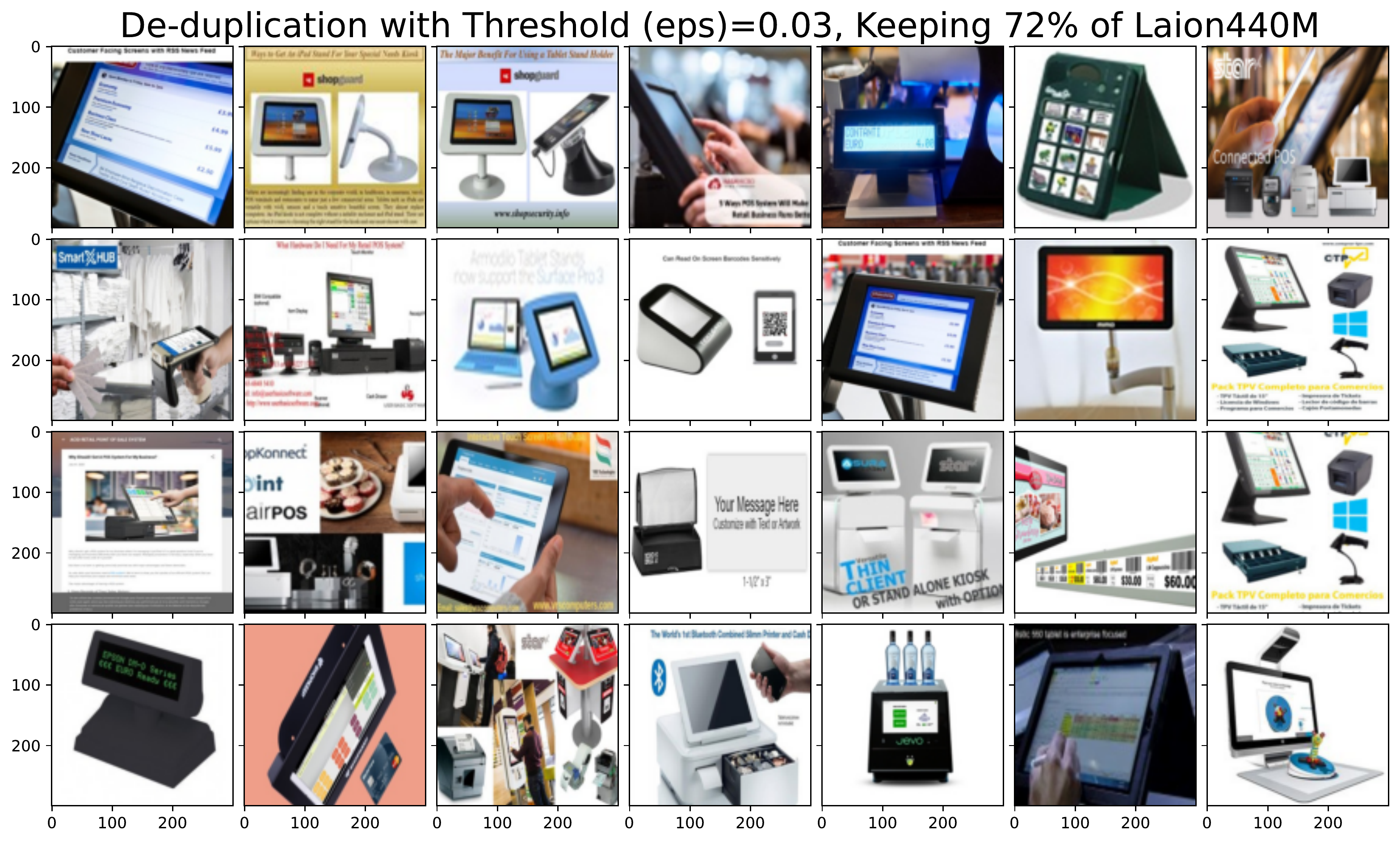}}
\centerline{\includegraphics[width = 0.75\textwidth]{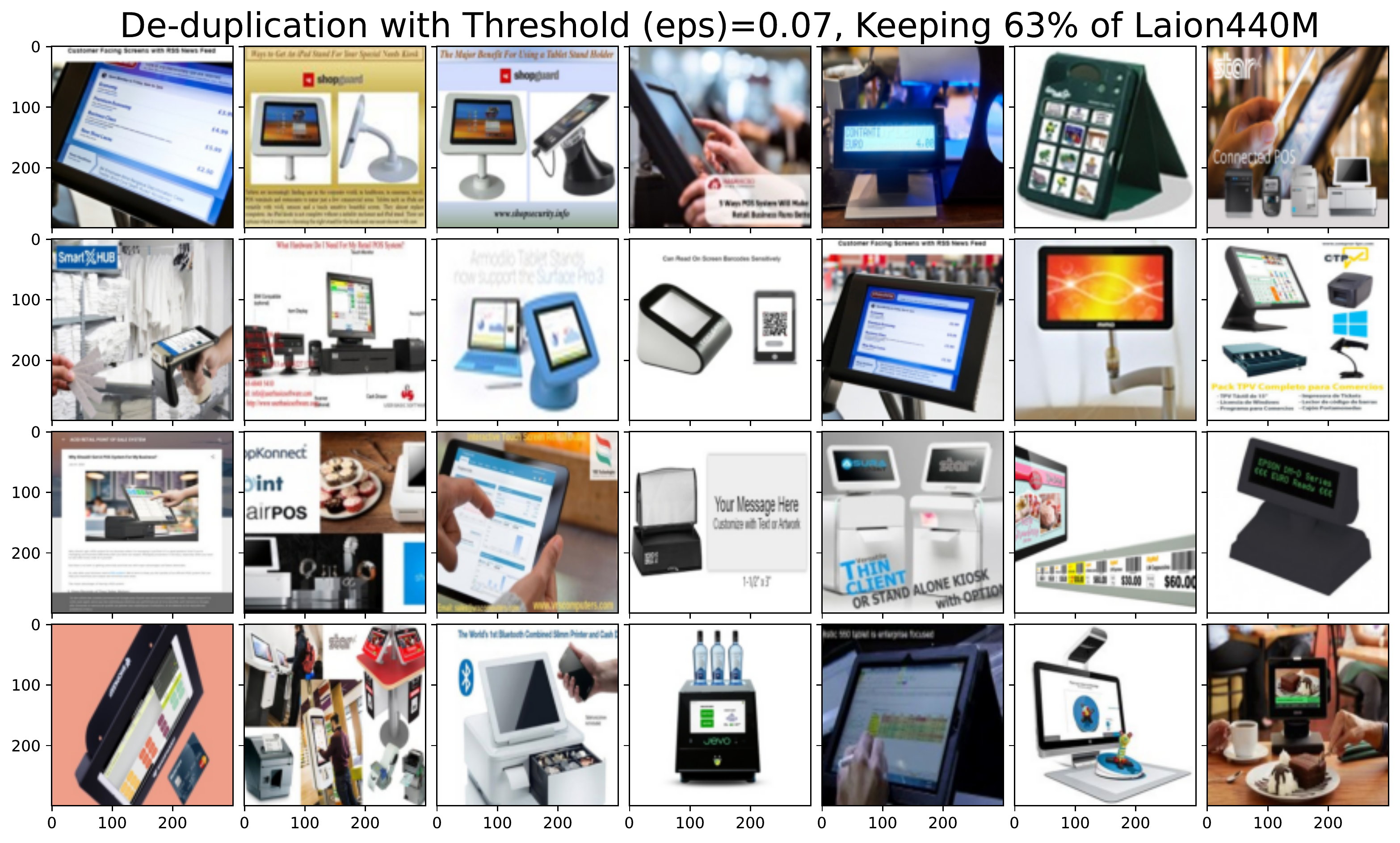}}

\caption{Examples from the same cluster from LAION-440M dataset before and after de-duplication. Images are sorted by cosine similarity to the cluster centroid. As we increase the deduplication threshold we start to see more unique images.}
\label{fig:before_and_after_1}
\end{center}
\vskip -0.2in
\end{figure}

\begin{figure}[ht]
\vskip 0.2in
\begin{center}
\centerline{\includegraphics[width = 0.75\textwidth]{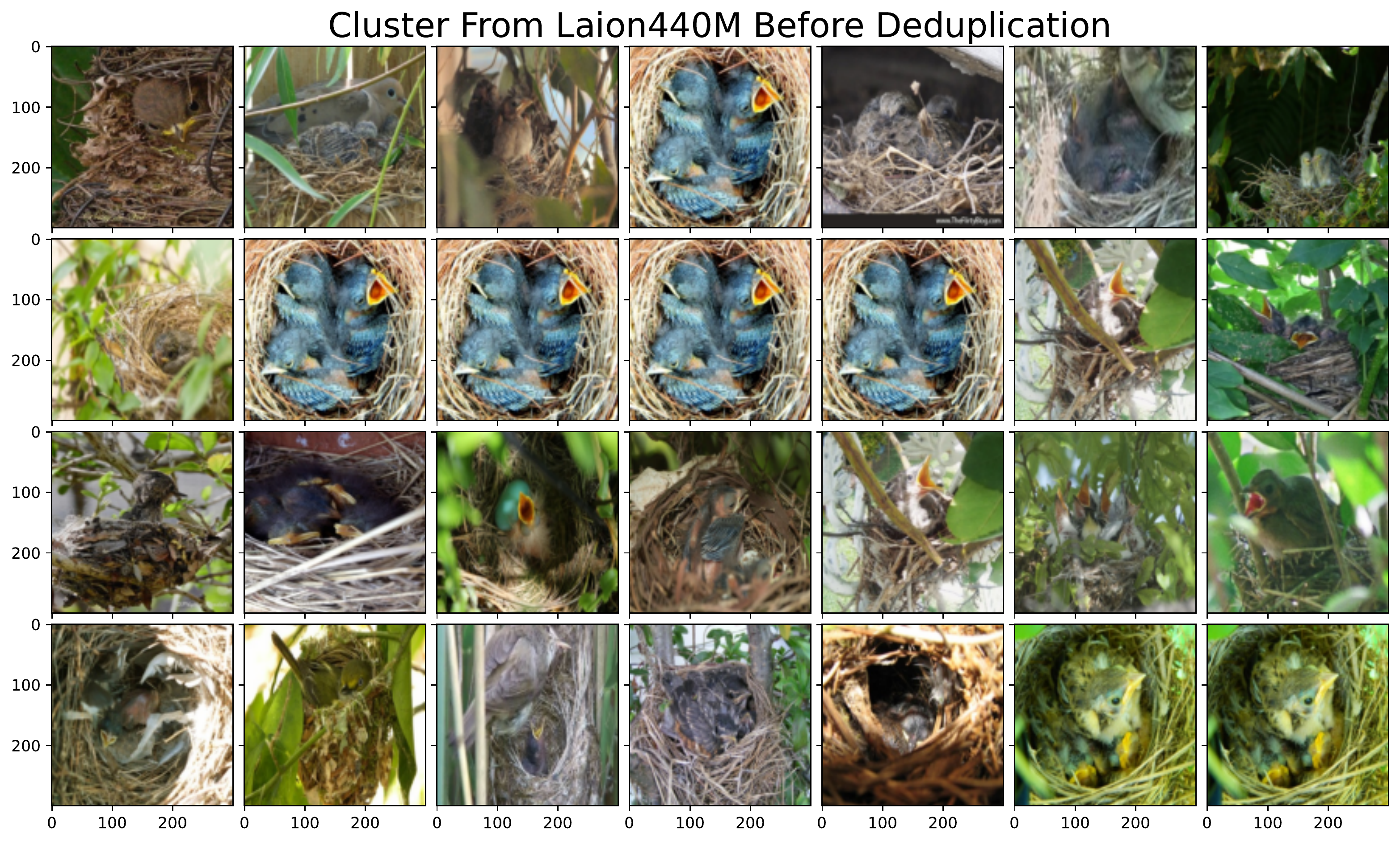}}
\centerline{\includegraphics[width = 0.75\textwidth]{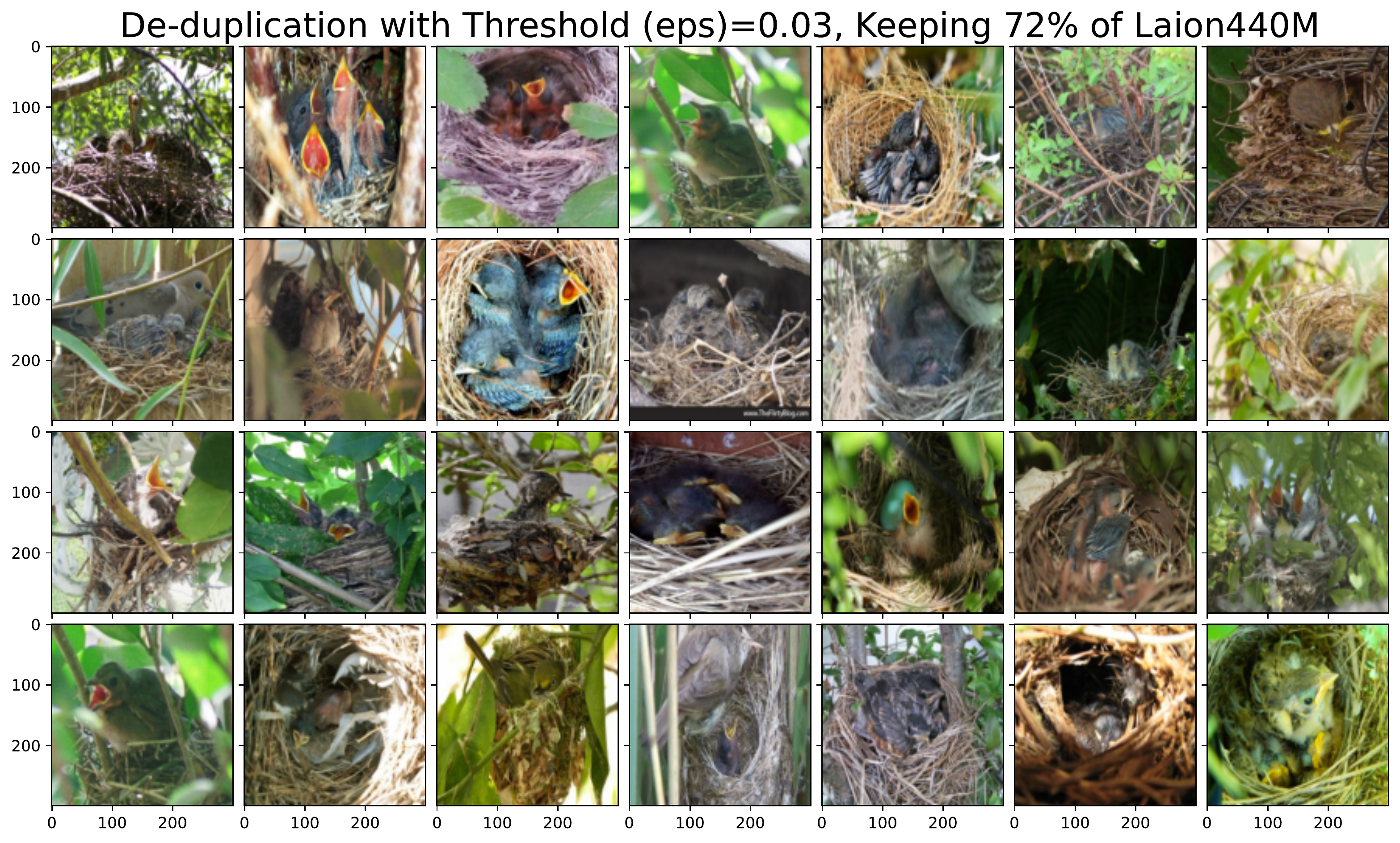}}
\centerline{\includegraphics[width = 0.75\textwidth]{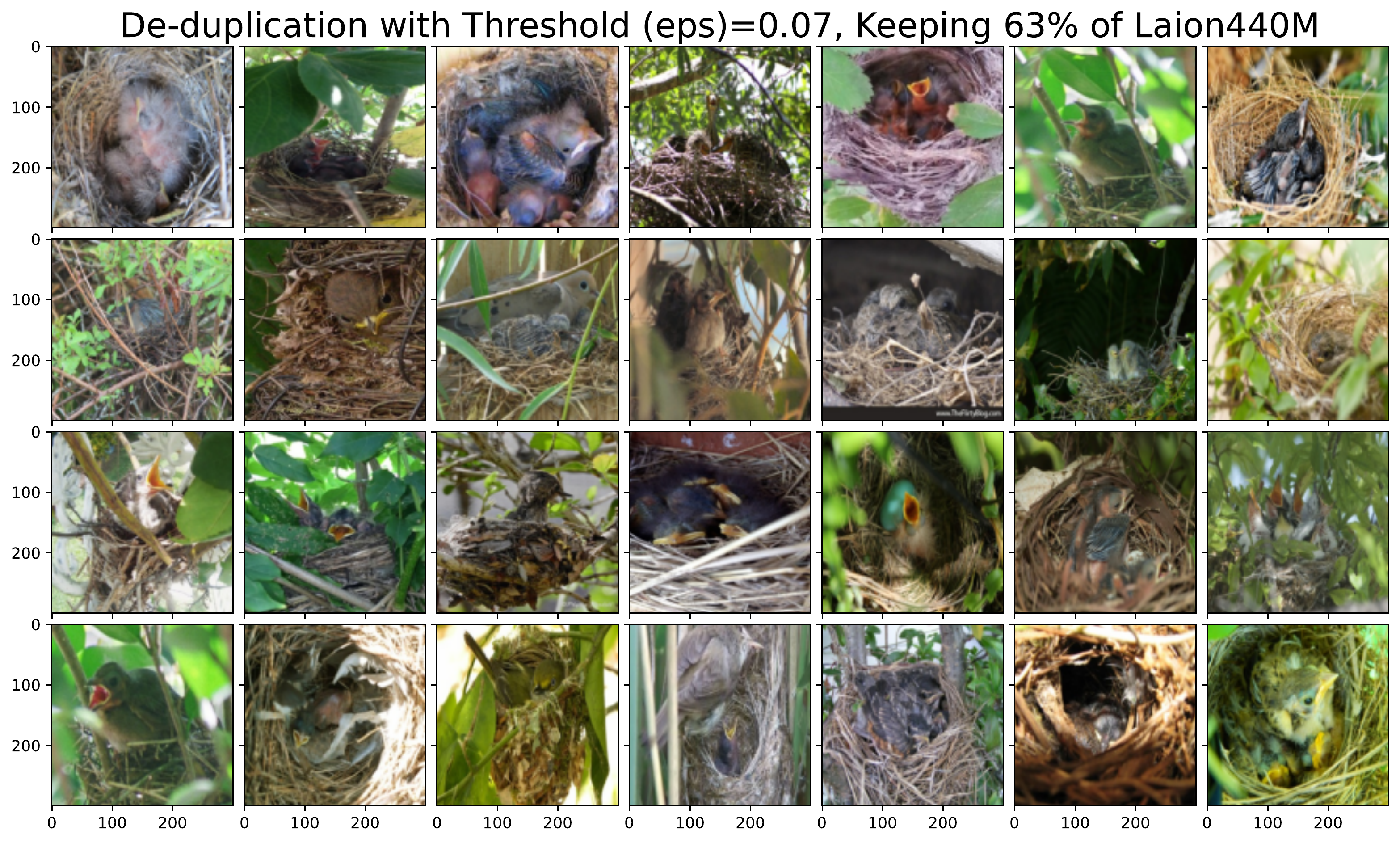}}

\caption{Examples from the same cluster from LAION440M dataset before and after de-duplication. Images are sorted by cosine similarity to the cluster centroid. As we increase the deduplication threshold we start to see more unique images.}
\label{fig:before_and_after_2}
\end{center}
\vskip -0.2in
\end{figure}

To visually show which images are removed by SemDeDup from LAION440M dataset, we visualize some images from a random cluster before and after deduplication. To do that, we choose a cluster randomly and sort its examples by the cosine similarity to the centroid. By doing that, we can show similar images next to each other in a sequence. Then we visualize a sequence of images before de-duplication. After that, we run SemDeDup, remove duplicates, and sort the remaining examples again. Finally, we visualize the sequence of images from the same indices we visualize before de-duplication. Figures (\ref{fig:before_and_after_1} and \ref{fig:before_and_after_2}) show that after applying SemDeDup with different values for the de-duplication threshold $\epsilon$, we keep the unique images. 

\begin{figure}[ht]
\begin{center}
\centerline{\includegraphics[width = 0.8\textwidth]{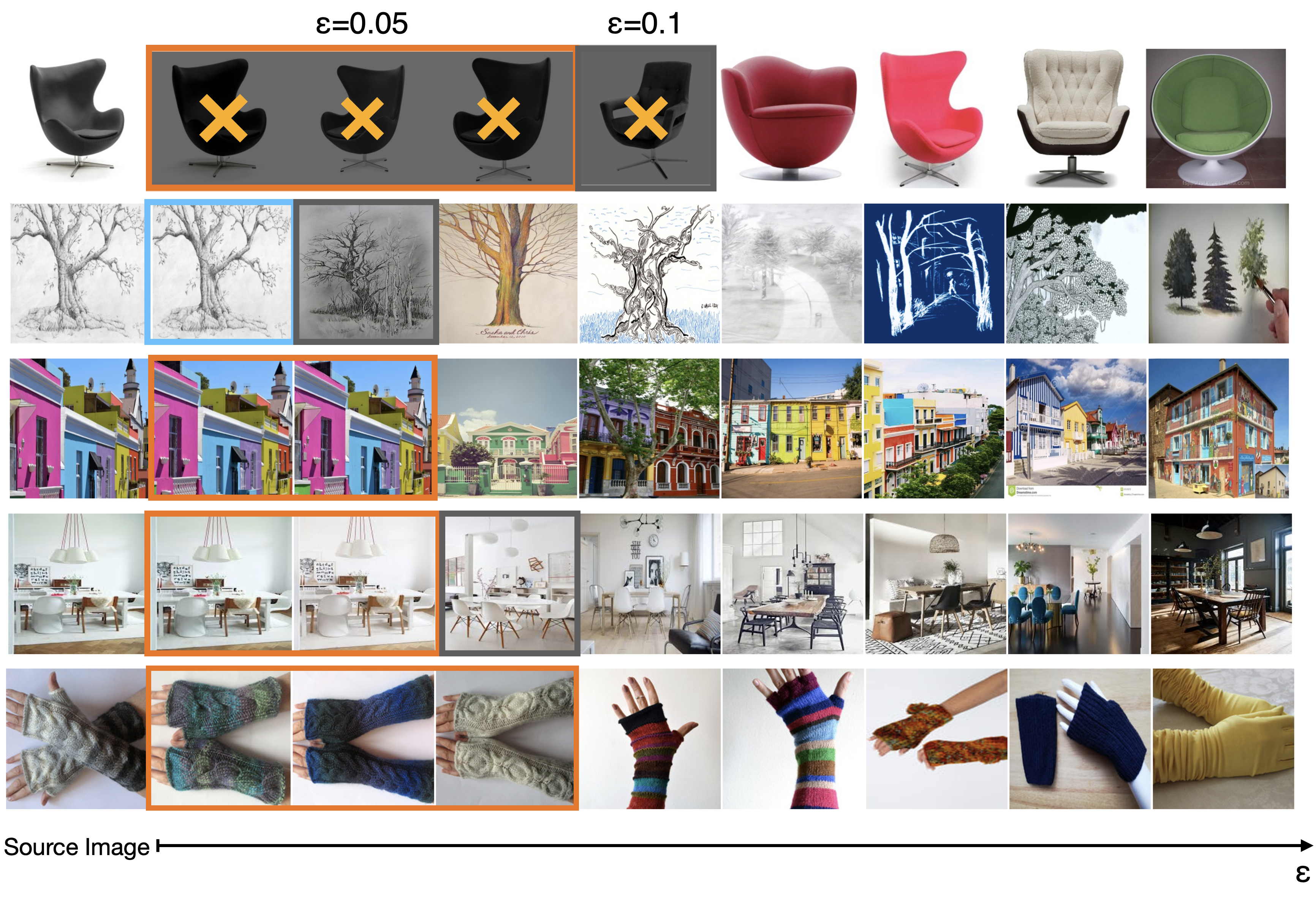}}

\caption{For each of the source images \textbf{(left column)}, we can retrieve a set of similar images from LAION440M. For each of  the source images, we show a set of images with the highest cosine similarity to it. Images are sorted from left to right by their cosine similarity (1- $\epsilon$) to the source image. By changing $\epsilon$ value, we can identify \textcolor{myblue}{perceptual duplicates}, \textcolor{mybrown}{semantic duplicates}, and \textcolor{mygray}{semantically redundant} examples for the source images. As we see in the \textbf{first row}, by increasing $\epsilon$ we can remove many examples that are semantically similar to the source image.}
\label{fig:semantically_similar_images_appendix}
\end{center}
\end{figure}



\section{Perplexity Values for SemDeDup on Language Modeling} \label{app:ppl}

\begin{figure}[ht]
\begin{center}
\begin{tabular}{L{3.5cm}C{2cm}C{2cm}C{2cm}C{2cm}}
\toprule
method & Baseline (no pruning) & NearDup from \citep{LeeDedup} &          Random &        SemDedup \\
validation set       &                       &                               &                 &                 \\
\midrule
C4                   &        38.95 +/- 0.07 &                39.46 +/- 0.14 &  39.51 +/- 0.07 &  \textbf{39.35} +/- 0.16 \\
opt\_valid            &        47.13 +/- 0.21 &                47.33 +/- 0.20 &  47.75 +/- 0.23 &  \textbf{47.18} +/- 0.29 \\
prompts\_with\_answers &        29.60 +/- 0.15 &                29.79 +/- 0.11 &  29.95 +/- 0.11 &  \textbf{29.69} +/- 0.19 \\
\bottomrule
\end{tabular}

\caption{Comparison of perplexity values for 125M OPT model after pruning via different methods at 96\% pruning. Note that \citep{LeeDedup} pruned 3.9 \% of examples, while above Random and SemDeDup prune 4\% of examples. Mean and standard deviation provided across 3 training seeds. Note that the Baseline column does not prune data (which is why the perplexities are lower) and bolded numbers compare between Random, SemDedup, and NearDup.}
\label{table:nlp_appendix_125m_c4_matched_epoch}
\end{center}
\end{figure}

\begin{figure}
\begin{center}
\begin{tabular}{llll}
\toprule
method & Baseline (no pruning) &          Random &        SemDedup \\
validation set       &                       &                 &                 \\
\midrule
C4                   &        38.95 +/- 0.07 &  42.16 +/- 0.03 &  \textbf{41.98} +/- 0.09 \\
opt\_valid            &        47.13 +/- 0.21 &  50.66 +/- 0.11 &  \textbf{49.04} +/- 0.16 \\
prompts\_with\_answers &        29.60 +/- 0.15 &  31.65 +/- 0.16 &  \textbf{30.98} +/- 0.13 \\
\bottomrule
\end{tabular}
\caption{Comparison of perplexity values for 125M OPT model after pruning via different methods at 80\% pruning. Mean and standard deviation provided across 3 training seeds. Note that the Baseline column does not prune data (which is why the perplexities are lower) and bolded numbers compare between Random and SemDedup.}
\label{table:nlp_appendix_125m_c4_matched_epoch_80}
\end{center}
\end{figure}

\begin{figure}
\begin{center}
\begin{tabular}{llll}
\toprule
method &        Baseline &          Random &        SemDedup \\
validation set       &                 &                 &                 \\
\midrule
C4                   &  38.95 +/- 0.07 &  87.09 +/- 0.21 &  \textbf{67.32} +/- 0.16 \\
opt\_valid            &  47.13 +/- 0.21 &  95.05 +/- 0.31 &  \textbf{70.17} +/- 0.16 \\
prompts\_with\_answers &  29.60 +/- 0.15 &  60.63 +/- 1.12 &  \textbf{43.16} +/- 0.19 \\
\bottomrule
\end{tabular}

\caption{Comparison of perplexity values for 125M OPT model after pruning via different methods at 20\% pruning. Mean and standard deviation provided across 3 training seeds. Note that the Baseline column does not prune data (which is why the perplexities are lower) and bolded numbers compare between Random and SemDedup.}
\label{table:nlp_appendix_125m_c4_matched_epoch_20}
\end{center}
\end{figure}

\begin{figure}
\begin{center}
\begin{tabular}{L{3.5cm}C{2cm}C{2cm}C{2cm}C{2cm}}
\toprule
method & Baseline (no pruning) & NearDup from \citep{LeeDedup} &          Random &        SemDedup \\
validation set       &                       &                               &                 &                 \\
\midrule
C4                   &        46.16 +/- 0.00 &                46.85 +/- 0.00 &  \textbf{46.15} +/- 0.00 &  46.56 +/- 0.00 \\
opt\_valid            &        55.69 +/- 0.00 &                55.27 +/- 0.00 &  55.20 +/- 0.00 &  \textbf{54.88} +/- 0.00 \\
prompts\_with\_answers &        34.04 +/- 0.00 &                33.93 +/- 0.00 &  33.91 +/- 0.00 &  \textbf{33.83} +/- 0.00 \\
\bottomrule
\end{tabular}
\caption{Comparison of perplexity values for 1.3b OPT model after pruning via different methods at 96\% pruning. Note that \citep{LeeDedup} pruned 3.9 \% of examples, while above Random and SemDeDup prune 4\% of examples. Due to compute restrictions we do not provide random seed standard deviations.}
\label{table:nlp_appendix_1.3b_c4_matched_epoch_96}
\end{center}
\end{figure}

\begin{figure}
\begin{center}
\begin{tabular}{llll}
\toprule
method & Baseline (no pruning) &           Random &         SemDedup \\
validation set       &                       &                  &                  \\
\midrule
C4                   &        46.16 +/- 0.00 &  303.71 +/- 0.00 &  \textbf{108.95} +/- 0.00 \\
opt\_valid            &        55.69 +/- 0.00 &  347.35 +/- 0.00 &  \textbf{109.68} +/- 0.00 \\
prompts\_with\_answers &        34.04 +/- 0.00 &  269.96 +/- 0.00 &   \textbf{72.64} +/- 0.00 \\
\bottomrule
\end{tabular}
\caption{Comparison of perplexity values for 1.3b OPT model after pruning via different methods at 20\% pruning.  Note that the Baseline column does not prune data (which is why the perplexities are lower) and bolded numbers compare between Random and SemDedup. Due to compute restrictions we do not provide random seed standard deviations.}
\label{table:nlp_appendix_1.3b_c4_matched_epoch_20}
\end{center}

\end{figure}

\section{Qualitative Examples of SemDeDup on C4} \label{app:qualitative}
\clearpage


\begin{figure}[ht]
\begin{center}
\includegraphics[width=0.49\columnwidth]{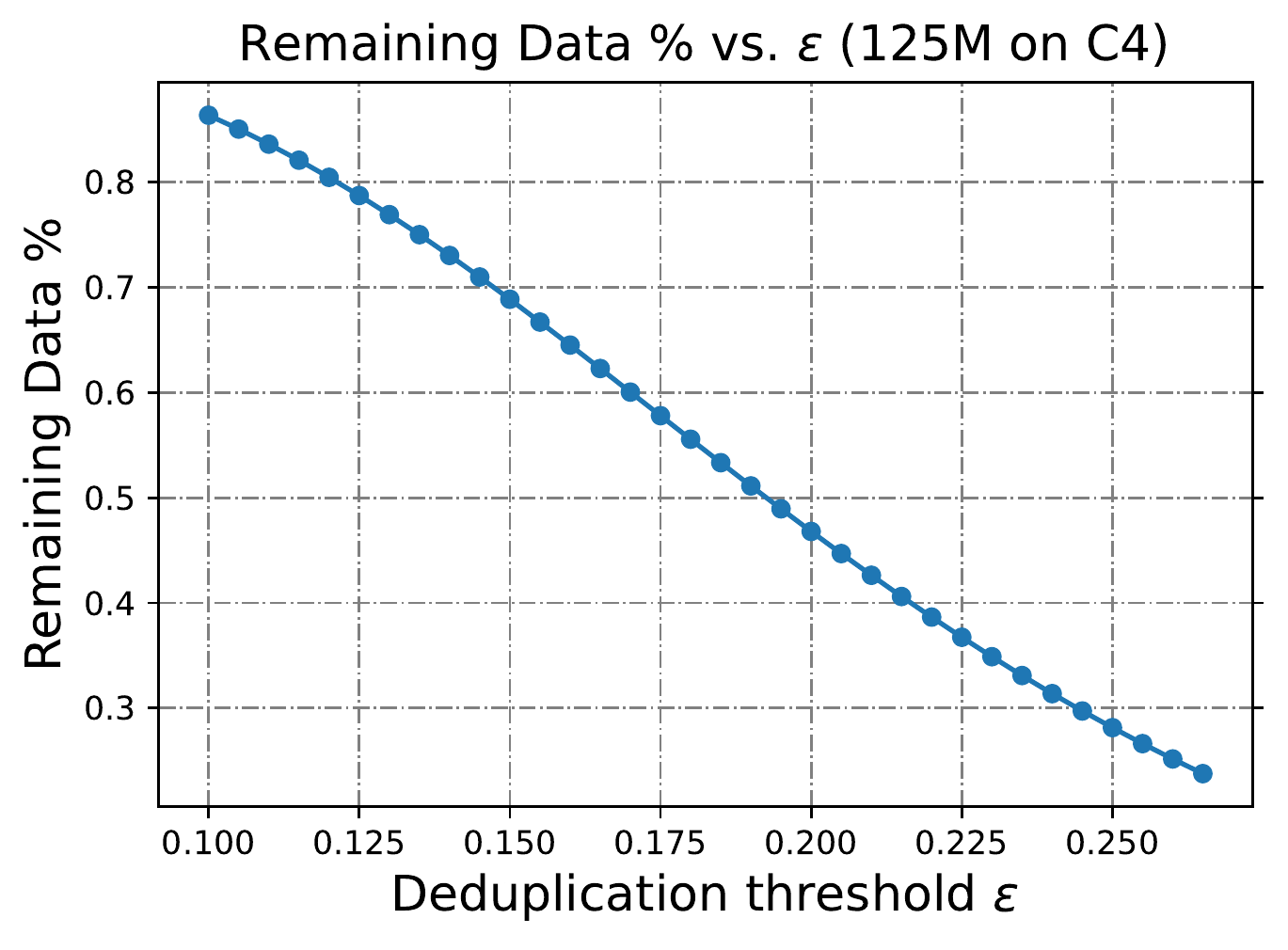}

\caption{\textbf{Percent Data Remaining versus $\epsilon$ for C4}.  The x-axis corresponds to different values of $\epsilon$ from Section~\ref{sec:methods}, and the y-axis represents the corresponding fraction of data in our subset of C4.}
\label{fig:nlp_appendix_125M_eps_vs_frac_data}
\end{center}
\end{figure}

\begin{figure}[ht]
\vskip 0.2in
\begin{center}
\centering
\begin{subfigure}{.40\columnwidth}
    \centering
    \includegraphics[width=\columnwidth]{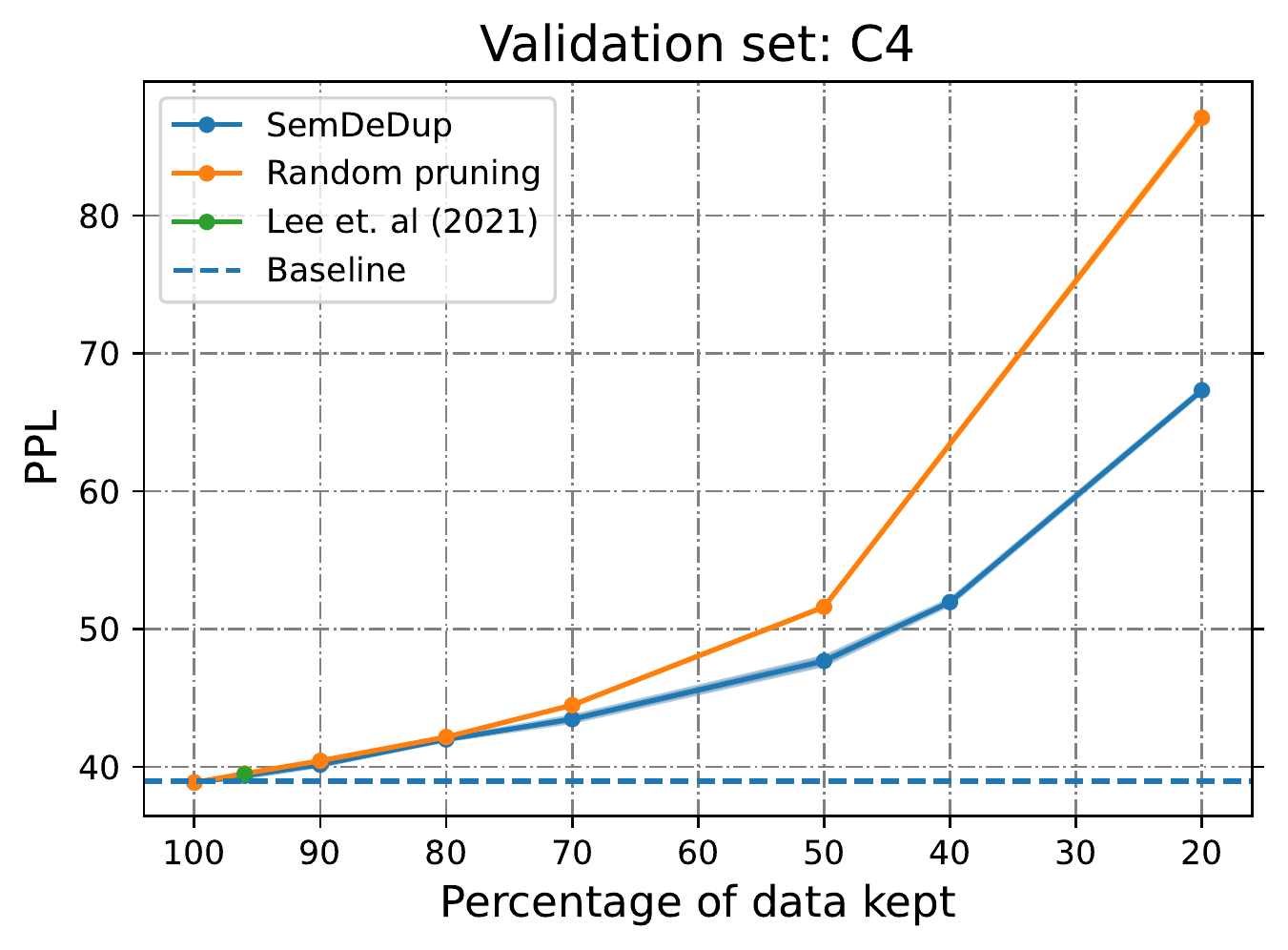}
    \caption{}
\end{subfigure}
\begin{subfigure}{.40\columnwidth}
    \centering
    \includegraphics[width=\columnwidth]{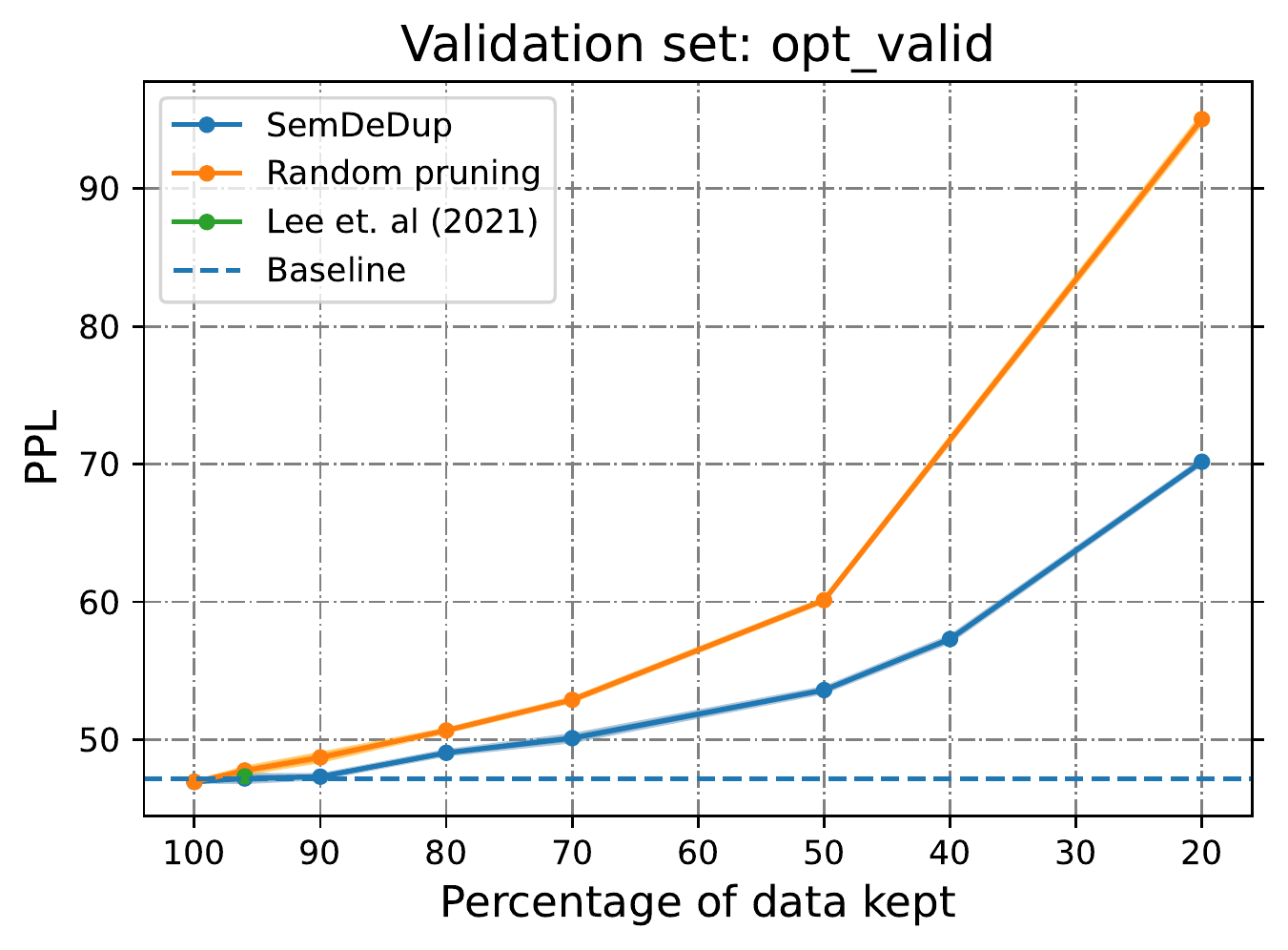}
    \caption{}
\end{subfigure}
\begin{subfigure}{.40\columnwidth}
    \centering
    \includegraphics[width=\columnwidth]{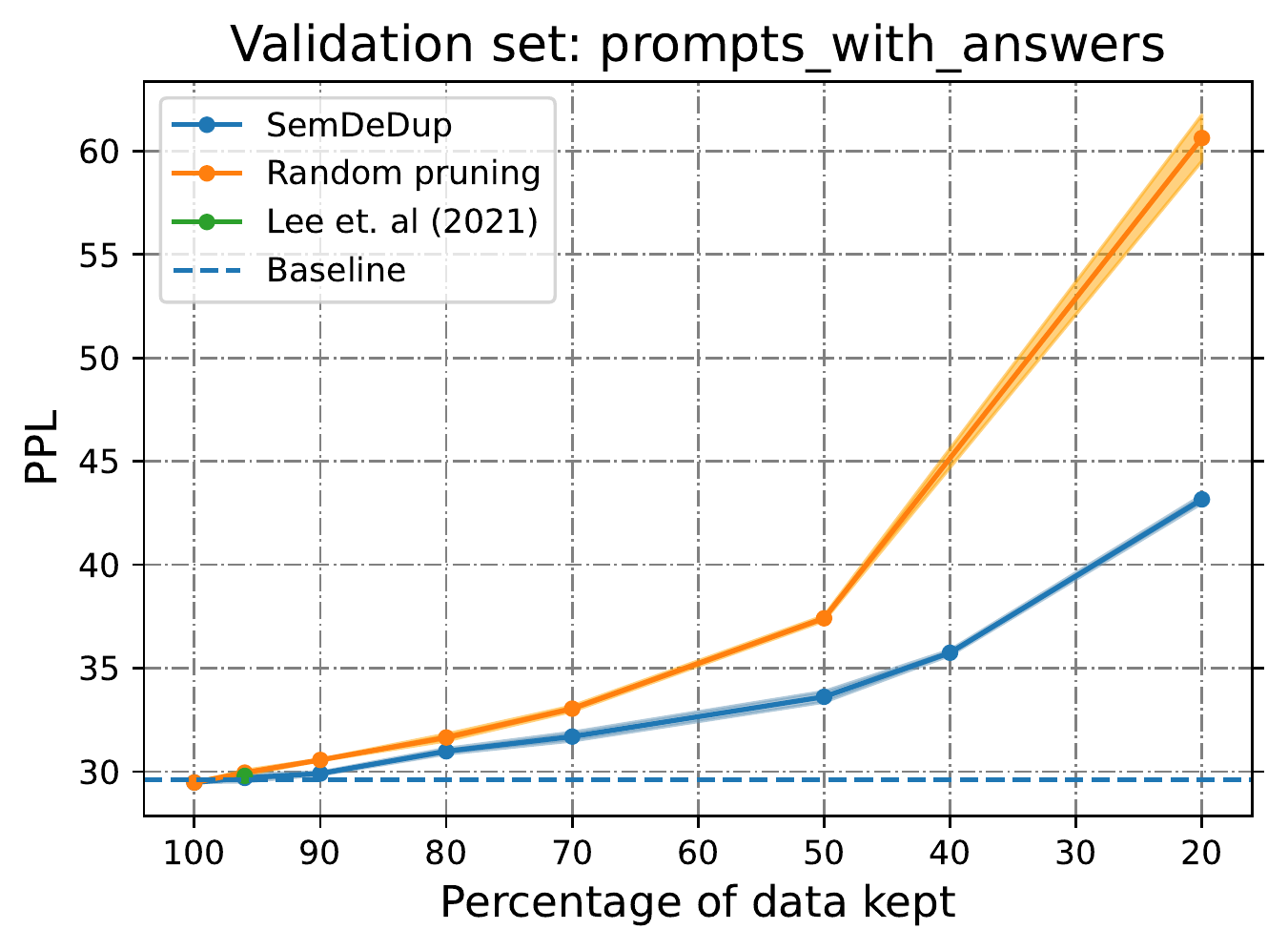}
    \caption{}
\end{subfigure}

\caption{\text{SemDeDup} performance at different fractions of data for the 125M OPT model. We show results for the C4 validation set (top left), opt\_valid (top right), and prompts\_with\_answers (bottoms). These are the same graphs as Figure~\ref{fig:nlp_125M_matched_epoch_c4}, but for a wider range of percentage of data kept. We note that \text{SemDeDup} consistently outperforms random pruning at lower percentages of data kept.}
\label{fig:nlp_appendix_125m_full_plots_matched_epoch}
\end{center}
\vskip -0.2in
\end{figure}

\begin{figure}[ht]
\vskip 0.2in
\begin{center}
\centering
\begin{subfigure}{.40\columnwidth}
    \centering
    \includegraphics[width=\columnwidth]{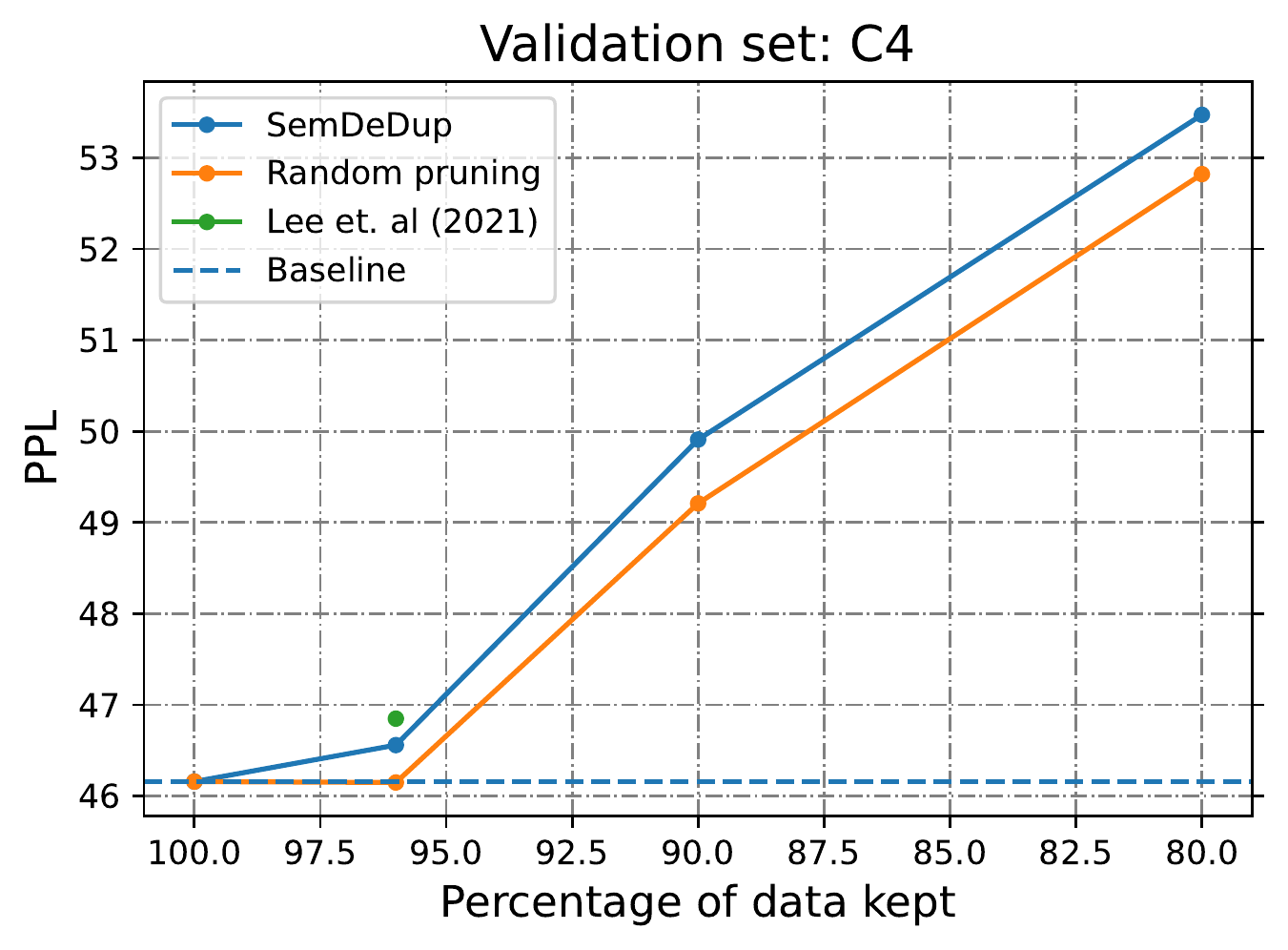}
    \caption{}
\end{subfigure}
\begin{subfigure}{.40\columnwidth}
    \centering
    \includegraphics[width=\columnwidth]{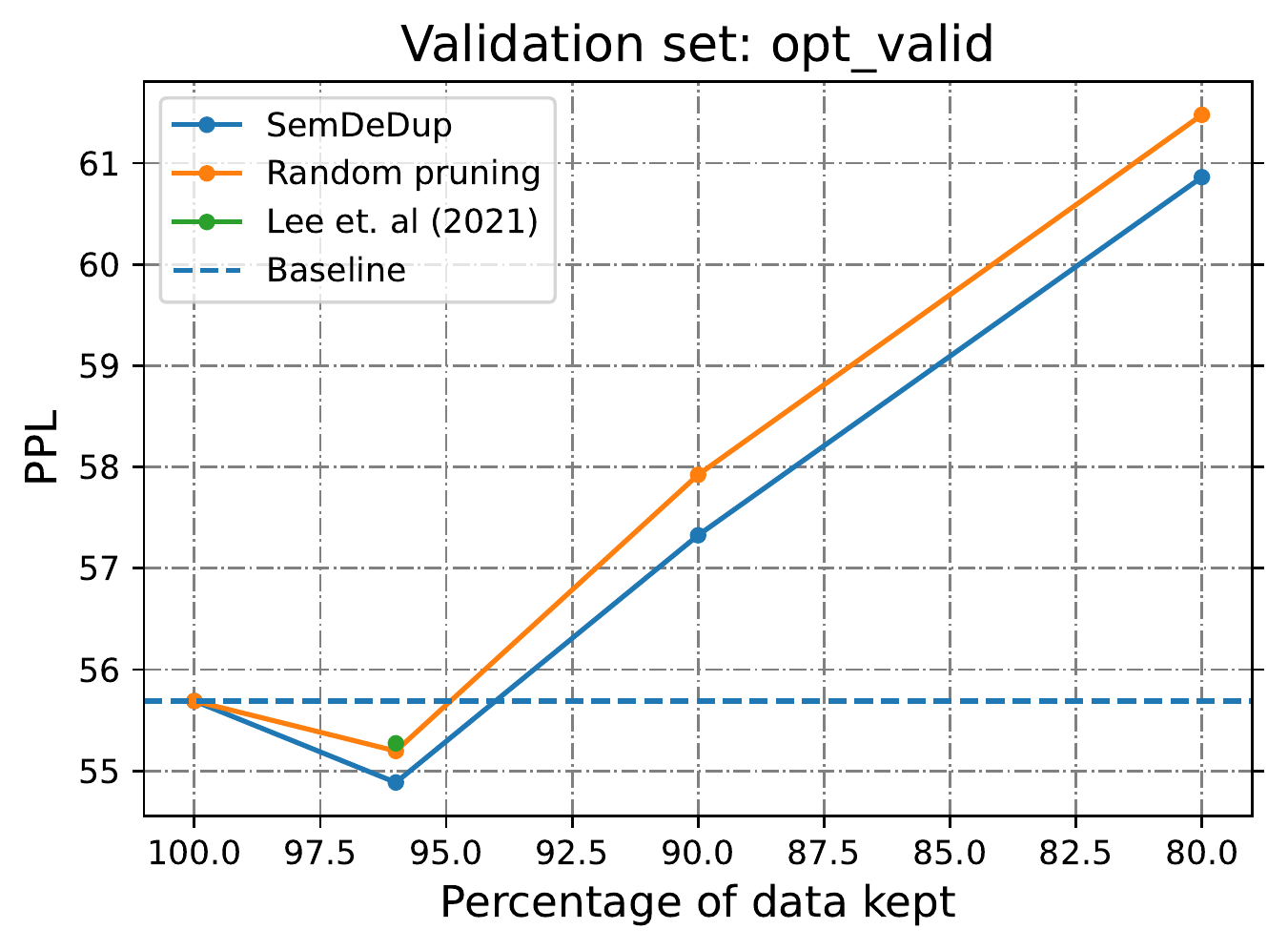}
    \caption{}
\end{subfigure}
\begin{subfigure}{.40\columnwidth}
    \centering
    \includegraphics[width=\columnwidth]{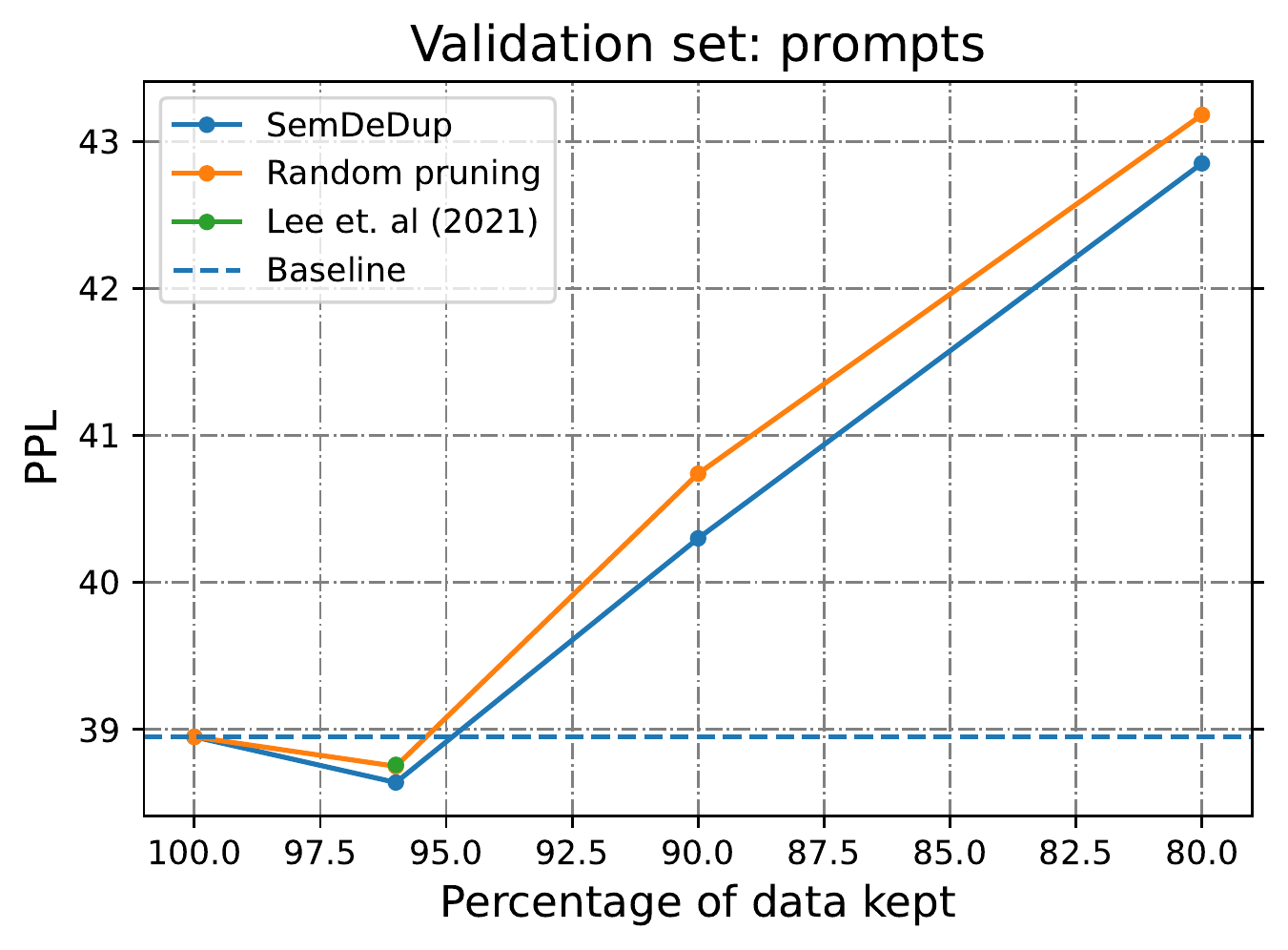}
    \caption{}
\end{subfigure}

\caption{\text{SemDeDup} performance at different fractions of data for the 1.3B OPT model. We show results for the C4 validation set (top left), opt\_valid (top right), and prompts\_with\_answers (bottoms). These are similar to tables~\ref{table:nlp_appendix_1.3b_c4_matched_epoch_96} and \ref{table:nlp_appendix_1.3b_c4_matched_epoch_20} but for a  range of percentage of data kept (96 \%, 90\%, 80\%). We note that \text{SemDeDup} consistently outperforms random pruning at lower percentages of data kept.}
\label{fig:nlp_appendix_1_3_b_full_plots_matched_epoch}
\end{center}
\vskip -0.2in
\end{figure}

\begin{figure}[ht]
\begin{center}
\includegraphics[width = 1.0\textwidth]{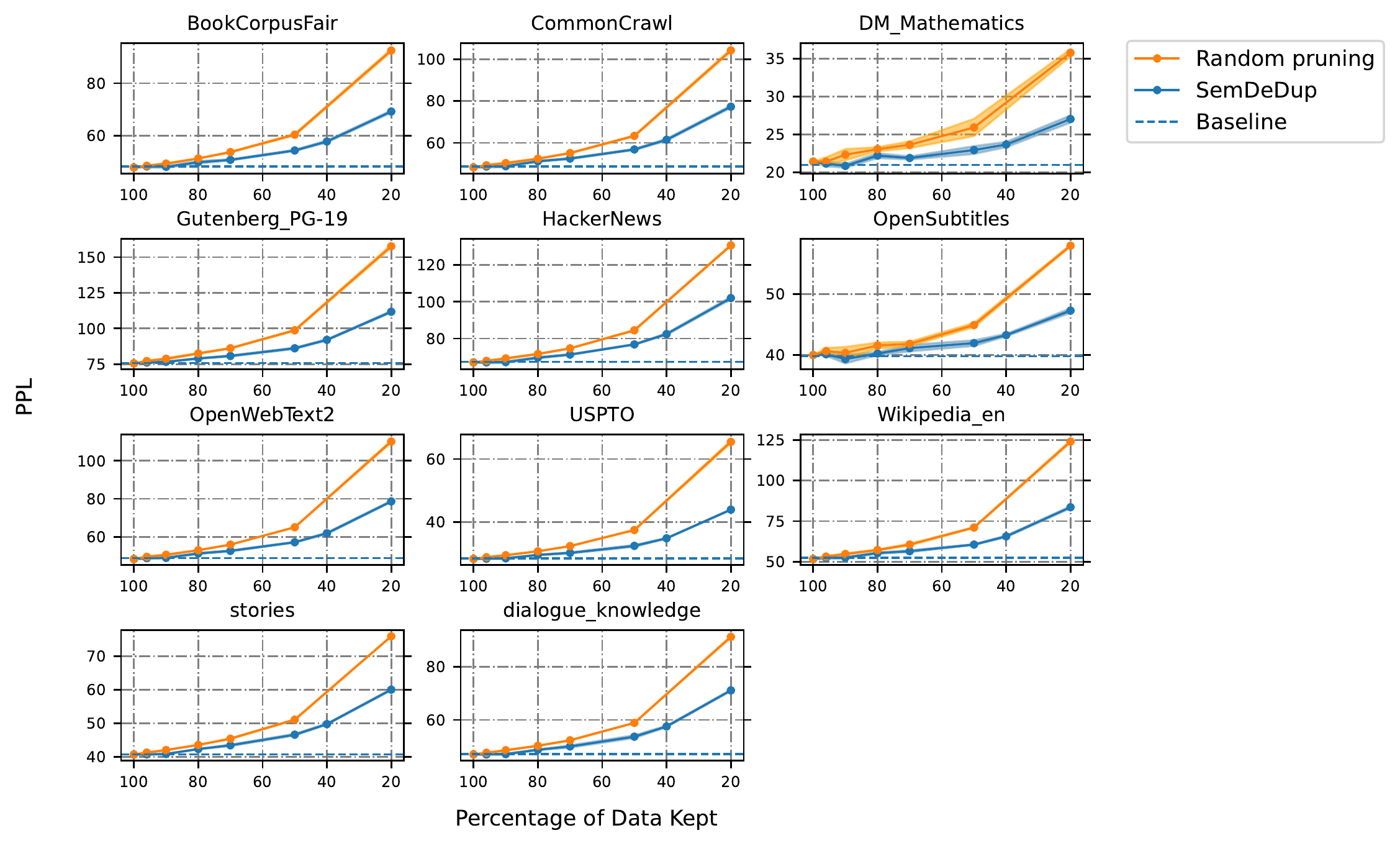}

\caption{Percentage of Data Kept vs. Perplexity on individual validation sets within opt\_valid. Runs are averages across 3 training seeds, and shaded regions represent 1 standard deviation from the mean. The title of each plot represents the name of the individual validation set within opt\_valid. Note that on all tasks, SemDedup significantly random pruning, especially at low percentages of data kept.}
\label{fig:nlp_appendix_big_ood_plot_matched_epochs}
\end{center}
\end{figure}

\begin{figure}[ht]
\begin{center}
\includegraphics[width = 1.0\textwidth]{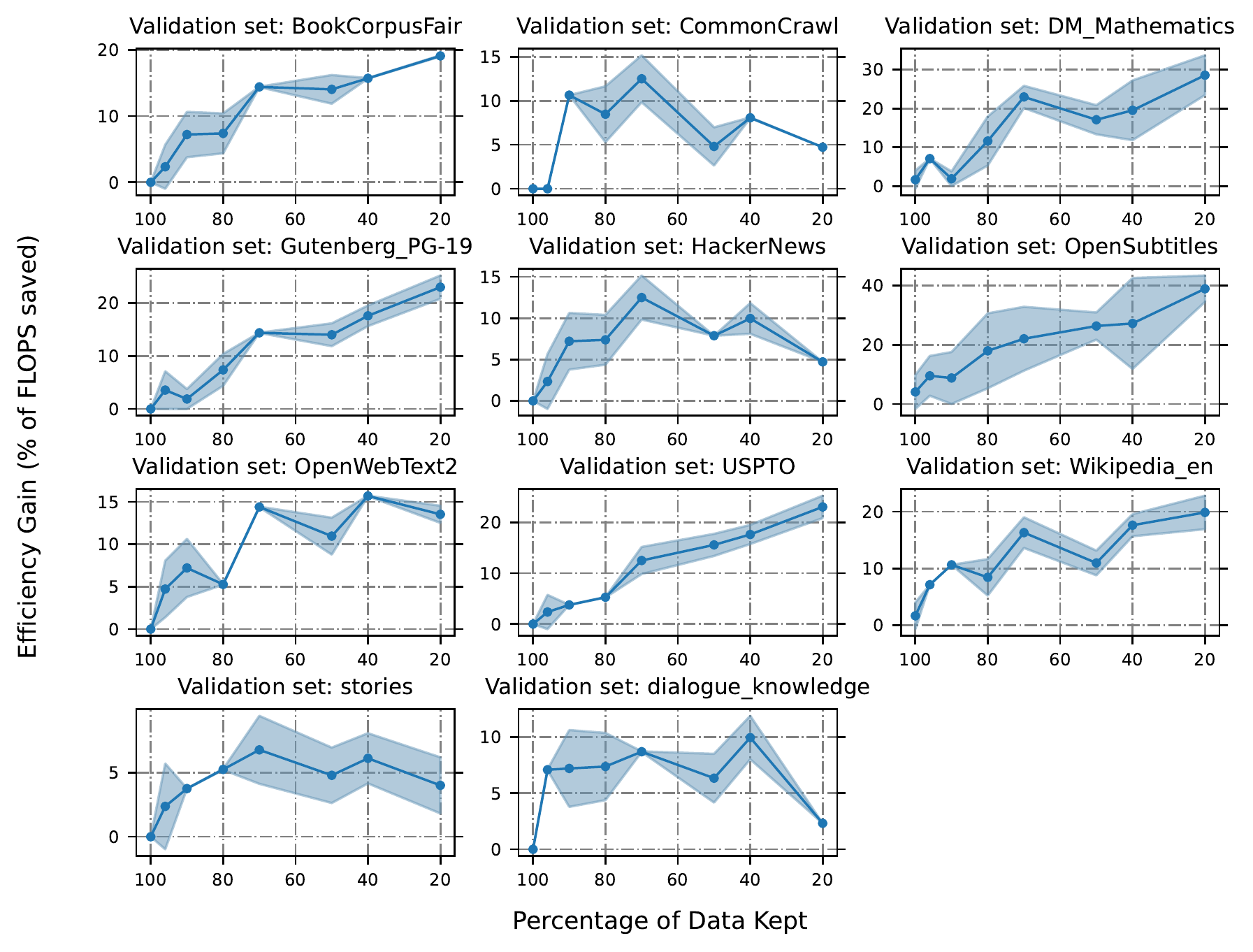}

\caption{Percentage of Data Kept vs. Efficiency Gain on individual validation sets within opt\_valid. Runs are averaged across training seeds where the model achieves baseline perplexity at some point in training, and shaded regions represent 1 standard deviation from the mean. The title of each plot represents the name of the individual validation set within opt\_valid.}
\label{fig:nlp_appendix_big_ood_plot_efficiency}
\end{center}
\end{figure}

\begin{figure}[ht]
\begin{center}
Keeping 90\% data
\begin{tabular}{L{14cm}}
\toprule
 text \\
\midrule
It appears that you already have an account on this site associated with . To connect your existing account... \\
\bottomrule
\end{tabular}

\vskip 0.5in
Keeping 100\% data (i.e. no pruning) 
\begin{tabular}{L{14cm}}
\toprule
text \\
\midrule
 It appears that you already have an account on this site associated with. To connect your existing account...   \\
 You are visiting the placeholder page for Wells Williams. This page is here because someone used our placeholder...\\
You are visiting the placeholder page for Mathew Barrett. This page is here because someone used our placeholder... \\
You are visiting the placeholder page for Marcus Slatar. This page is here because someone used our placeholder... \\
 You are visiting the placeholder page for Bernice Andrews. This page is here because someone used our placeholder... \\
 You are visiting the placeholder page for Emiko Chille. This page is here because someone used our placeholder... \\
 You are visiting the placeholder page for Landon Buckland. This page is here because
 Someone used our placeholder... \\
.... \\
You are visiting the placeholder page for Kylie Dickens. This page is here because someone used our placeholder utility ... \\
\bottomrule
\end{tabular}
\caption{Example of semantic de-duplication with SemDeDup (cluster 4500)}
\label{table:nlp_appendix_examples_one}
\end{center}
\end{figure}

\begin{figure}[ht]
\begin{center}
Keeping 20\% data
\begin{tabular}{L{14cm}}
\toprule
text \\
\midrule
cheap jordan shoes from china free shipping,order maroon foams , jordan blue retro 12 , jordans sz 10 , all white 14s... \\
Booming business thanks to Cristiano Ronaldo! Nike Presents Cristiano Ronaldo – CR7 Winter Collection. Cristiano Ronaldo ... \\
\bottomrule
\end{tabular}

\vskip 0.5in
Keeping 90\% data
\begin{tabular}{L{14cm}}
\toprule
 text \\
\midrule
Purchase from us, you can get max discount and free shipping.Free shipping and returns on Nike Jordans at Nordstrom.com.... \\
Product range. Adidas collections are divided into three groups: Sport Performance, Originals, and Sport Style. Originals that ... \\
cool jordans for boys , foamposite paranorman , new black and white foams , lebron 1's ,cheap jordans online for sale ... \\
This Comfortable Nike Huarache Free Basketball And Running has 1600 x 900 pixel resolution with jpeg format. .. \\
Top Rating: “Best high performance product.” Performance efficiency. That is the motto of our textile engineers by... \\
cheap jordan shoes online free shipping order cheap jordans for sale free shipping. Air Jordan 1's new theme color matching ... \\
... \\
Trendy Men's Nike Kyrie 1 Best Seller 'All Star' Multicolor at high discount. Buy Nike Trainers - The Kyrie 1 All Star comes ... \\
\bottomrule
\end{tabular}
\end{center}
\caption{Example of semantically redundant de-duplication with SemDeDup (cluster 4900)}
\label{table:nlp_appendix_examples_two}
\end{figure}

\section{K-means Clustering Details} \label{sec:kmeans_clustering}
 We use the $faiss$ library for clustering. $faiss$ is a library for efficient clustering on millions of vectors with GPU support. We use Spherical k-means as we found it better for clustering on ImageNet.  Spherical k-means normalizes the cluster centroids after every iteration to have a unit length. This requires the data to also be normalized before clustering. In all our experiments, we run 100 clustering iterations for LAION440M and 20 iterations for C4. We found that centroids do not move after this number of iterations.
\end{document}